\documentclass[journal]{IEEEtran}
\usepackage{subfig}
\usepackage{graphicx}

\usepackage{gensymb}
\usepackage{comment}

\usepackage{floatrow}
\usepackage{comment}
\usepackage[table,xcdraw]{xcolor}
\usepackage{pdfpages}
\usepackage{algorithm,algpseudocode}
\renewcommand{\algorithmicrequire}{\textbf{Input:}}
\renewcommand{\algorithmicensure}{\textbf{Output:}}
\graphicspath{{./Figures/}}

\usepackage{amsmath}
\usepackage{amsfonts}

\usepackage{bm}

\usepackage{cleveref}
\begin{document}
	%
	\title{Image-to-Height Domain Translation for Synthetic Aperture Sonar}
	%
	%
	%
	
	\author{Dylan~Stewart,~\IEEEmembership{Member,~IEEE,}
		Shawn~F.~Johnson,~\IEEEmembership{Senior~Member,~IEEE,}
		and~Alina~Zare,~\IEEEmembership{Senior~Member,~IEEE}
	}
	
	%
	%

	\markboth{Journal of \LaTeX\ Class Files,~Vol.~14, No.~8, August~2015}%
	{Shell \MakeLowercase{\textit{et al.}}: Bare Demo of IEEEtran.cls for IEEE Journals}
	%



	\maketitle

	\begin{abstract}
 	 Observations of seabed texture with synthetic aperture sonar are dependent upon several factors. In this work, we focus on collection geometry with respect to isotropic and anisotropic textures. The low grazing angle of the collection geometry, combined with orientation of the sonar path relative to anisotropic texture, poses a significant challenge for image-alignment and other multi-view scene understanding frameworks. We previously proposed using features captured from estimated seabed relief to improve scene understanding. While several methods have been developed to estimate seabed relief via intensity, no large-scale study exists in the literature. Furthermore, a dataset of coregistered seabed relief maps and sonar imagery is nonexistent to learn this domain translation. We address these problems by producing a large simulated dataset containing coregistered pairs of seabed relief and intensity maps from two unique sonar data simulation techniques. We apply three types of models, with varying complexity, to translate intensity imagery to seabed relief: a Gaussian Markov Random Field approach (GMRF), a conditional Generative Adversarial Network (cGAN), and UNet architectures. Methods are compared in reference to the coregistered simulated datasets using $L_1$ error. Additionally, predictions on simulated and real SAS imagery are shown. Finally, models are compared on two datasets of hand-aligned SAS imagery and evaluated in terms of $L_1$ error across multiple aspects in comparison to using intensity. Our comprehensive experiments show that the proposed UNet architectures outperform the GMRF and pix2pix cGAN models on seabed relief estimation for simulated and real SAS imagery.
	\end{abstract}
	

	%
	\IEEEpeerreviewmaketitle

	\section{Introduction}
\begin{figure*}[ht]
 		\centering
 		 \includegraphics[width=.3\linewidth]{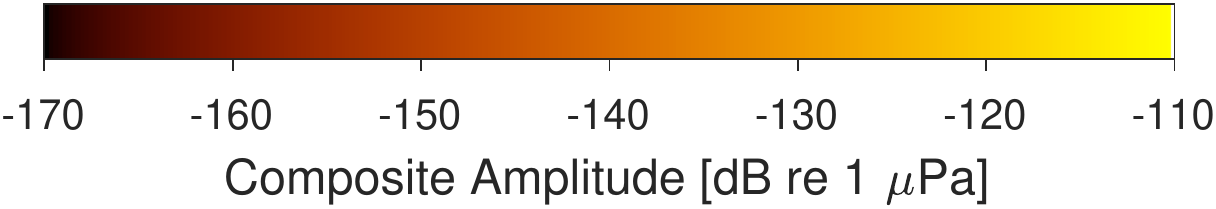}\\ 
 		\subfloat[Ripple response from the sensor. \label{fig:sonarv}]{\includegraphics[width=.5\linewidth]{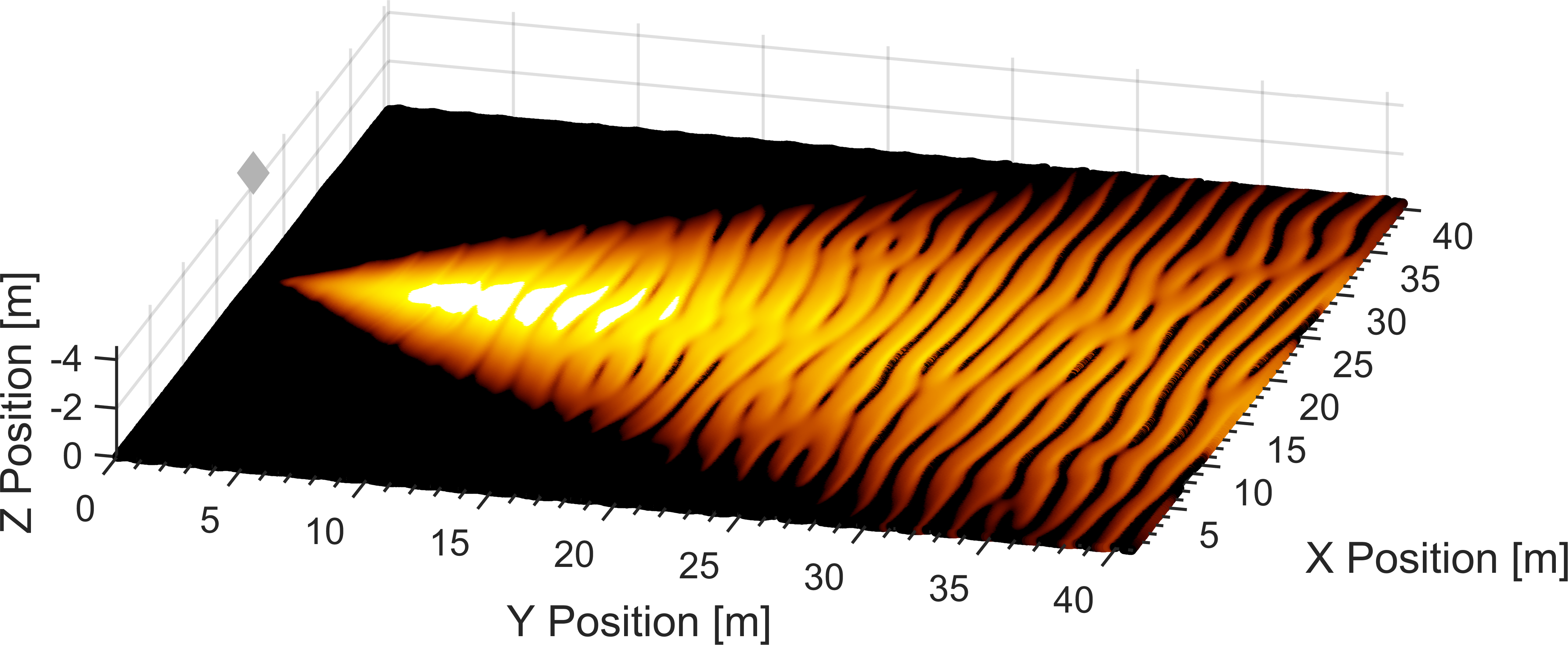}} 
 		\hfill
       \subfloat[Flat response from the sensor.\label{fig:sripple}]{\includegraphics[width=.5\linewidth]{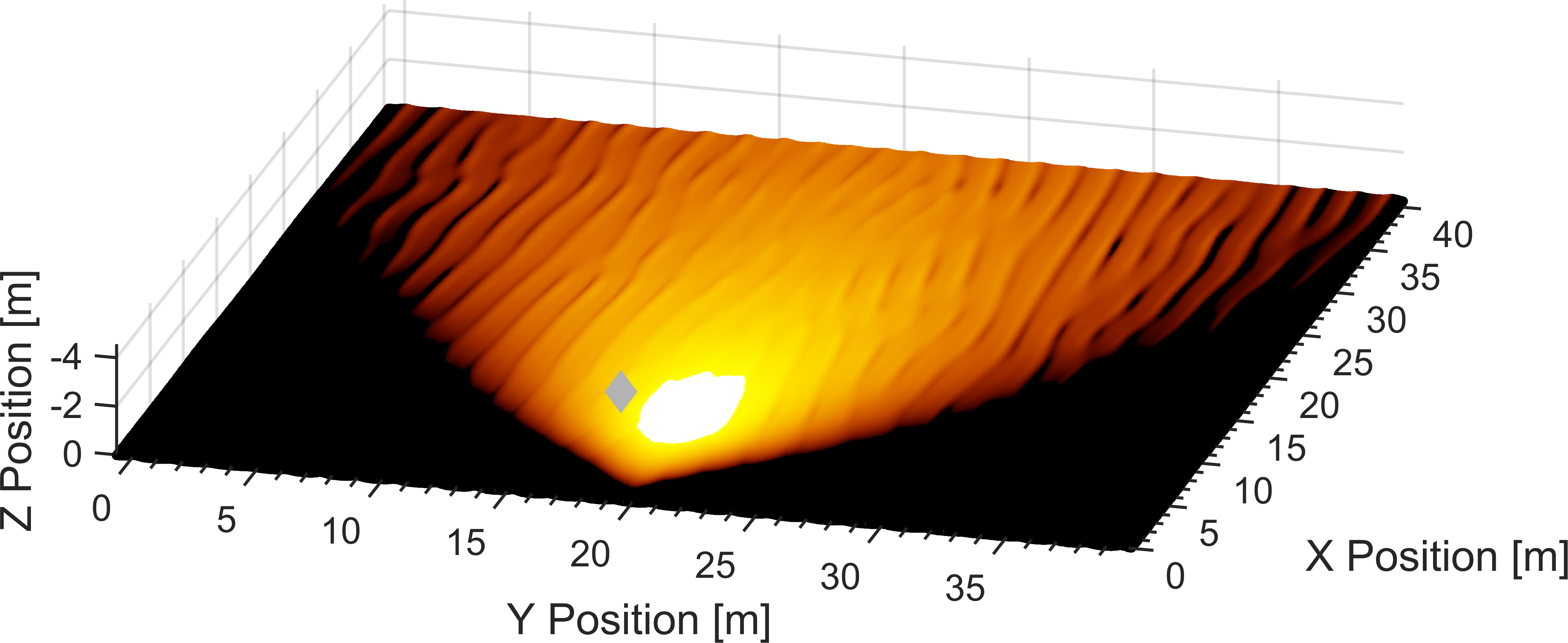}} 
 		\caption{Example intermediate products of PoSSM showing the composite acoustic model amplitudes for each of the seafloor points being evaluated. For the left image, the sonar along-track dimension is parallel to the ripple crests, producing a sharp contrast of the ensonified ripple texture. In the right image, the sonar along-track dimension is orthogonal to the ripple crests, which produces a more ``washed out" ensonification pattern. When the range of seafloor slopes is ensonified in combination with the grazing angle, a wider or narrower range of backscattering angles are visible for the left and right image respectively.}
 		\label{fig:2paths}
 	\end{figure*}

	%
	%
	%
	%
	\IEEEPARstart{S}{ynthetic} aperture sonar (SAS) surveys produce high-quality imagery of the seafloor. In this work, we study changes in sonar array direction. Due to the low grazing angle geometry of side-scan sensors, views of the same feature on the seafloor manifest as positional offsets. The troughs of the ripples are occluded. In regards to unnatural artifacts present within the look, seabed relief may obscure or occlude an object in the trough of the ripples. This phenomenon can be a significant issue when objects are smaller than the ripples they are located in \cite{Jackson02,Piper02}. For example, when the sonar along-track direction is parallel to the sand-ripple as shown in Figure \ref{fig:sonarv}, the troughs are not visible. However, when the along-track direction is orthogonal to the ripple propagation, the area appears flat (see Figure \ref{fig:sripple}). As demonstrated by this example, coregistered intensity from multiple looks, captured at different times and/or aspects may not correlate highly. This can lead to issues in image registration \cite{Tesfaye16} and noncoherent change detection \cite{Lyons13,Tesfaye19}. Reducing variability between multi-look SAS imagery may improve mine countermeasures, ocean health studies, and glacier monitoring. 

	Our team previously proposed a Gaussian Markov Random Field (GMRF) to produce aspect invariant features of SAS by estimating seabed relief from intensity imagery \cite{Chen2014invariant}. We hypothesize that using estimated seabed relief would provide improvements for scene understanding. In this study, we aim to improve upon our previous findings. We develop two coregistered datasets consisting of simulated seabed relief and intensity for use as training data for real multi-look SAS imagery. We compare the GMRF approach with two trainable methods for intensity to seabed relief translation: UNet architectures \cite{Unet} and a conditional Generative Adversarial Network (cGAN) \cite{pix2pix}. We analyze the ability of each approach to translate intensity to seabed relief in terms of $L_1$ error on simulated SAS imagery. We demonstrate the ability of the UNet architectures to produce estimated seabed relief information that varies less across aspect than intensity for real multi-look SAS. We apply UNet models to two multi-aspect datasets and demonstrate higher similarity between multi-look data than intensity data.

	\subsection{Related Work}
	Seabed relief estimation from sonar imagery within the literature is not a new problem. Over the past 25 years, many methods have been introduced to estimate seabed relief from intensity. These can be categorized as unsupervised and supervised methods.

	Several methods in the literature perform estimations of seabed relief without ground-truth data. Cuschieri and Hebert computed seabed relief from simple geometric estimations relying on the sonar range, altitude of the towfish, and swath of the sensor \cite{Cuschieri90}. Their method does not account for possible shadows nor cluttered environments. Langer and Hebert estimated a seabed relief map using a geometric reflection model and information about shadows and highlights \cite{Langer91}. Given a group of SAS images taken in a circular pattern around an object, the influence of shadows on seabed relief estimation should be reduced. Based on this assumption, the method developed by Langer and Hebert \cite{Langer91} was later extended to multiple images taken in a circular pattern \cite{Zerr96}.  Standard SAS surveys contain many shadows and occlusions that have high influence on the beamformed imagery. Bikonis \textit{et al.} proposed a seabed relief estimation technique which relies on localized information for each pixel seabed relief and utilizes the linear approaches from \cite{Pentland88} and \cite{TsaiShah93}. This approach does not take into account the gradients which exist in SAS data. Two recent methods rely on Expectation Maximization (EM) approaches \cite{Moon96}. Coiras \textit{et al.} use a Lambertian model and a multi-resolution optimization approach to extract the seabed roughness \cite{CoirasPetillot07}. At each iteration the reflectance map, roughness map, and the intensity map are updated using stochastic gradient descent. This varies from the recent work of our team \cite{Chen2014invariant} because the intensity map is updated rather than being fixed. 
	
	Multiple supervised methods have shown success in seabed relief estimation from SAS data. Johnson \textit{et al.} rely on sparse bathymetric data to produce an initial estimate for the elevation map. The elevation map is then refined to fit side-scan sonar imagery by minimizing a global error function \cite{Johnson1996}. The cost function is adapted from previous work by \cite{Horn89} and uses three constraints: an $L_2$-norm between the intensity and Lambertian scattering model, the smoothness constraint, and a second order constraint on the gradients of the estimated roughness in each direction. Li and Pai use the minimization methods of \cite{FrankotChellappa88} to improve existing bathymetric data from side-scan sonar pings \cite{LiPai91}. Dura \textit{et al.} modified the linear approach from Pentland \cite{Pentland88} to reconstruct seabed elevation in reference to ground-truth bathymetric maps \cite{DuraBell04}. Outside of SAS applications, many have recently used conditional Generative Adversarial Networks (cGANs) to perform image-to-image translation \cite{Mirza14}.
	
	In regular GANs, the discriminator and generator only use the input data and noise vector respectively. With a cGAN, the discriminator and generator use the input and condition their individual outputs based on the structure of the output. A few authors have used cGANs for domain translation. Denton \textit{et al.} stacked a pyramid of cGANs to produce natural images \cite{pix2pixGAN13}.  Radford \textit{et al.} generated images conditioned on keypoints and segmentation masks \cite{pix2pixGAN44}. Fick \textit{et al.} used a cGAN to map temporal hyperspectral data from one year to the next \cite{Fick19}. Salimans \textit{et al.} developed two cGANs and fused them. The Structure-GAN generates a surface normal map and the Style-GAN takes the surface normal map and generates the 2D image \cite{pix2pixGAN52}. Zhao \textit{et al.} colored grayscale images using cGANs \cite{pix2pixGAN63}. Recently, a novel approach was designed combining a UNet \cite{Unet} architecture as the generator network and a Patch-GAN architecture \cite{LiWand16} as the discriminator network \cite{pix2pix}. This approach is known as pix2pix and has been used for many applications including: background removal, palette generation, sketches to portraits, sketches to Pokemon, pose transformation, and producing photos. We include comparisons to the pix2pix network trained with a gradient-penalty, Wasserstein, and $L_2$ loss function by combining methods from Isola \textit{et al.} and Arjovsky \textit{et al.} \cite{pix2pix,arjovsky2017wasserstein}.
	\subsection{Seabed Relief Estimation from Gaussian Markov Random Field}
	Given a backscattering model, seabed relief can be estimated from SAS intensity \cite{Johnson1996}. Chen \textit{et al.} (our team) developed a Gaussian Markov Random Field (GMRF) approach to estimate seabed relief from an intensity image \cite{Chen2014invariant}. The change in seabed relief, $\Delta \mathcal{H}_{ij}$, can be estimated as shown in Equation \ref{eq:LamH},
	\begin{equation}
		\label{eq:LamH}
		\Delta \mathcal{H}_{ij} = 
		\begin{cases}
			\frac{\Delta U}{\tan(\arctan(\frac{A_s}{j})+\arccos(I_{ij}))} & \quad \text{if } M_{ij} = 0, \\
			-2 & \quad \text{otherwise,}
		\end{cases}
	\end{equation}
	where $\Delta U$ is the change in range from the UUV, $M_{ij}$ is the shadow map such that $M_{ij}=1$ if that pixel is shadowed and $M_{ij}=0$ otherwise. The height of the sonar vehicle, $A_s$ comes from metadata. The shadow map is computed by calculating the mean and variance in a local window. Any pixel in that window that has small mean and small variance below some chosen threshold is considered a shadowed pixel. Based on this change in seabed relief, the estimated seabed relief is from the row-wise cumulative sum shown in Equation \ref{eq:LamHest},
	\begin{equation}
		\label{eq:LamHest}
		\hat{\mathcal{H}}_{ij} = \sum_{k=1}^j \Delta \mathcal{H}_{ik}.
	\end{equation} 
	
	After estimating the initial seabed relief profile, the GMRF parameters are updated to refine the seabed relief map using EM with iterated conditional modes \cite{Li2009,ICMBesag}. The expectation step improves the quality of the seabed relief map while the maximization step updates the GMRF parameters. These updates are computed using Maximum pseudo-likelihood estimation (MPLE). MPLE is the product of local conditional pdfs at each pixel given the neighbors. Given updated GMRF parameters, the seabed relief map can be refined. Two errors are considered to alter seabed relief maps, occlusion error and reconstruction error. Shadowed pixels where the seabed relief is unknown causes occlusion errors. Noise and beamforming artifacts increase reconstruction error. These errors are treated as Gaussian random variables with zero-mean and the variance is estimated. The GMRF parameters are updated accordingly to update the seabed relief map.
	\section{Datasets}
	
	We utilized simulated synthetic aperture sonar imagery for training our models. Each simulated sample used in our approach consists of a height map and intensity map. This allows us to leverage "perfect" ground-truth where each pixel is shared across the two domains of interest to train domain translation models. Collected imagery is used as test sets for the various approaches and will be described in Section \ref{sec:Experiments}. In this section, we describe the simulated dataset generation approaches.
	
	Two distinct approaches were utilized to generate the simulated imagery dataset used here: a lower-physics fidelity approach which \textit{emulates} the synthetic aperture image formation process, and a higher-physics fidelity approach which outputs time-series suitable for \textit{stimulating} synthetic aperture image formation algorithms. Both approaches utilize a simulated seabed texture height-map to produce their respective co-registered imagery outputs. A total of 148,000 m$^2$ of seafloor textures and 296,000 m$^2$ of simulated imagery were generated.

	A variety of seabed texture height-maps were generated which serve as input to each of the two image simulation approaches. Each of the height-maps consist of rippled-sand texture \cite{heightMapGen}, power-law texture \cite{Jackson_2007}, or a mixture of the two. To achieve our goal of studying multi-aspect sonar imagery, we simulated 37 rotations each of 10 textures (see Table \ref{tab:HeightMap}).
	
	\begin{table}[h]
		\centering
		\caption{Base and range of parameters for simulated seabed relief maps. All base parameters were obtained from Johnson \cite{SJohnsonThesis}. Each individual texture has a unique combination of dominant ripple wavelength and spectral strength of roughness. Six unique sand-ripple textures were produced by fixing the spectral strength and varying the dominant wavelength. The wavelength was increased by 0.35 m for each of the six ripple textures while the spectral strength was fixed at the base parameter. Four roughness textures were simulated by changing the spectral strength by a factor of 100 and negating the sand-ripple component.}
		\label{tab:HeightMap}
		\begin{tabular}{ccccc}
			\multicolumn{5}{c}{Simulated Height-Map Parameters}                    \\
			Par.   & Base     & Lower & Step       & Upper  \\ \hline
			\rowcolor[HTML]{E0E0E0} 
			$\lambda_0$ & 1.14 m  & 0.25 m      & 0.35 m     & 2 m         \\
			\rowcolor[HTML]{FFFFFF} 
			$\phi_r$  & 0 $\degree$ & $-90\degree$   & $5\degree$    & $90\degree$    \\
			\rowcolor[HTML]{E0E0E0} 
			$\omega_2$  & 4.3E-5 m$^4$  & 4.3E-12 m$^4$    & 100$\times$ & 4.3E-6 m$^4$     \\
		\end{tabular}
	\end{table}

	The first of two simulated imagery datasets were generated using the height-maps described above combined with the pseudo-image approach, which we refer to as PISAS, developed by Johnson \cite{SJohnsonThesis,PISAS10,PISAS11}. The purpose of the pseudo-image technique is the ability to \textit{emulate} the output of a synthetic aperture imagery formation process. Pseudo-image formation is a computationally efficient process by making several approximations to the physics involved with acoustic propagation and scattering and directly outputting an image product that can approximate a SAS image. Although PISAS imagery is "low" physics fidelity, it conveys several key attributes of a collected SAS image: the seafloor scattering response at low-grazing angles dependence on slope, constant-with-range resolution, and an approximation of coherent-imagine speckle.
	
	The second of two simulated imagery datasets was generated using the height maps described above with the Point-based Sonar Scattering Model developed by Johnson \textit{et. al} \cite{PoSSM17,PoSSM18,PoSSM19}. The model outputs calibrated time-series for each sonar hydrophone. We then applied synthetic aperture image formation techniques, effectively \textit{stimulating} a representative SAS signal processing chain. This approach captures significantly more of the acoustic propagation and scattering physics, including two-dimensional sonar transmit and receive beampatterns. Additionally, these "high" physics fidelity images are expected to more closely represent collected SAS imagery. 
	
	After image formation, seabed relief and intensity maps are tiled into $256 \times 256$ sized tiles. The datasets were split into five cross-validation folds. A held out set of 500 coregistered pairs of intensity and seabed relief maps from each set are used for testing. For each fold, 22,205 pairs are used for training and 1,000 pairs are used for validation. For each training split, the maximum and mininum of the intensity and seabed relief map for each beamformer is recorded. Samples are normalized between zero and one by subtracting the minimum and dividing by the maximum.

	\section{Methods}
	Our goal is to learn the function, $f(I)\rightarrow\mathcal{H}$, where $I \in \mathbb{R}^{256\times256}$ is an input intensity image and $\mathcal{H} \in \mathbb{R}^{256\times256}$ is the desired seabed relief map. To learn this mapping, we compare a variety of UNet architectures to a GMRF model and the pix2pix cGAN model. We aim to obtain the least complex model that can perform domain translation of intensity to seabed relief.
	
	In a traditional convolutional autoencoder, convolutional layers are used to project a given input to a lower dimensional space at the bottleneck layer and deconvolutional layers upsample the projected vector back to a desired space. A UNet is a special type of convolutional autoencoder that contains skip connections \cite{Unet}. The skip connections in a UNet are used to minimize the loss of spatial information on the decoder side of the network. Outputs with larger spatial windows from the encoder side are concatenated onto upsampled inputs at later layers of the network to reduce the loss of spatial information. 
	
\begin{figure*}[t]
	\includegraphics[width=\linewidth]{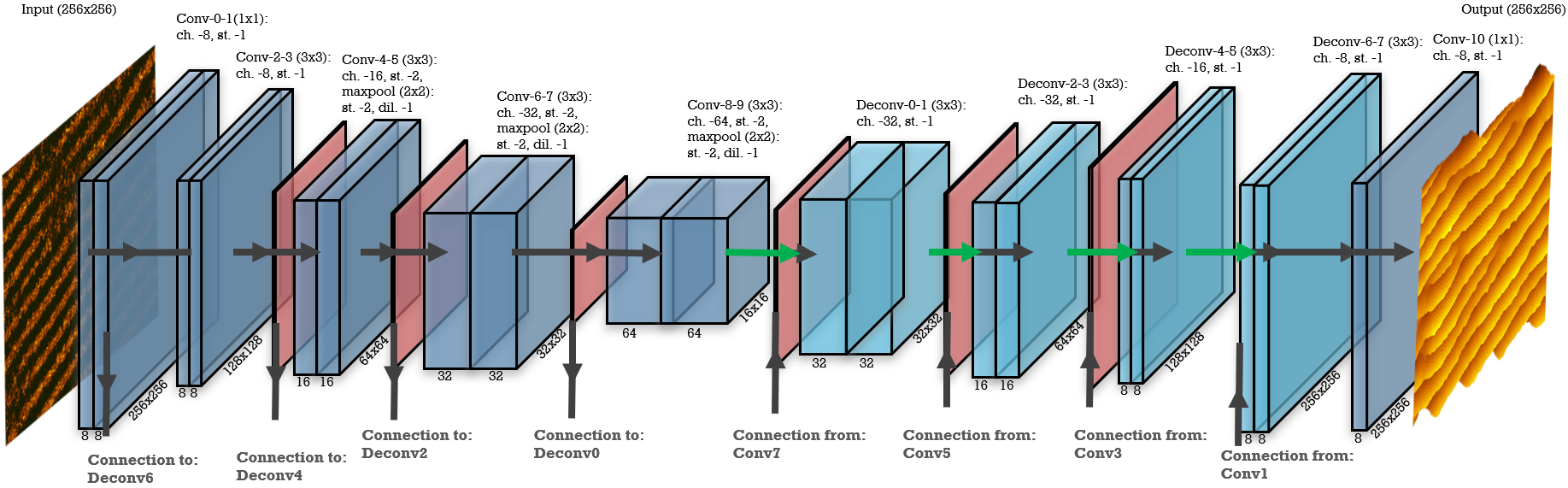}
	\caption{Diagram of the UNet-opt architecture for translating intensity to seabed relief. Dark blue boxes are standard convolutional layers, red boxes are $2\times 2$ maxpooling layers, and light blue boxes are deconvolutional layers. Each convolutional and deconvolutional layer is followed by a batch normalization layer (not shown in the figure for clarity). Each gray arrow is a standard feedforward connection. Green arrows are connections where the output of the previous layer is upsampled by a factor of two and fed into the following layer. }
	\label{fig:UNetArch} 
\end{figure*}

	In this work, we compare variations of a standard UNet model where each subsequent model has a variety of different channels and depths. These modifications are proposed in order to produce the simplest model that is robust on our dataset. Descriptions for each of the proposed models are provided in Table \ref{tab:UNetModels}. We provide a diagram of the best performing architecture (see Figure \ref{fig:UNetArch}). We will refer to this architecture as UNet-opt (opt meaning optimal). The details of each layer of UNet-opt are provided in Table \ref{tab:UNetDetails}.

	The network architecture can be summarized in two stages. The first stage sequentially encodes the input into smaller spatial scales with more features until the bottleneck layer. At the bottleneck, the spatial size is smallest with the most number of features. The second stage decodes the representation at the bottleneck back to the desired space through increasing the spatial size and decreasing the number of features. 
	
	The first stage consists of 10 convolutional layers. Each convolutional layer has kernel size of three and zero padding followed by a batchnorm and ReLU activation. Convolutional layers are stacked in sequential pairs. Other than the first, each pair is followed by a maxpooling layer with window size two. The outputs of each pair of convolutional layers from the first stage are connected via skip connections to inputs of the second stage. 
	
	The second stage consists of eight deconvolutional layers and an output convolutional layer. Each deconvolutional layer consists of a pair of inputs. The first input is from the previous sequential layer and the second is from a skip connection. The first input is upsampled with scale factor of two and bilinear interpolation is applied to fill in missing samples. The second input, which is via a skip connection (see Figure \ref{fig:UNetArch} or Table \ref{tab:UNetDetails} for specific connections), is concatenated onto the upsampled tensor. The output convolutional layer has a kernel size of one and projects the data from a multi-channel representation to single-channel.
	
	During training, batches of coregistered intensity and seabed relief maps are used as the input and desired samples respectively. We fixed the batch size at 20 due to memory and loading constraints. The average $L_1$ loss is used as the error measurement between predicted and desired samples. We selected the $L_1$ loss rather than a more complex loss function due to the success of using $L_1$ regularization for cGAN architectures \cite{pix2pix}. The $L_1$ loss is shown in Equation \ref{eq:lossfn},
	\begin{equation}
		\label{eq:lossfn}
L(\mathcal{H},\hat{\mathcal{H}}) =\frac{1}{B}\sum_{i=1}^B|\mathcal{H}_i-\hat{\mathcal{H}}_i|,
	\end{equation}
	where $B$ is the batch size.
	
	After training the entire network end-to-end with the PISAS dataset, we compared fine tuning schemes using the PoSSM dataset. One approach relies on fine-tuning the entire network \cite{Li2018}. We also fine-tune the output convolutional layer while fixing all other weights and biases \cite{amiri2020}.

	\begin{table}[h]
		\caption{\label{tab:UNetModels}  Descriptions of each UNet model as well as the bottleneck size and number of trainable parameters for our experiments. Channel-reduced (c) and depth-reduced (d) have been shorted for brevity. Each number of channels is reduced by a factor of two between the UNet-opt and UNet-c2 model. The depth is reduced by removing two convolutional layers from the first stage and two deconvolutional layers from the second stage of the UNet-c2 model. This produces the UNet-d model. Additional models (UNet-d2 and UNet-d3) are of the same structure as the UNet-d model but each channel depth is increased by a factor of two respectively.}
	\begin{tabular}{cccc}
		\multicolumn{4}{c}{UNet model descriptions}   \\
		Model    & Desc.     & Bn. & Par.      \\ \hline
		\rowcolor[HTML]{E0E0E0} 
		UNet     & 8C + 8DC & 256        & 1,081,745 \\
		\rowcolor[HTML]{FFFFFF} 
		UNet-opt     & 8C + 8DC  & 128        & 271,305   \\
		\rowcolor[HTML]{E0E0E0} 
		UNet-c2 & 8C + 8DC  & 64         & 68,261    \\
		\rowcolor[HTML]{FFFFFF} 
		UNet-d  & 6C + 6DC  & 64         & 67,689    \\
		\rowcolor[HTML]{E0E0E0} 
		UNet-d2 & 6C + 6DC  & 128        & 269,009   \\
		\rowcolor[HTML]{FFFFFF} 
		UNet-d3 & 6C + 6DC  & 256        & 1,072,545
	\end{tabular}
	\end{table} 

\begin{table*}[h!]
	\begin{tabular}{cccccccccc}
		\multicolumn{5}{c}{Layer Details}                & \multicolumn{5}{c}{Layer Details}               \\
		Name     & Ker.       & Ch. & Param. & Input(s)  & Name     & Ker.       & Ch. & Param. & Input(s) \\ \hline
		\rowcolor[HTML]{E0E0E0} 
		Conv-0   & 3$\times$3        & 8   & 80     & Img       & Conv-1   & 3$\times$3        & 8   & 584    & C-0      \\
		\rowcolor[HTML]{FFFFFF} 
		Conv-2   & 3$\times$3        & 16  & 1168   & C-1       & Conv-3   & 3$\times$3 & 16  & 2320   & C-2      \\
		\rowcolor[HTML]{E0E0E0} 
		Conv-4   & 3$\times$3 & 32  & 4640   & C-3       & Conv-5   & 3$\times$3 & 32  & 9248   & C-4      \\
		\rowcolor[HTML]{FFFFFF} 
		Conv-6   & 3$\times$3 & 64  & 18496  & C-5       & Conv-7   & 3$\times$3 & 64  & 36928  & C-6      \\
		\rowcolor[HTML]{E0E0E0} 
		Conv-8   & 3$\times$3 & 64  & 36928  & C-7       & Conv-9   & 3$\times$3 & 64  & 36928  & C-8      \\
		\rowcolor[HTML]{FFFFFF} 
		Deconv-0 & 3$\times$3 & 64  & 73792  & C-7, C-9  & Deconv-1 & 3$\times$3 & 64  & 18464  & DC-0     \\
		\rowcolor[HTML]{E0E0E0} 
		Deconv-2 & 3$\times$3 & 32  & 18464  & C-5, DC-1 & Deconv-3 & 3$\times$3 & 16  & 4624   & DC-2     \\
		\rowcolor[HTML]{FFFFFF} 
		Deconv-4 & 3$\times$3 & 16  & 4624   & C-3, DC-3 & Deconv-5 & 3$\times$3 & 8   & 1160   & DC-4     \\
		\rowcolor[HTML]{E0E0E0} 
		Deconv-6 & 3$\times$3 & 8   & 1160   & C-1, DC-5 & Deconv-7 & 3$\times$3 & 8   & 584    & DC-6     \\
		\rowcolor[HTML]{FFFFFF} 
		Conv-10  & 1$\times$1 & 1   & 9      & DC-7      &          &            &     &        &          \\ \hline
	\end{tabular}
	\label{tab:UNetDetails}  
	\caption{A summary of each layer for the UNet-opt architecture. Some input names are shortened for brevity. Convolutional layers are shortened to C and deconvolutional layers are represented by DC.}
\end{table*}  
     
	\section{Experiments}
	\label{sec:Experiments}
	In the following, the GMRF model, pix2pix model, and variations of the UNet are evaluated and compared using simulated and real SAS data. We trained the proposed models on the PISAS dataset and evaluated their performance using five-fold cross-validation. For each type of model, the one which performed the best on the validation set was selected for comparison. Testing error for each of the best models on our PISAS dataset is shown in Table \ref{tab:testPISAS}. Given pre-training on the PISAS dataset, we fine-tune the UNet-opt model with the PoSSM dataset and evaluate performance on two multi-aspect datasets. One consists of hand-aligned side-scan SAS and the other contains pairs of hand-aligned circular SAS.
	
	\subsection{UNet Architecture Selection}
	All UNet models were trained for 500 epochs. During preliminary experiments, we experimented with training up to 2000 epochs. However, no significant decrease in the loss was recorded after 500 epochs. Additionally, we used an initial learning rate of 0.1 with the Adam optimizer with default values \cite{Adam}. Training was performed on a single NVidia RTX 2080 Ti with 11Gb of VRAM. 
	
	We compare each variation of the UNet architecture to the Gaussian Markov Random Field model \cite{Chen2014invariant}. Rather than relying on training thousands of parameters, the GMRF model estimates seabed relief information directly from the intensity given an initial seabed relief estimate and a specific number of iterations. A grid search was conducted for the best initial seabed relief parameter and number of iterations to train the model. The parameters that produced the smallest error in the validation set were selected for testing. During experiments with the training data, the estimated seabed relief did not change after 50 updates of the GMRF model and the best initial seabed relief was $A_s = 0.8$. 
	
	Additionally, we trained a pix2pix cGAN model for 4000 epochs. The generator network used for the cGAN is same architecture as the UNet-opt model. The discriminator is a standard patchGAN \cite{LiWand16}. A patchGAN attempts to classify patches of an image as real or fake rather than individual pixels or an entire image at a time. The patch size was fixed at $70\times 70$ pixels. The loss function to train the pix2pix network is a Wasserstein cGAN loss with $L_1$ regularization. Similar to the literature, we train the discriminator several iterations for each iteration the generator is trained \cite{arjovsky2017wasserstein}. During validation we found that training the discriminator 10 times for each generator iteration provided stable training. To select the best parameters, the $L_1$ loss between generated samples and the validation seabed relief maps were used. In each run, the validation error increased after 3000 epochs, therefore we used the best pix2pix model trained for 3000 epochs in our comparisons.  

\begin{figure}[h]
	\centering
	\subfloat[$I_1$\label{fig:Intensity1}]{%
		\includegraphics[trim={1cm 1cm 1cm 1cm},clip,width=0.17\linewidth,height=.05\textheight]{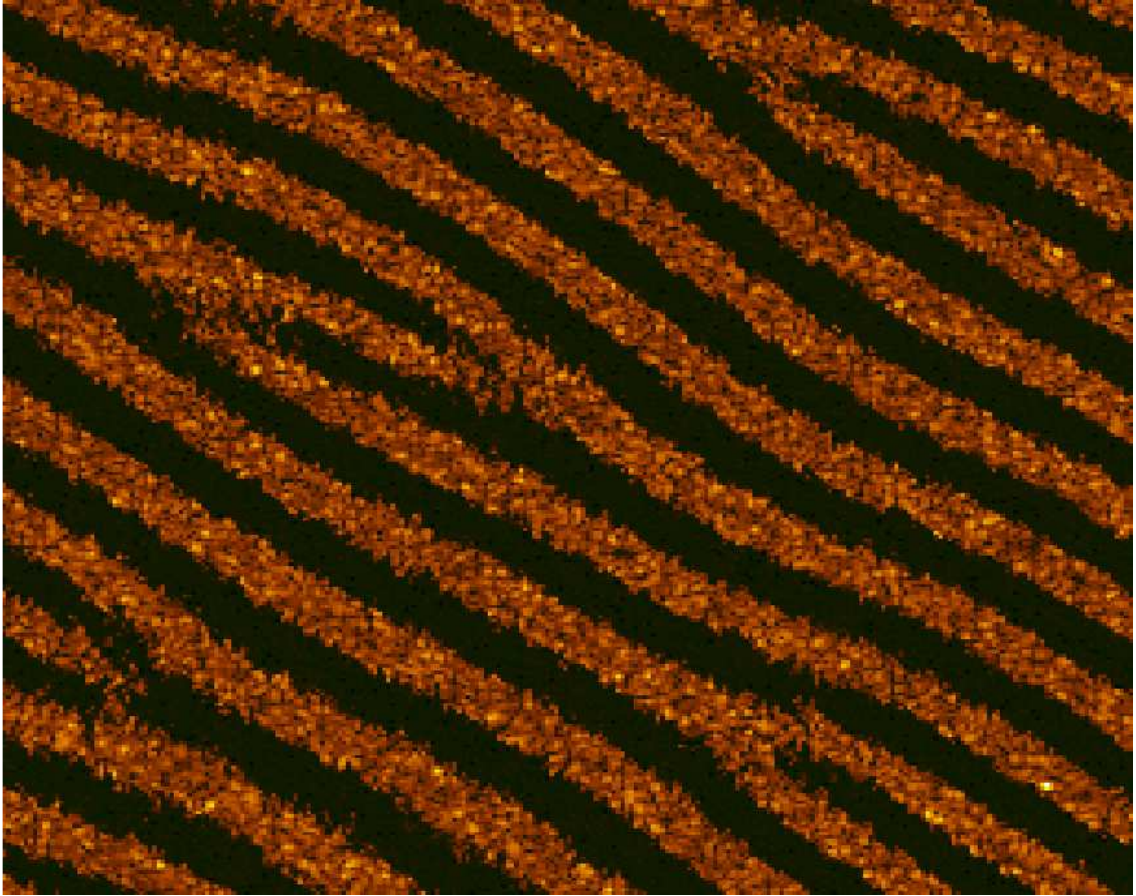}}
	\hfill
	\subfloat[$\mathcal{H}_1$\label{fig:Height1}]{%
		\includegraphics[trim={1cm 1cm 2cm 1cm},clip,width=0.17\linewidth,height=.05\textheight]{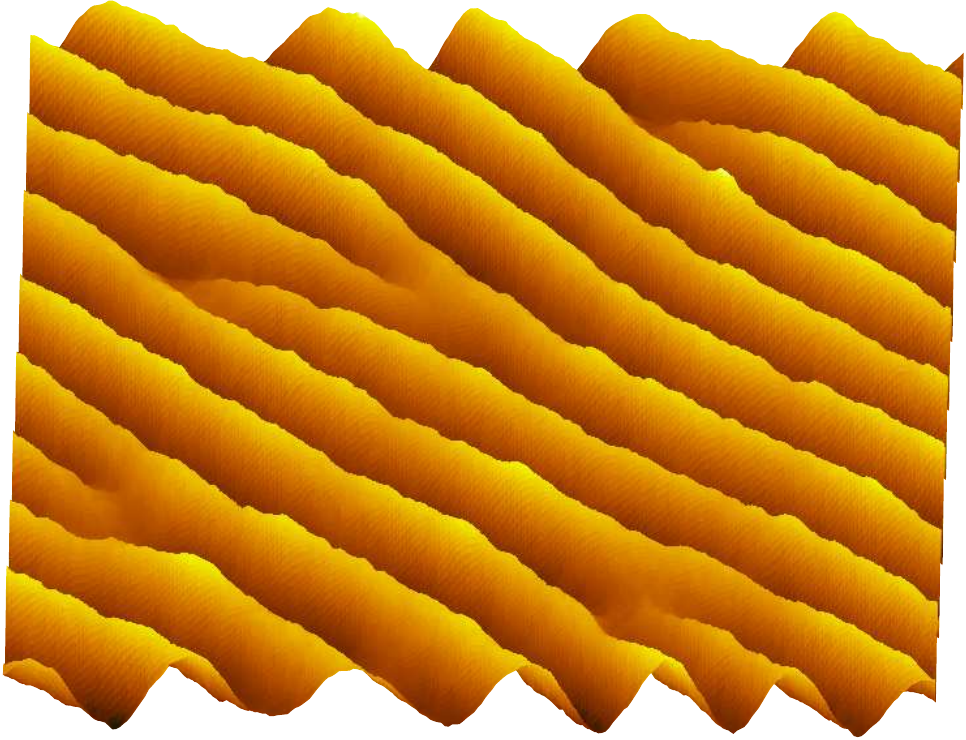}}
	\hfill
	\subfloat[GMRF\label{fig:GMRFestimate1}]{%
		\includegraphics[trim={1cm 1cm 2cm 1cm},clip,width=0.17\linewidth,height=.05\textheight]{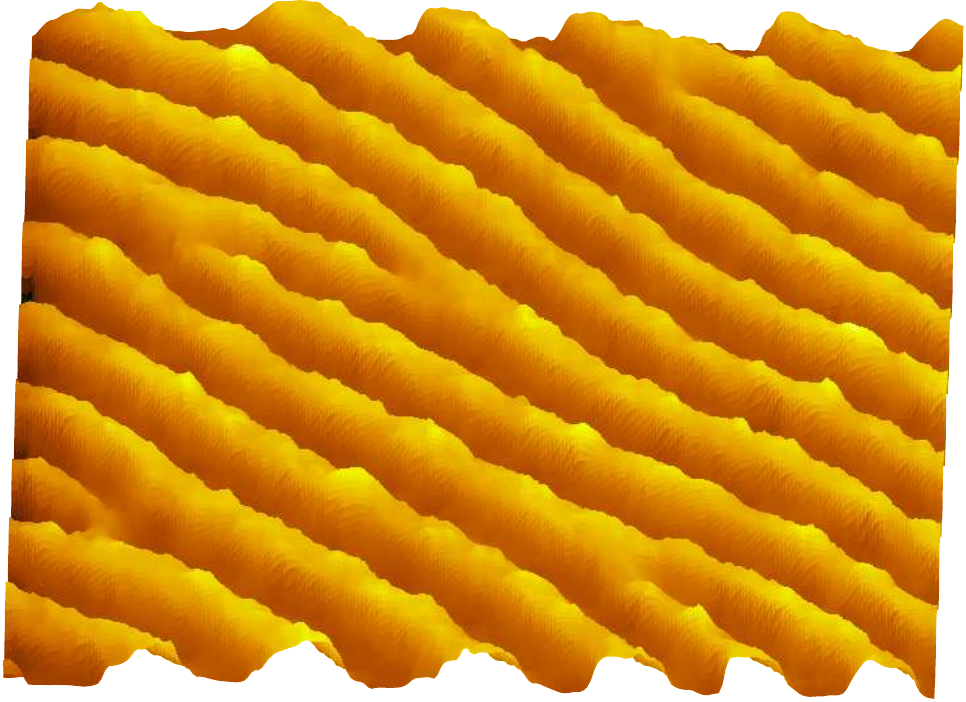}}
	\hfill
	\subfloat[pix2pix\label{fig:UNetestimate1}]{%
		\includegraphics[trim={1cm 1cm 2cm 1cm},clip,width=0.17\linewidth,height=.05\textheight]{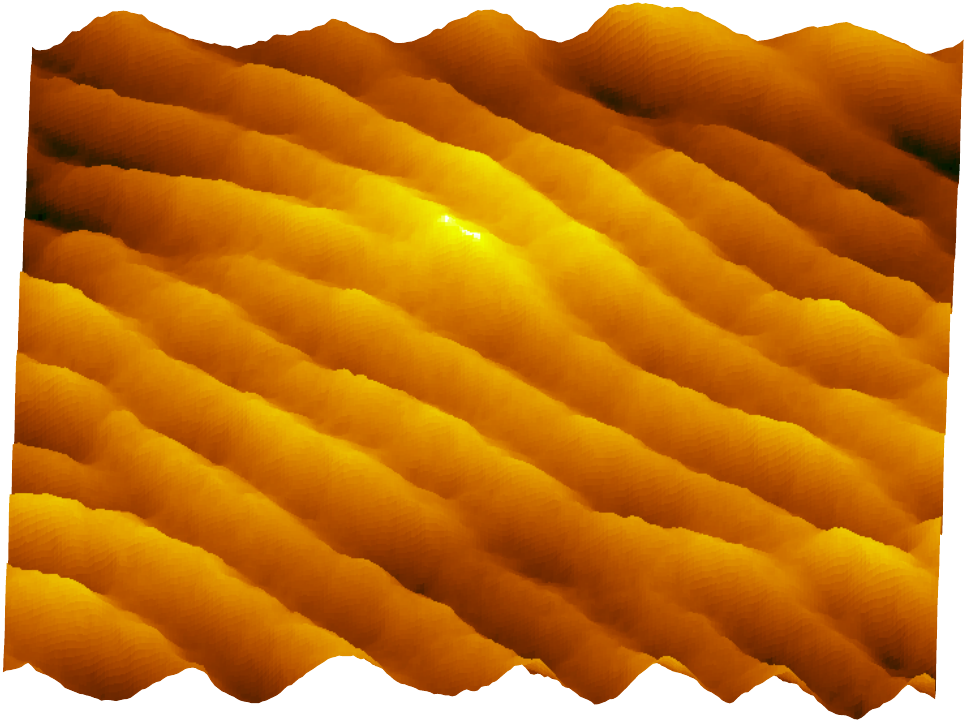}}
	\hfill
	\subfloat[UNet-opt\label{fig:cGANestimate1}]{%
		\includegraphics[trim={1cm 1cm 2cm 1cm},clip,width=0.17\linewidth,height=.05\textheight]{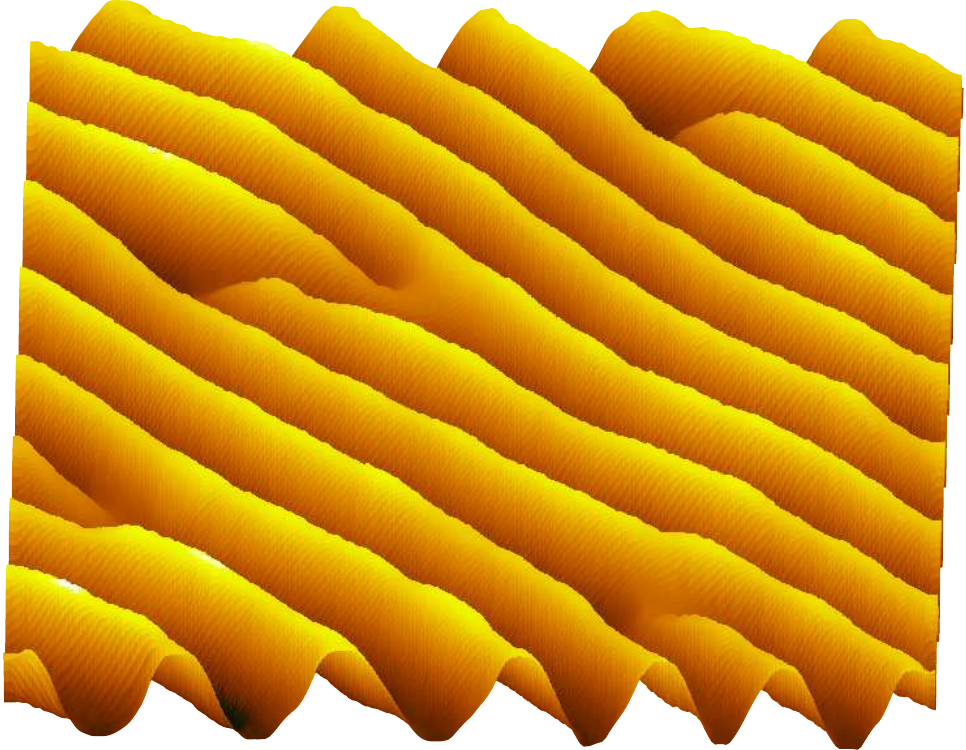}}\\
	
	\subfloat[$I_2$\label{fig:Intensity2}]{%
		\includegraphics[trim={1cm 1cm 1cm 1cm},clip,width=0.17\linewidth,height=.05\textheight]{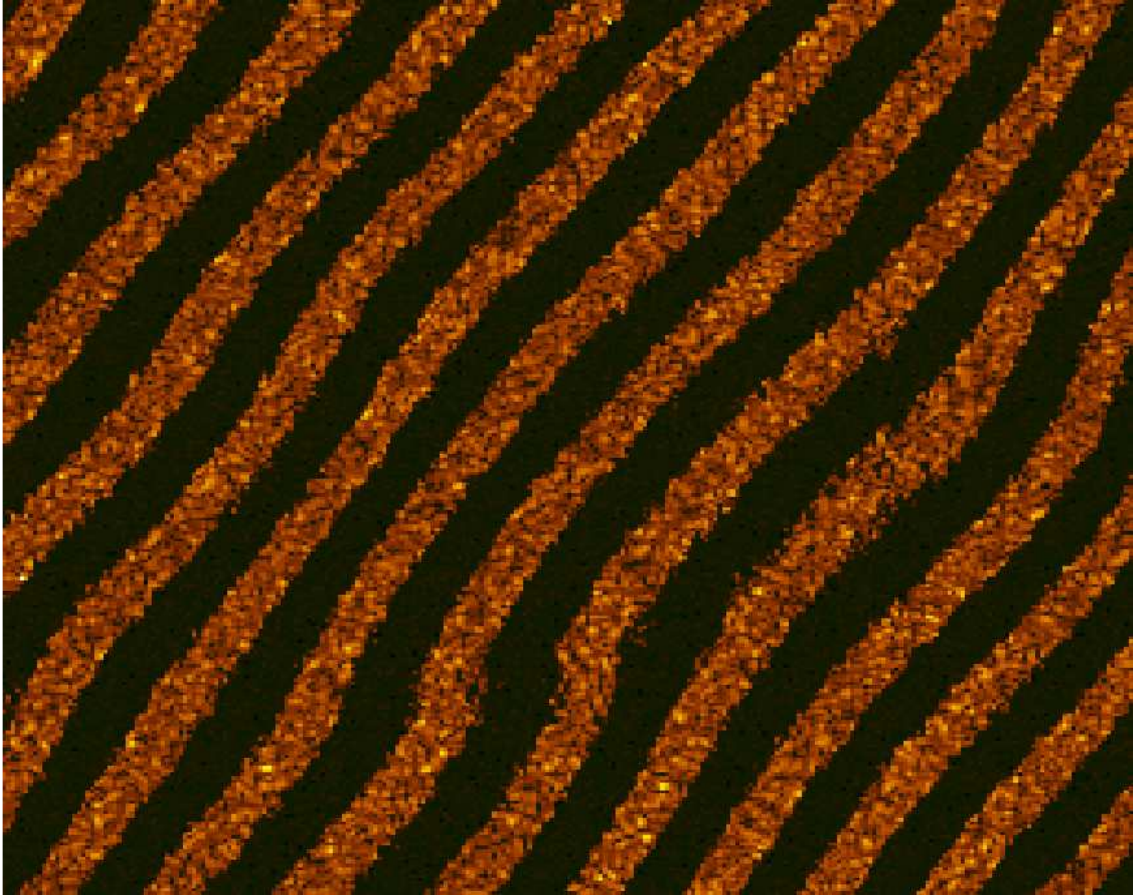}}
	\hfill
	\subfloat[$\mathcal{H}_2$\label{fig:Height2}]{%
		\includegraphics[trim={1cm 1cm 2cm 1cm},clip,width=0.17\linewidth,height=.05\textheight]{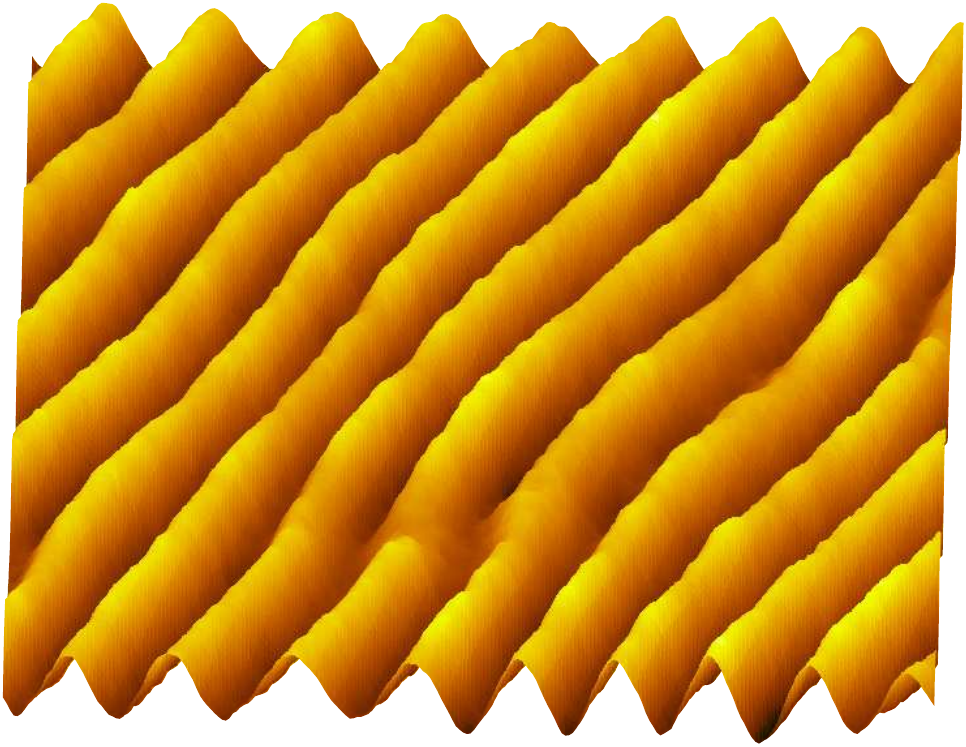}}
	\hfill
	\subfloat[GMRF\label{fig:GMRFestimate2}]{%
		\includegraphics[trim={1cm 1cm 2cm 1cm},clip,width=0.17\linewidth,height=.05\textheight]{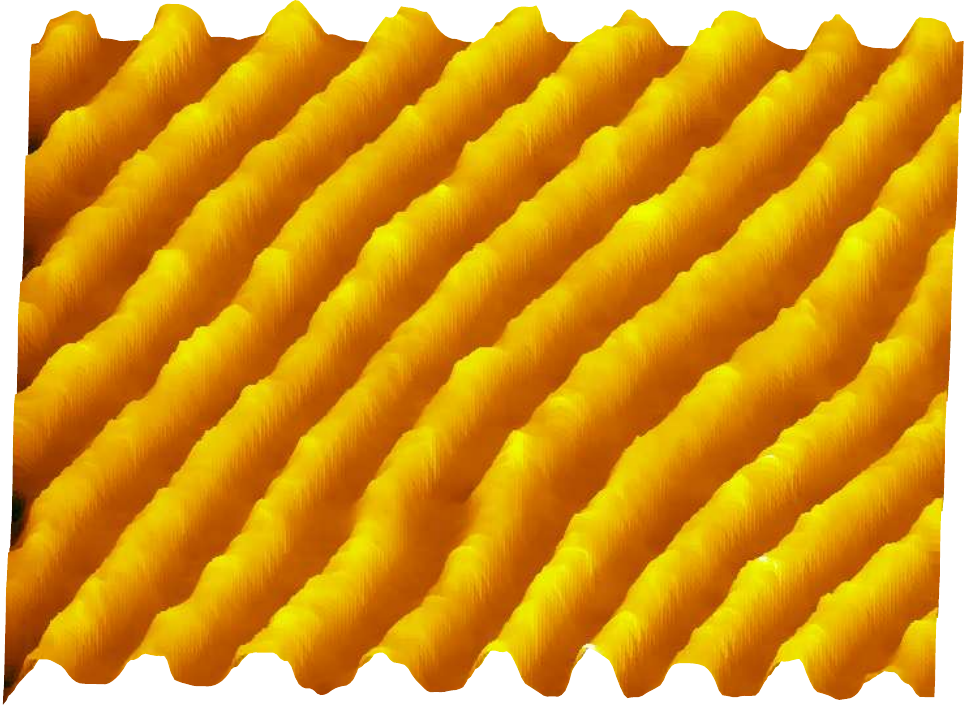}}
	\hfill
	\subfloat[pix2pix\label{fig:UNetestimate2}]{%
		\includegraphics[trim={1cm 1cm 2cm 1cm},clip,width=0.17\linewidth,height=.05\textheight]{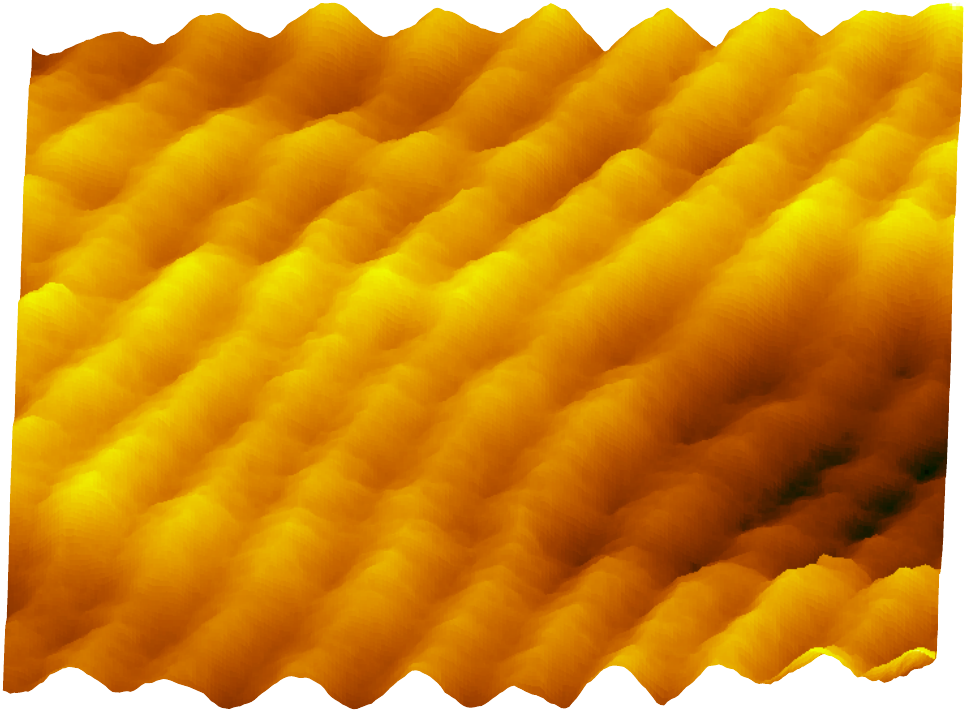}}
	\hfill
	\subfloat[UNet-opt\label{fig:cGANestimate2}]{%
		\includegraphics[trim={1cm 1cm 2cm 1cm},clip,width=0.17\linewidth,height=.05\textheight]{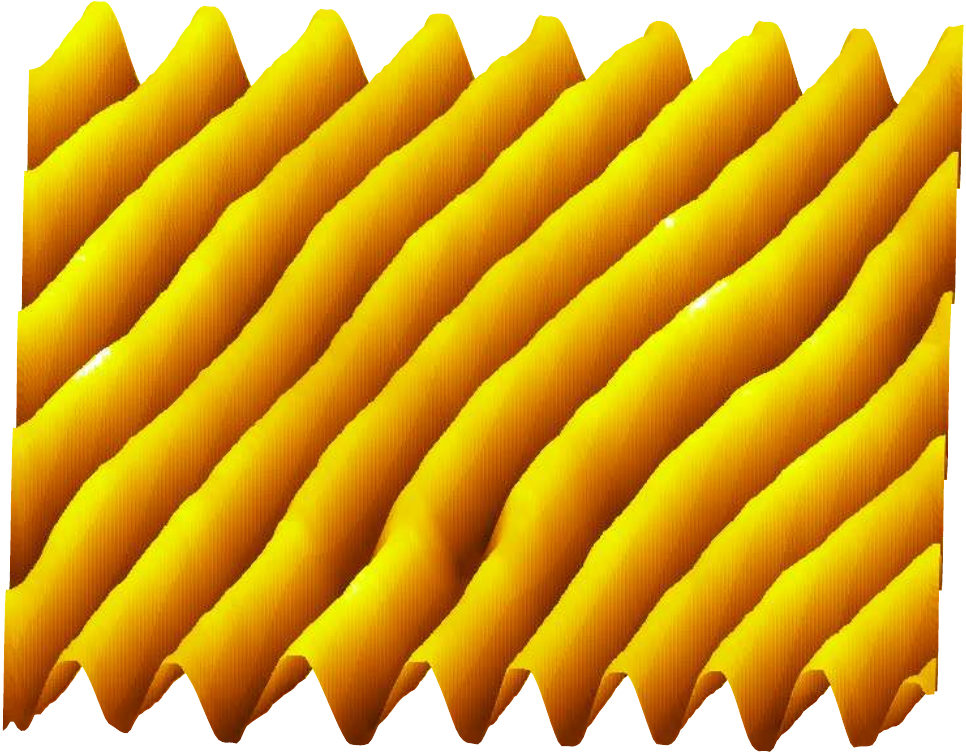}} \\
	
	\subfloat[$I_3$\label{fig:Intensity3}]{%
		\includegraphics[trim={1cm 1cm 1cm 1cm},clip,width=0.17\linewidth,height=.05\textheight]{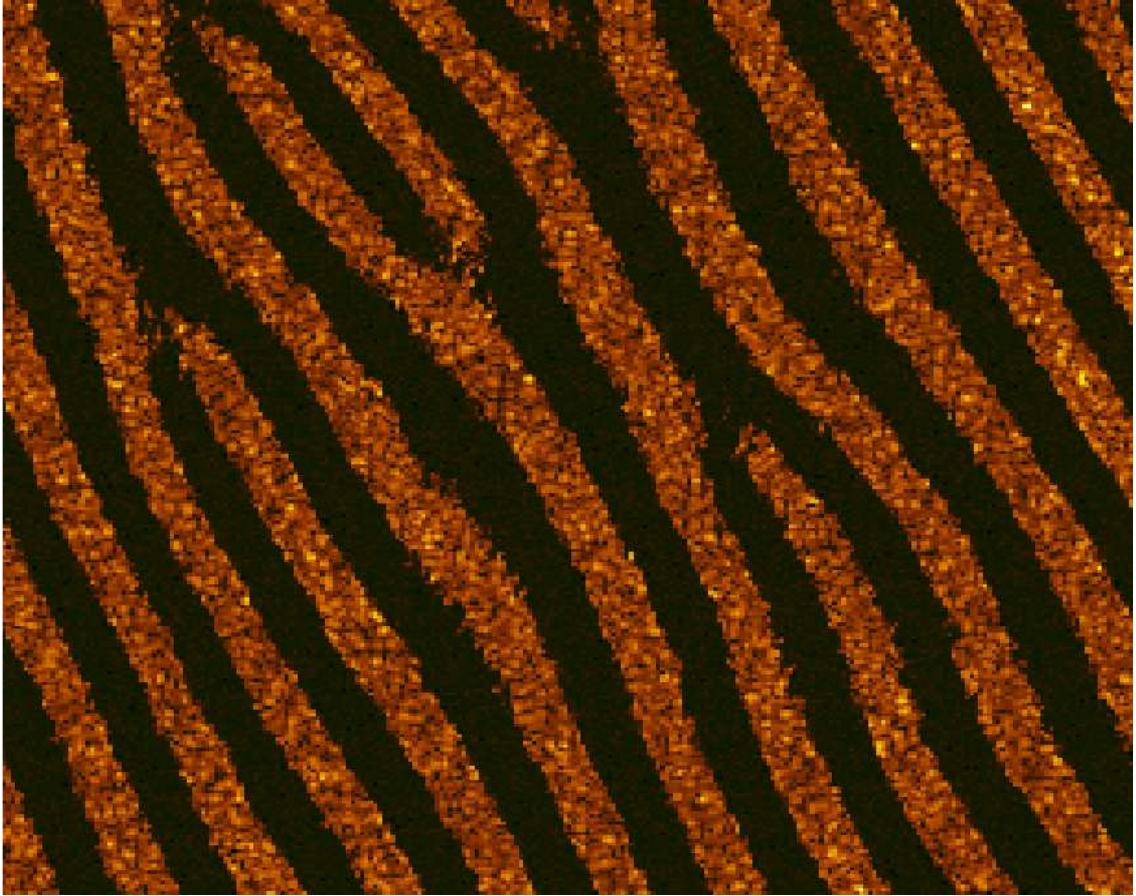}}
	\hfill
	\subfloat[$\mathcal{H}_3$\label{fig:Height3}]{%
		\includegraphics[trim={1cm 1cm 2cm 1cm},clip,width=0.17\linewidth,height=.05\textheight]{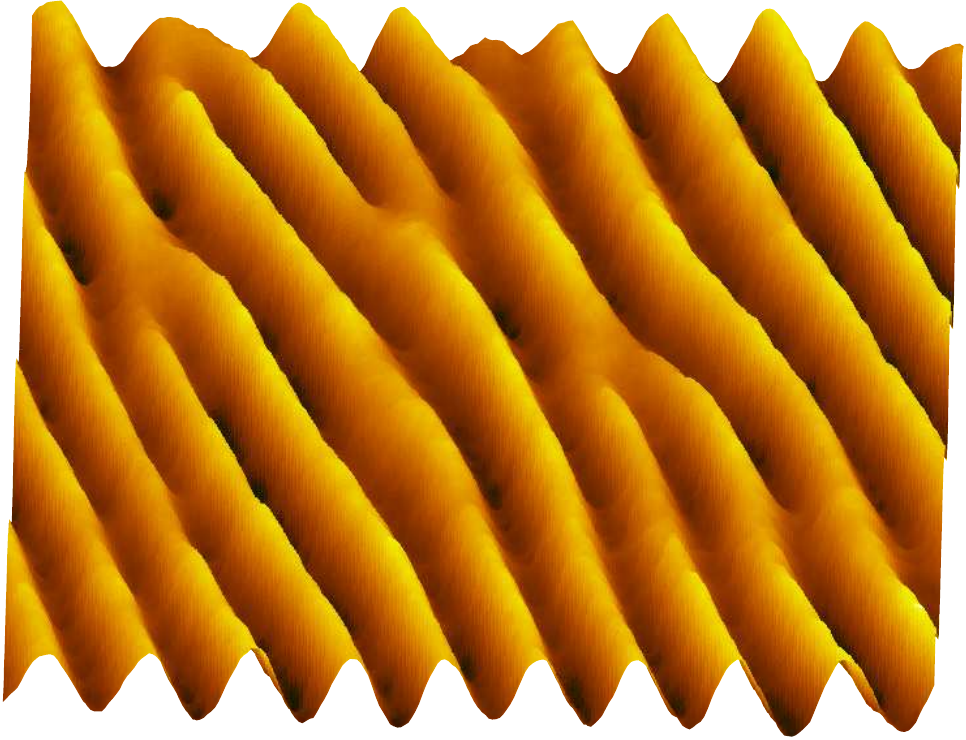}}
	\hfill
	\subfloat[GMRF\label{fig:GMRFestimate3}]{%
		\includegraphics[trim={1cm 1cm 2cm 1cm},clip,width=0.17\linewidth,height=.05\textheight]{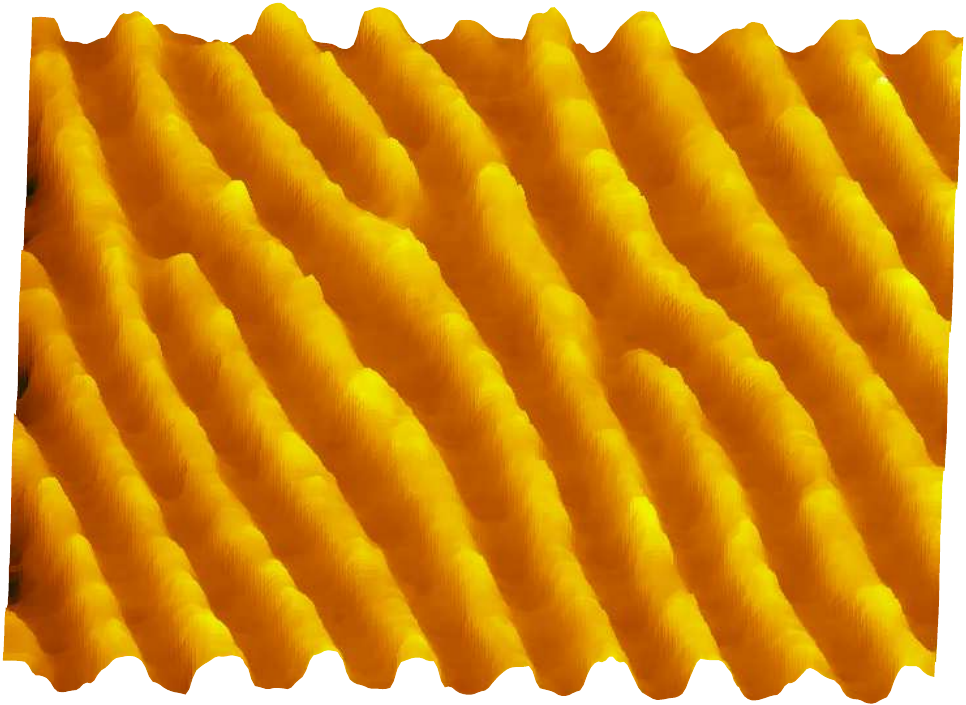}}
	\hfill
	\subfloat[pix2pix\label{fig:UNetestimate3}]{%
		\includegraphics[trim={1cm 1cm 2cm 1cm},clip,width=0.17\linewidth,height=.05\textheight]{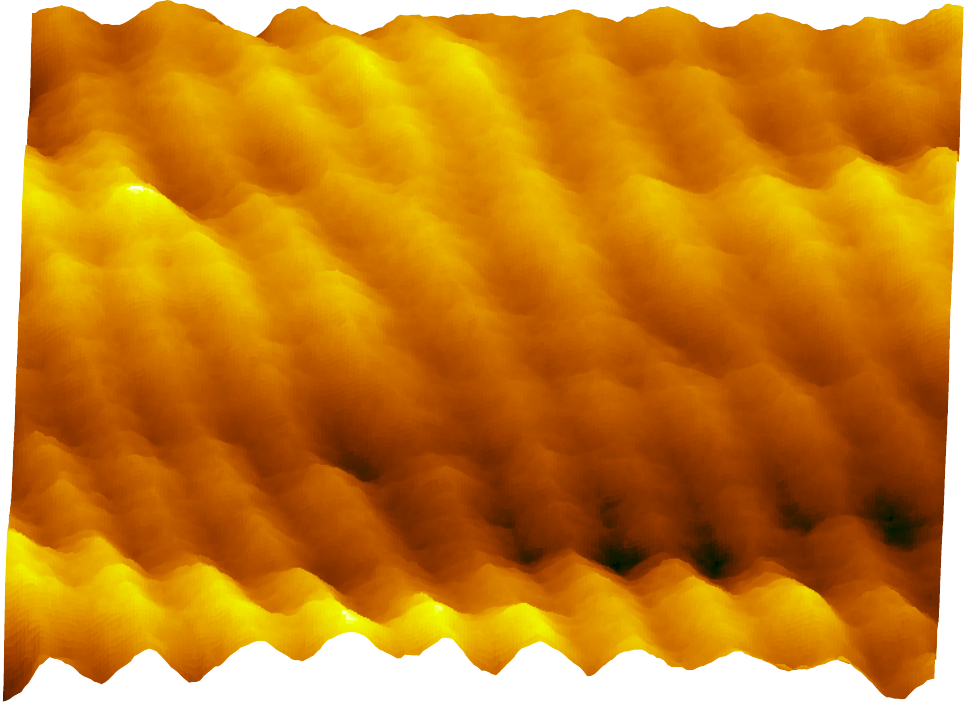}} 
	\hfill
	\subfloat[UNet-opt\label{fig:cGANestimate3}]{%
		\includegraphics[trim={1cm 1cm 2cm 1cm},clip,width=0.17\linewidth,height=.05\textheight]{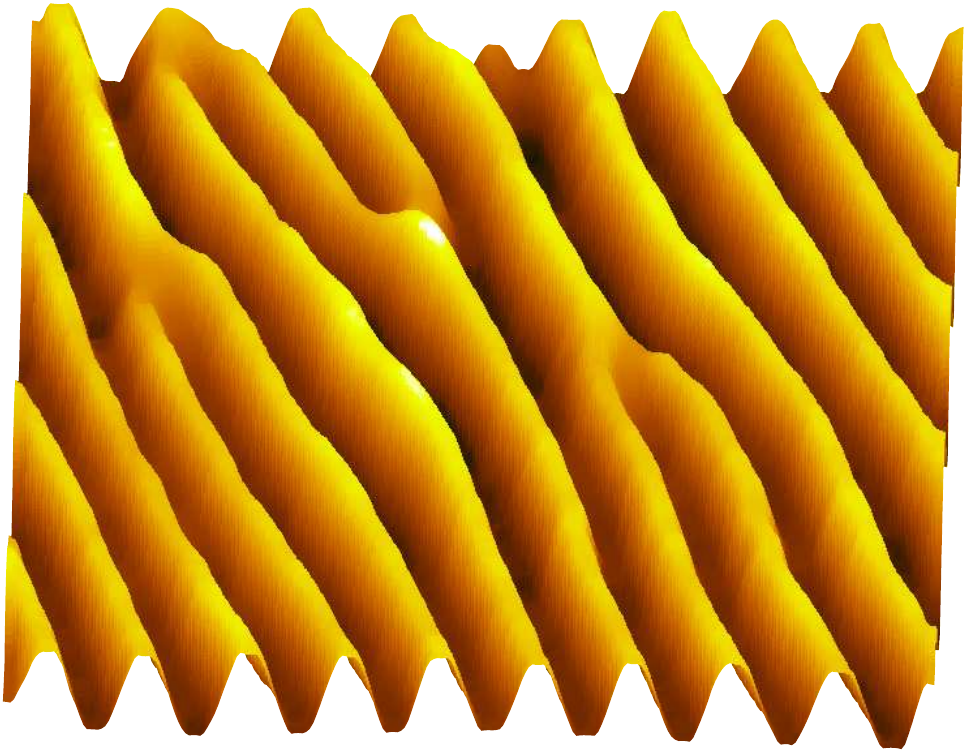}}\\
	\caption{Examples of seabed relief estimation on pure sand-ripple textures. Three intensity images (a,f,k) and their coregistered seabed relief maps (b,g,l) are shown. GMRF estimates (c,h,m) and UNet-opt estimates (e,j,o) do not flatten the seabed relief maps corresponding to sand-ripple textures like the pix2pix estimates (d,i,n) do. These examples show that although pix2pix is the most sophisticated model, it can oversmooth sand-ripple textures.}
	\label{fig:sand-ripples} 
\end{figure}

\begin{figure}[h!]
	\centering
	\subfloat[$I_1$\label{4a}]{%
		\includegraphics[trim={1cm 1cm 1cm 1cm},clip,width=0.17\linewidth,height=.05\textheight]{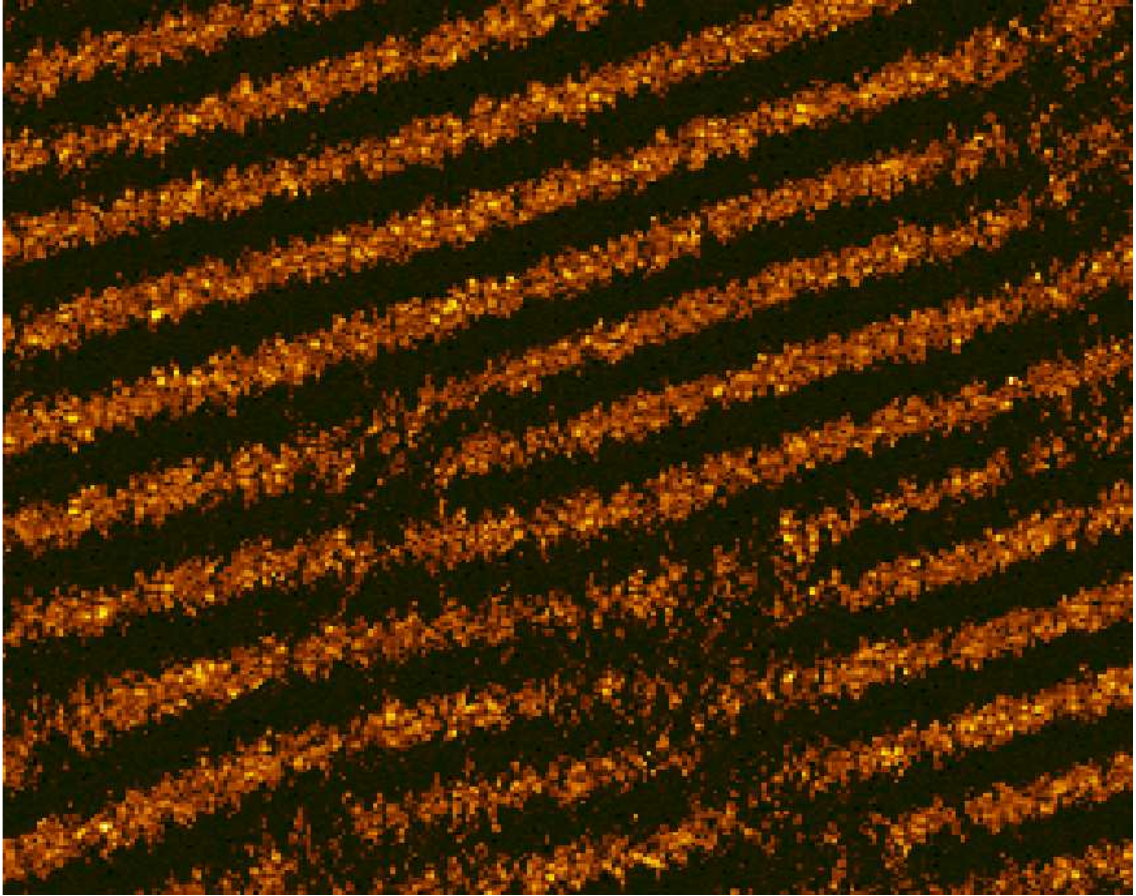}}
	\hfill
	\subfloat[$\mathcal{H}_1$\label{4b}]{%
		\includegraphics[trim={1cm 1cm 2cm 1cm},clip,width=0.17\linewidth,height=.05\textheight]{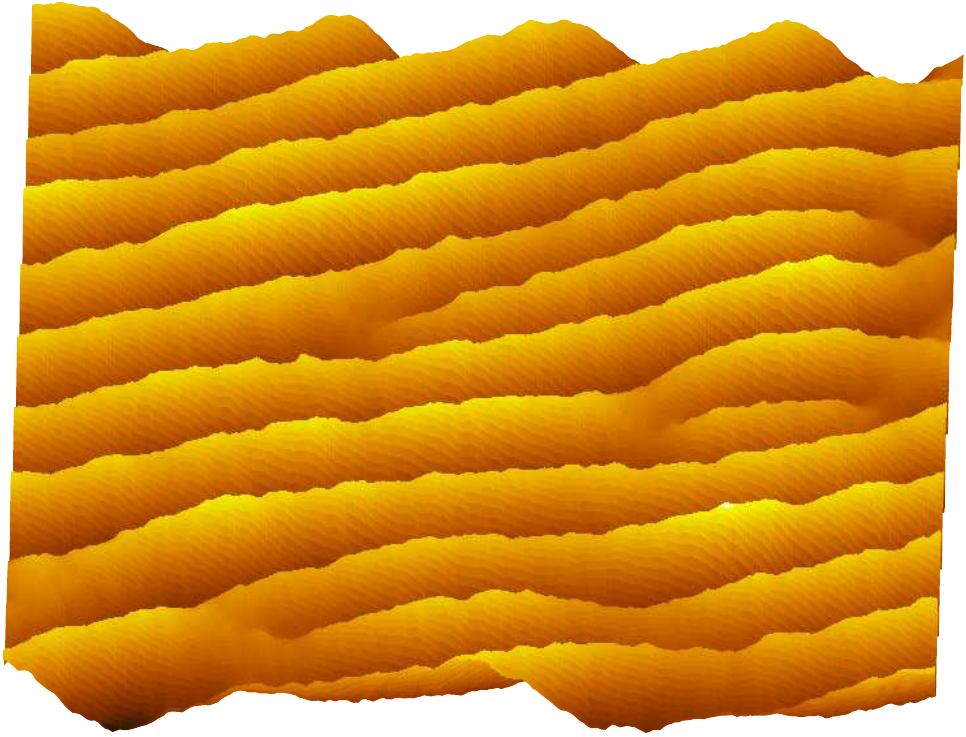}}
	\hfill
	\subfloat[GMRF\label{4c}]{%
		\includegraphics[trim={1cm 1cm 2cm 1cm},clip,width=0.17\linewidth,height=.05\textheight]{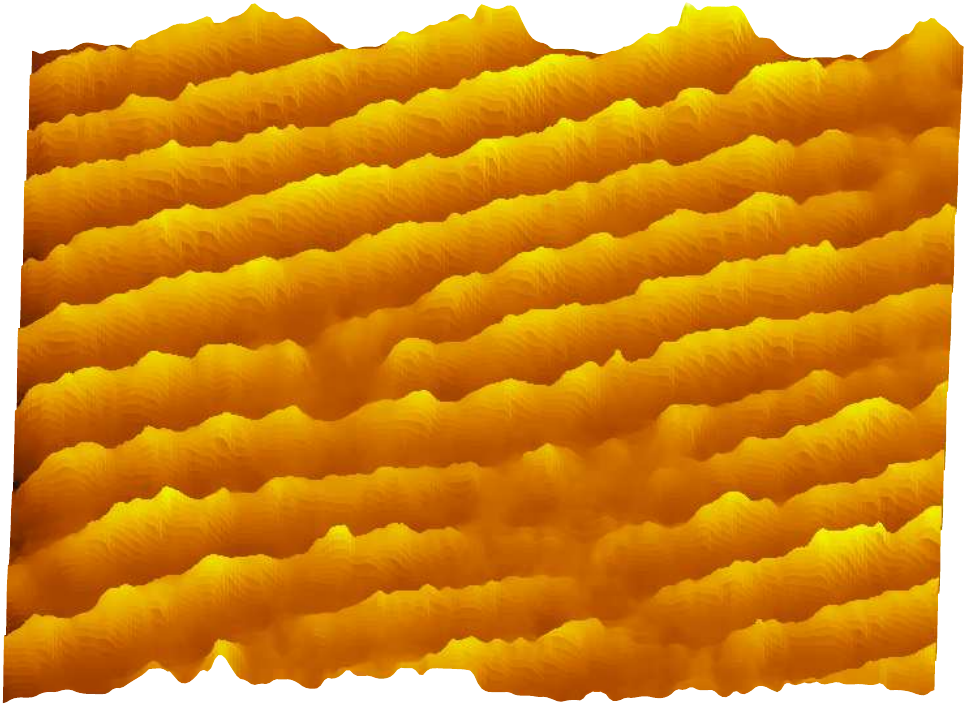}}
	\hfill
	\subfloat[pix2pix\label{4d}]{%
		\includegraphics[trim={1cm 1cm 2cm 1cm},clip,width=0.17\linewidth,height=.05\textheight]{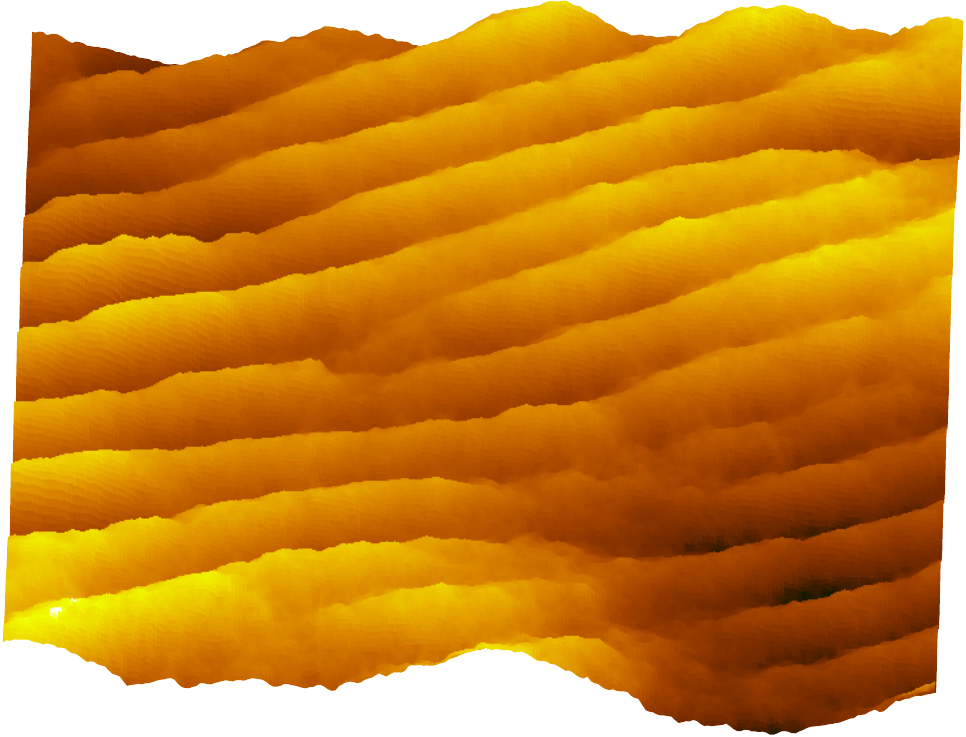}}  
	\hfill
	\subfloat[UNet-opt\label{4e}]{%
		\includegraphics[trim={1cm 1cm 2cm 1cm},clip,width=0.17\linewidth,height=.05\textheight]{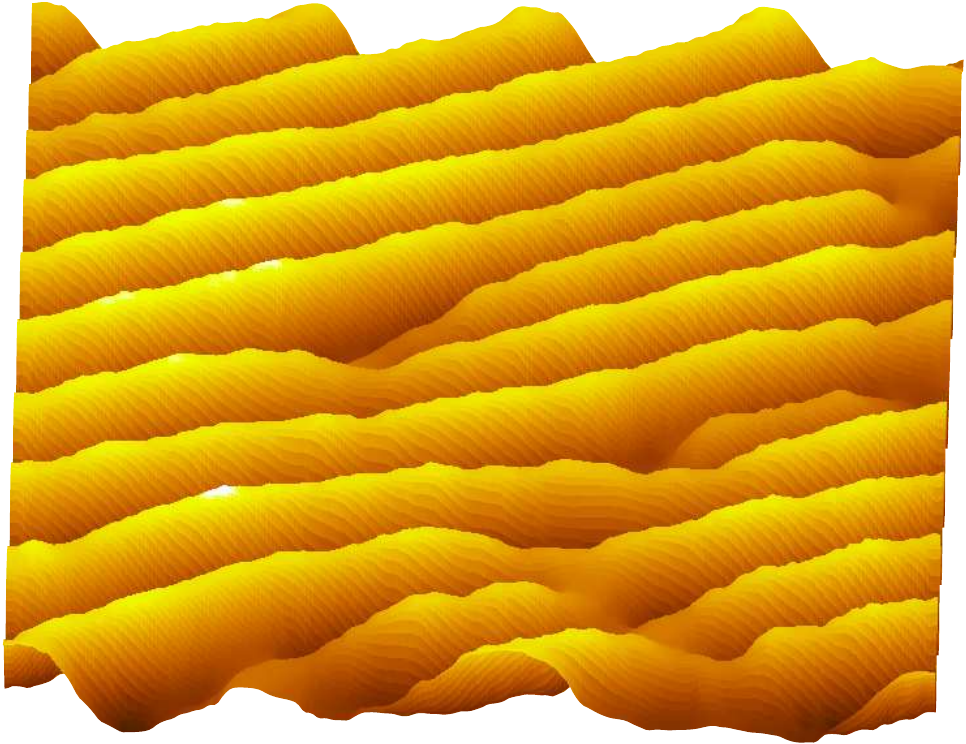}}\\
	
	\subfloat[$I_2$\label{5a}]{%
		\includegraphics[trim={1cm 1cm 1cm 1cm},clip,width=0.17\linewidth,height=.05\textheight]{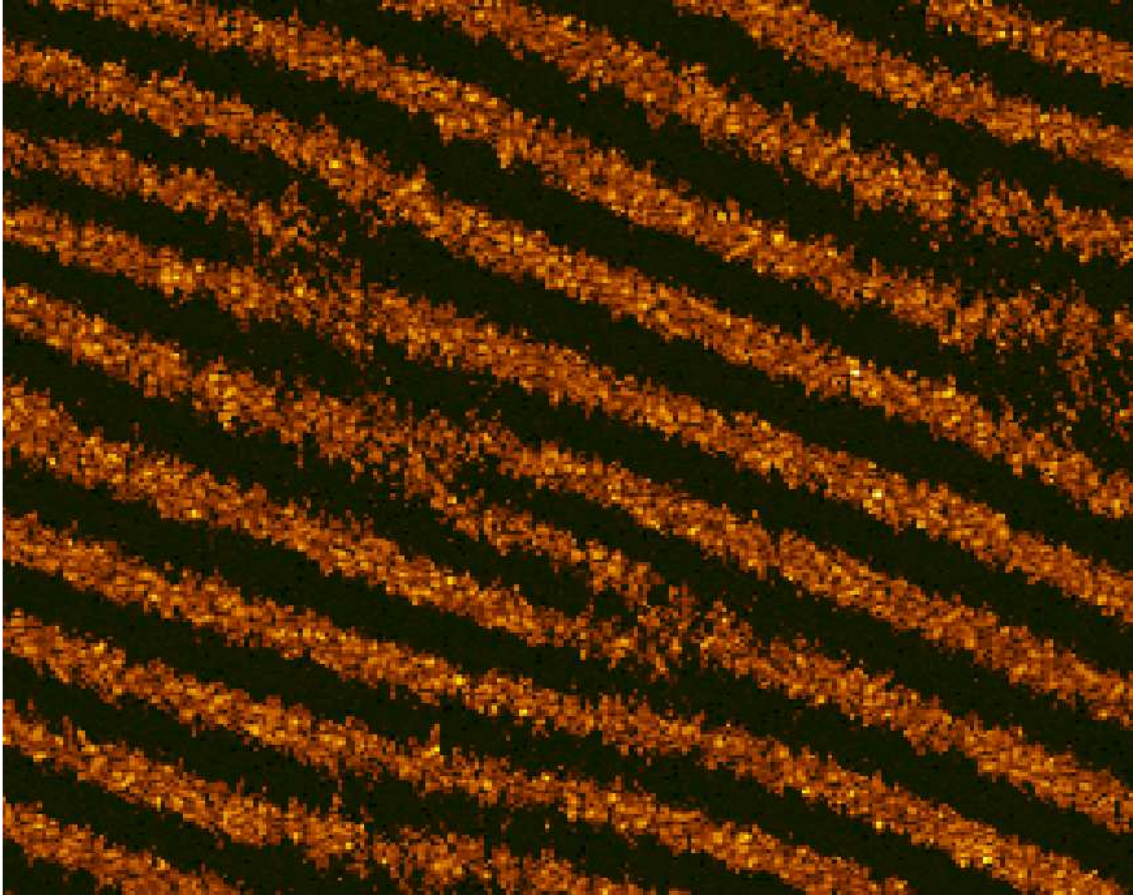}}
	\hfill
	\subfloat[$\mathcal{H}_2$\label{5b}]{%
		\includegraphics[trim={1cm 1cm 2cm 1cm},clip,width=0.17\linewidth,height=.05\textheight]{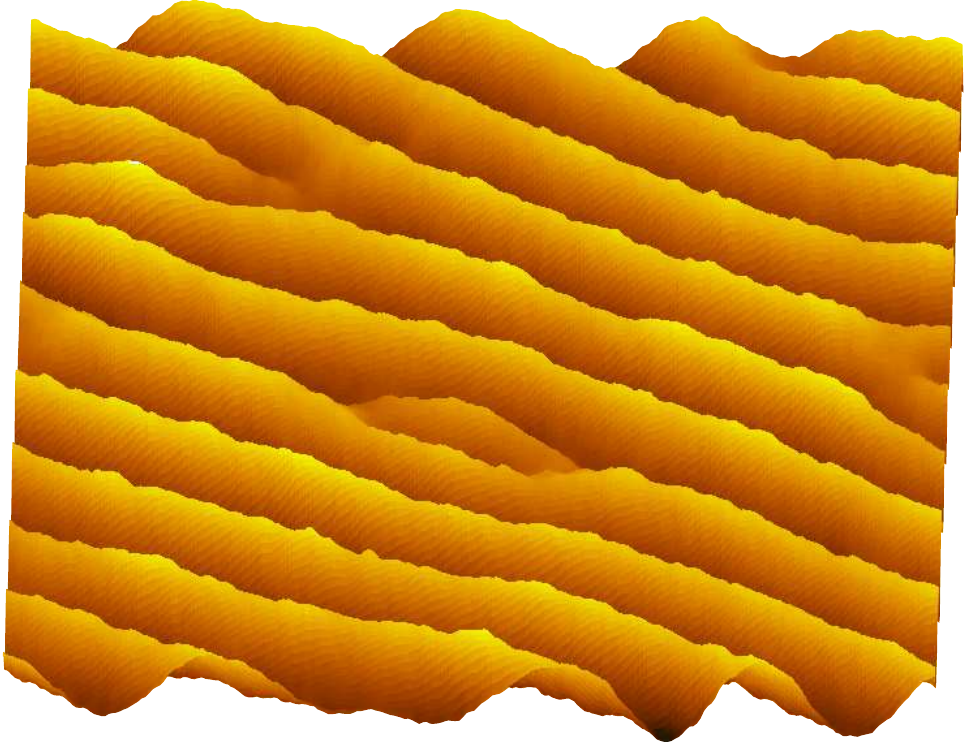}}
	\hfill
	\subfloat[GMRF\label{5c}]{%
		\includegraphics[trim={1cm 1cm 2cm 1cm},clip,width=0.17\linewidth,height=.05\textheight]{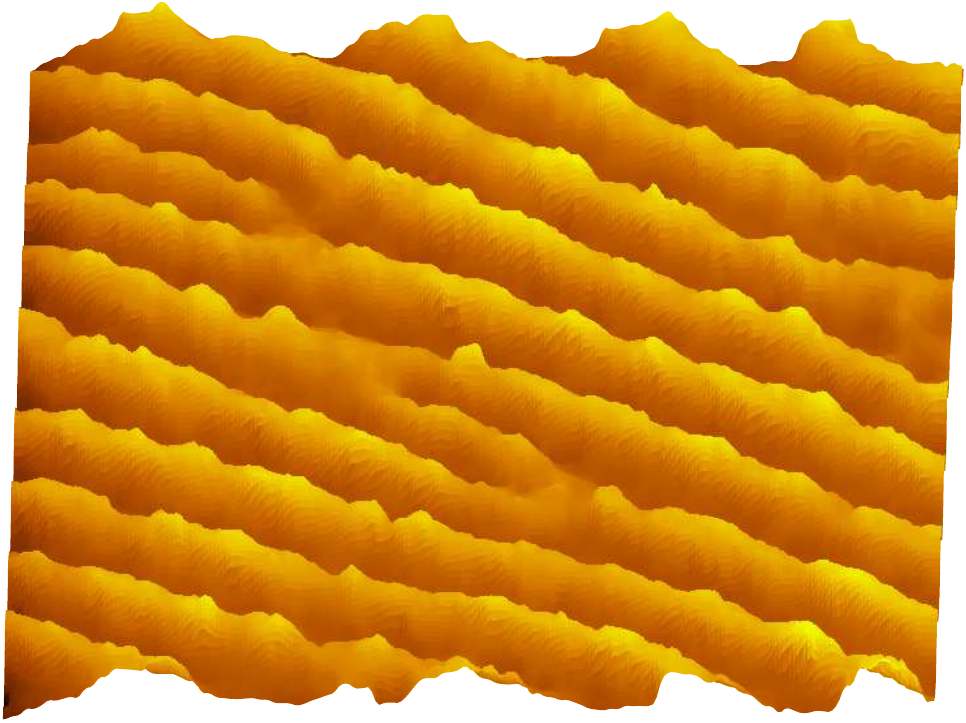}}
	\hfill
	\subfloat[pix2pix\label{5d}]{%
		\includegraphics[trim={1cm 1cm 2cm 1cm},clip,width=0.17\linewidth,height=.05\textheight]{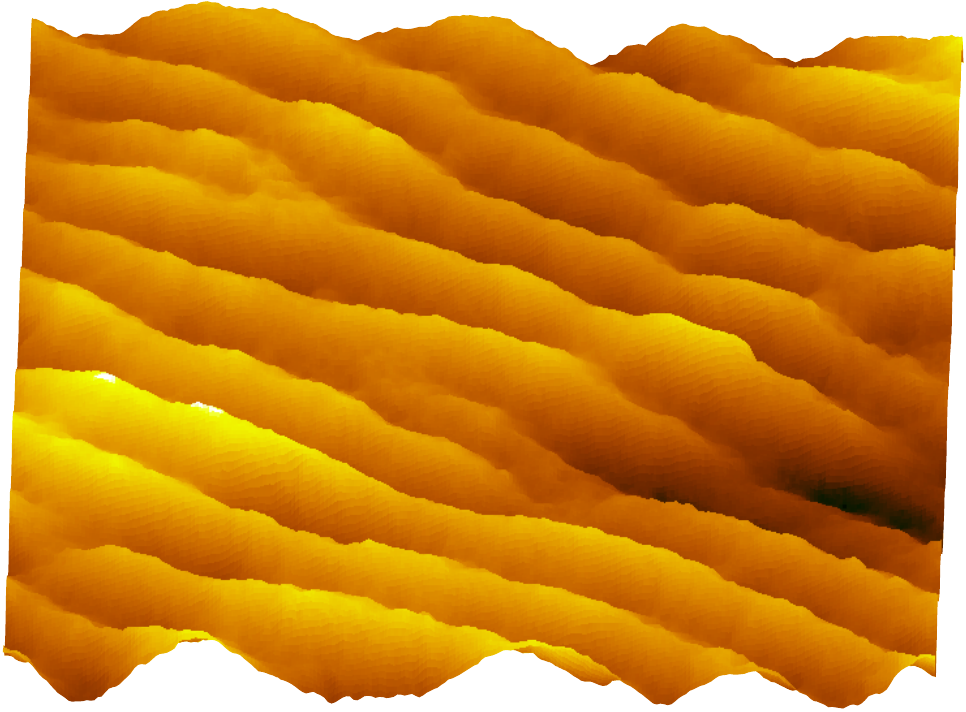}} 
	\hfill
	\subfloat[UNet-opt\label{5e}]{%
		\includegraphics[trim={1cm 1cm 2cm 1cm},clip,width=0.17\linewidth,height=.05\textheight]{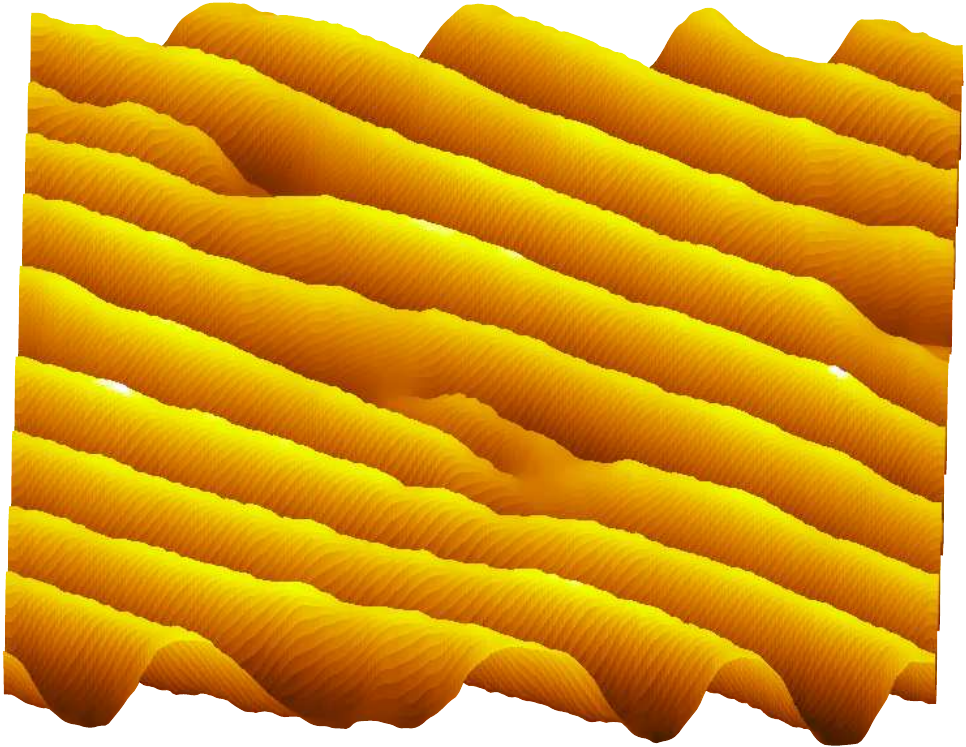}}\\
	
	\subfloat[$I_3$\label{6a}]{%
		\includegraphics[trim={1cm 1cm 1cm 1cm},clip,width=0.17\linewidth,height=.05\textheight]{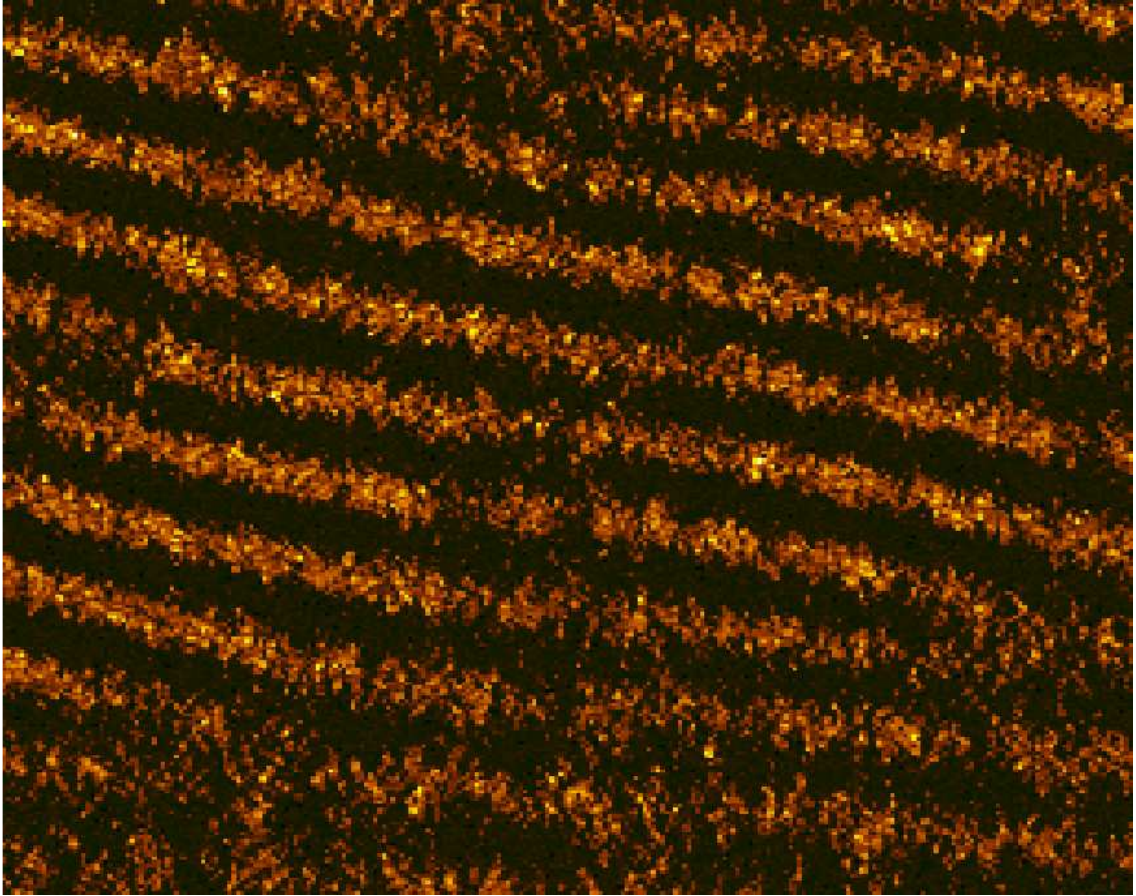}}
	\hfill
	\subfloat[$\mathcal{H}_3$\label{6b}]{%
		\includegraphics[trim={1cm 1cm 2cm 1cm},clip,width=0.17\linewidth,height=.05\textheight]{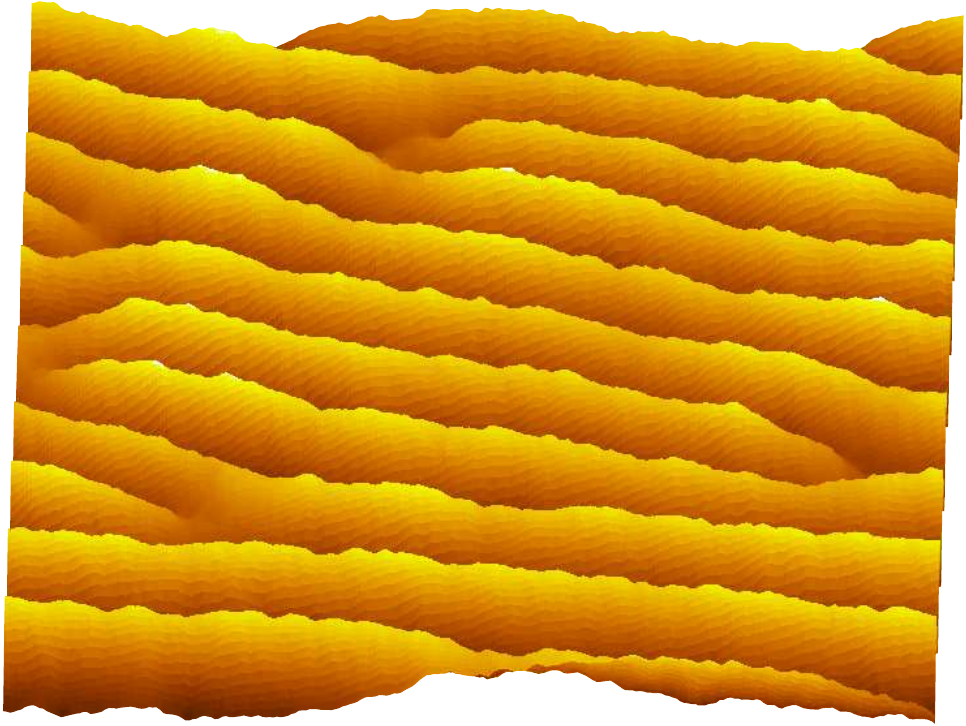}}
	\hfill
	\subfloat[GMRF\label{6c}]{%
		\includegraphics[trim={1cm 1cm 2cm 1cm},clip,width=0.17\linewidth,height=.05\textheight]{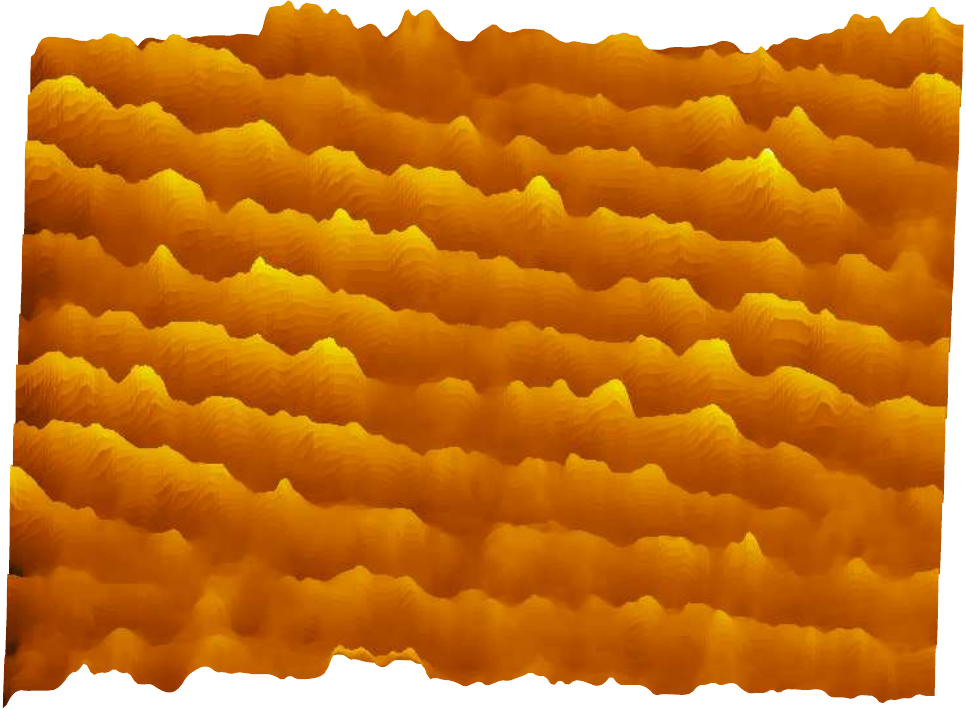}}
	\hfill
	\subfloat[pix2pix\label{6d}]{%
		\includegraphics[trim={1cm 1cm 2cm 1cm},clip,width=0.17\linewidth,height=.05\textheight]{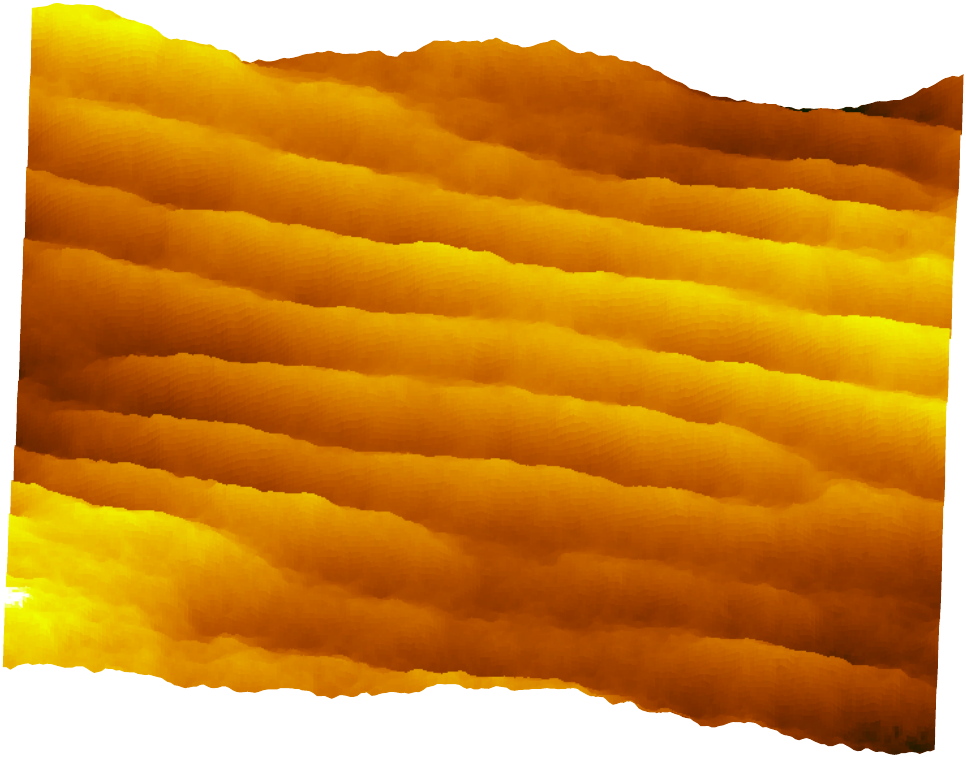}} 
	\hfill
	\subfloat[UNet-opt\label{6e}]{%
		\includegraphics[trim={1cm 1cm 2cm 1cm},clip,width=0.17\linewidth,height=.05\textheight]{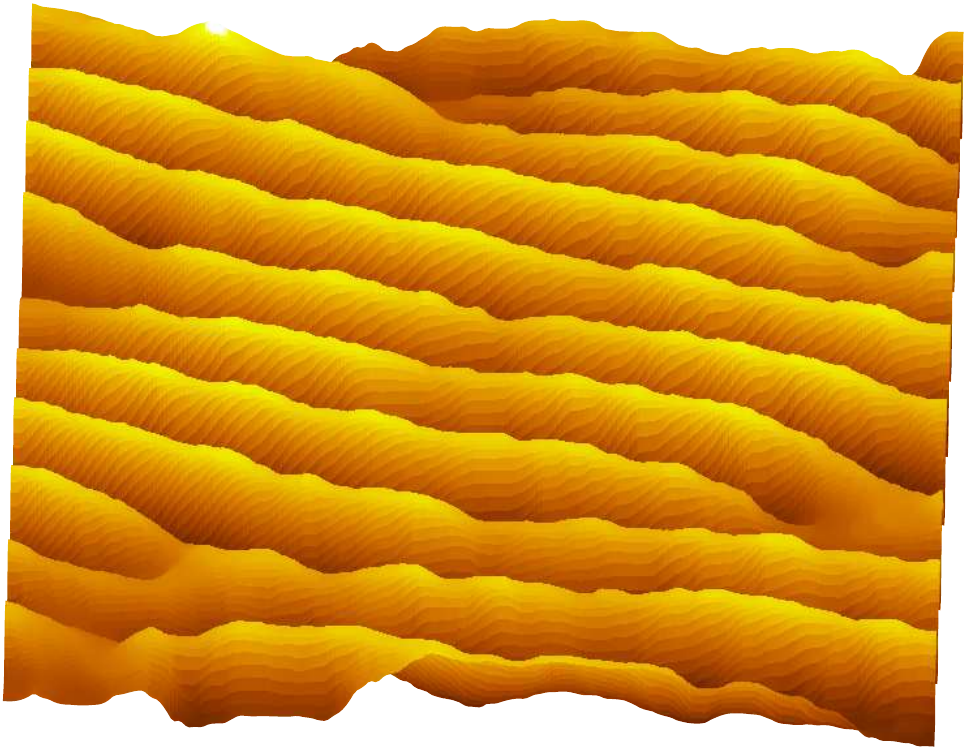}} \\
	\caption{Examples of seabed relief estimation on textures containing a mixture of small sand-ripple and roughness. Three intensity images (a,f,k) and their coregistered seabed relief maps (b,g,l) are shown. The pix2pix model (e,j,o) oversmoothes while the GMRF (c,h,m) and UNet-opt (d,i,n) models appear to be visually closer to the desired seabed relief in each of the three examples.}
	\label{fig:small-mix}
\end{figure}

We compare the $L_1$ error between the predicted and desired seabed relief to compare performance on the PISAS dataset (see Table \ref{tab:testPISAS}). While the GMRF model requires few parameters compared to each of the UNet architectures (Table \ref{tab:UNetModels}) and the pix2pix model, it outperforms the most complex network on the PISAS dataset. Upon qualitative inspection of the highest error generated samples from the cGAN, we noted some of the sand-ripple seabed relief maps were oversmoothed by the generator as shown in \Cref{fig:sand-ripples,fig:small-mix}.

	\begin{table}[htb]
		\centering
		\caption{\label{tab:testPISAS} $L_1$ error between predicted and desired seabed relief for each model applied to the PISAS test set. Channel-reduced (c) and depth-reduced (d) have been shortened for conciseness. Models with fewer parameters (GMRF and UNet variants) outperform the pix2pix network. The best model is the UNet-opt.}
\begin{tabular}{cccc}
\multicolumn{4}{c}{PISAS Test Error}                                                                \\
\rowcolor[HTML]{FFFFFF} 
Model   & \multicolumn{1}{c|}{\cellcolor[HTML]{FFFFFF}Error}           & Model    & Error           \\ \hline
\rowcolor[HTML]{E0E0E0} 
GMRF    & \multicolumn{1}{c|}{\cellcolor[HTML]{E0E0E0}$0.17\pm0.02$} & UNet     & $0.14\pm0.04$ \\
\rowcolor[HTML]{FFFFFF} 
pix2pix & \multicolumn{1}{c|}{\cellcolor[HTML]{FFFFFF}$0.19\pm0.06$} & UNet-opt & $\bm{0.13\pm0.02}$ \\
\rowcolor[HTML]{E0E0E0} 
UNet-c2 & \multicolumn{1}{c|}{\cellcolor[HTML]{E0E0E0}$0.14\pm0.02$} & UNet-d   & $0.14\pm0.02$ \\
\rowcolor[HTML]{FFFFFF} 
UNet-d2 & \multicolumn{1}{c|}{\cellcolor[HTML]{FFFFFF}$0.14\pm0.02$} & UNet-d3  & $0.14\pm0.02$
\end{tabular}
	\end{table}

The sand-ripple textures (Figure \ref{fig:sand-ripples}) are some of the least complex seabed relief maps that are present in the dataset. On these samples, the GMRF model produces seabed relief maps that qualitatively match the desired seabed relief. However, each contains slight bumps on the peaks of the ripples. Both the UNet-opt and pix2pix models produce smooth ripples. Unlike the UNet-opt, the pix2pix model oversmoothes the ripples. 

\begin{figure}[h!]
	\centering
	\subfloat[$I_1$\label{fig:Intensity4}]{%
		\includegraphics[trim={1cm 1cm 1cm 1cm},clip,width=0.17\linewidth,height=.05\textheight]{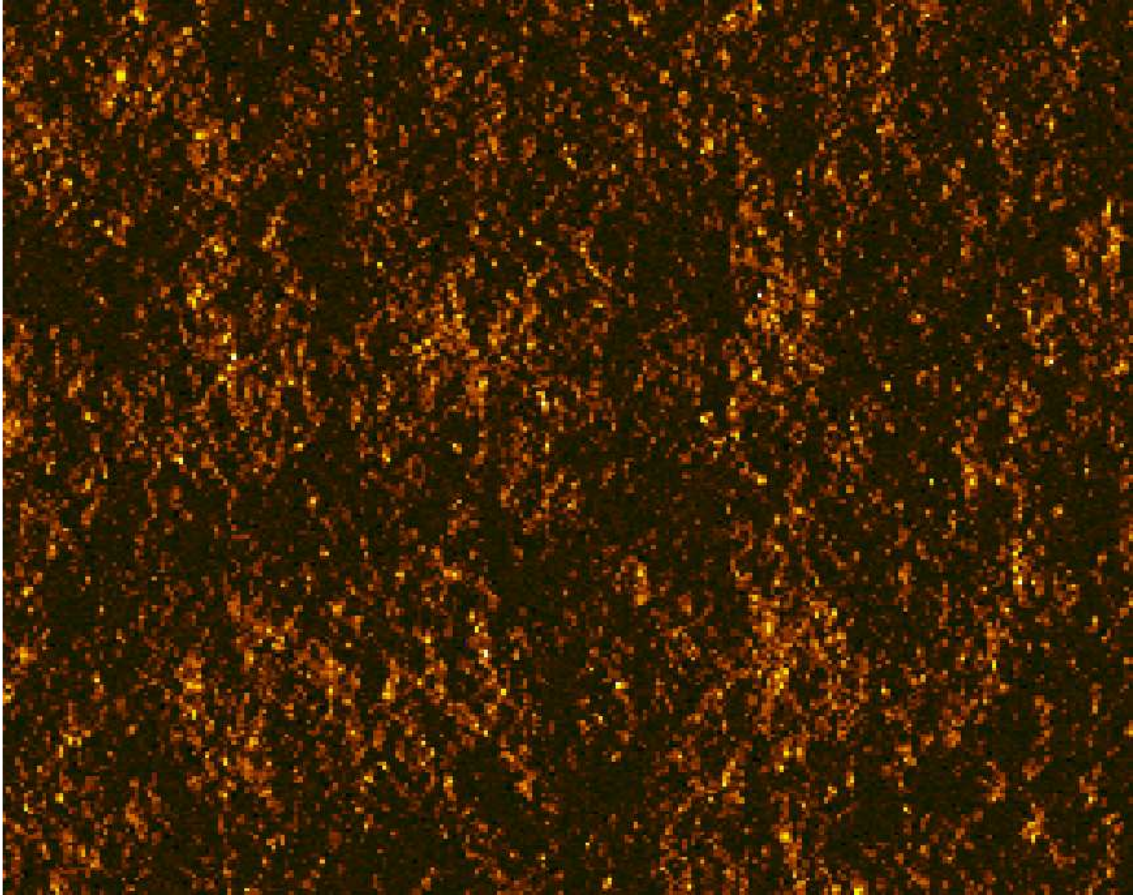}}
	\hfill
	\subfloat[$\mathcal{H}_1$\label{fig:Height4}]{%
		\includegraphics[trim={1cm 1cm 2cm 1cm},clip,width=0.17\linewidth,height=.05\textheight]{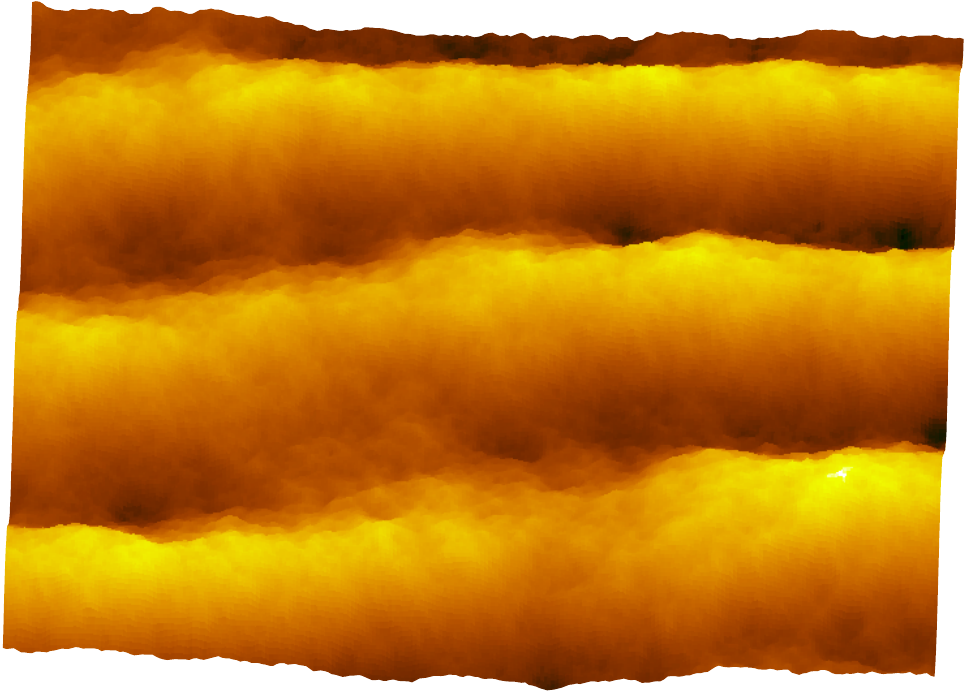}}
	\hfill
	\subfloat[GMRF\label{fig:GMRFestimate4}]{%
		\includegraphics[trim={1cm 1cm 2cm 1cm},clip,width=0.17\linewidth,height=.05\textheight]{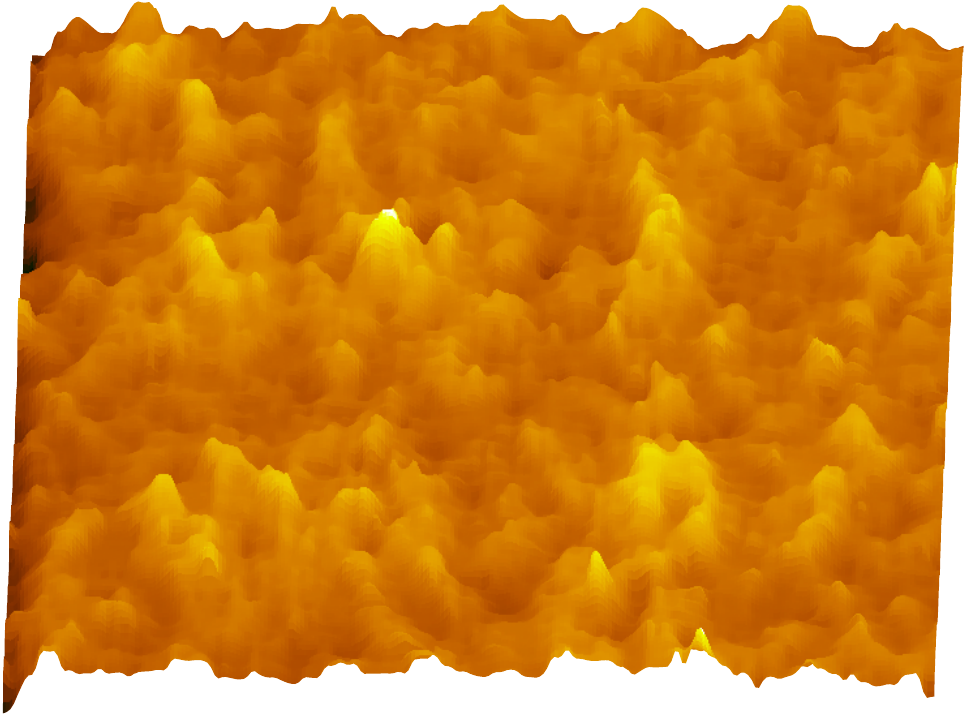}}
	\hfill
	\subfloat[pix2pix\label{fig:UNetestimate4}]{%
		\includegraphics[trim={1cm 1cm 2cm 1cm},clip,width=0.17\linewidth,height=.05\textheight]{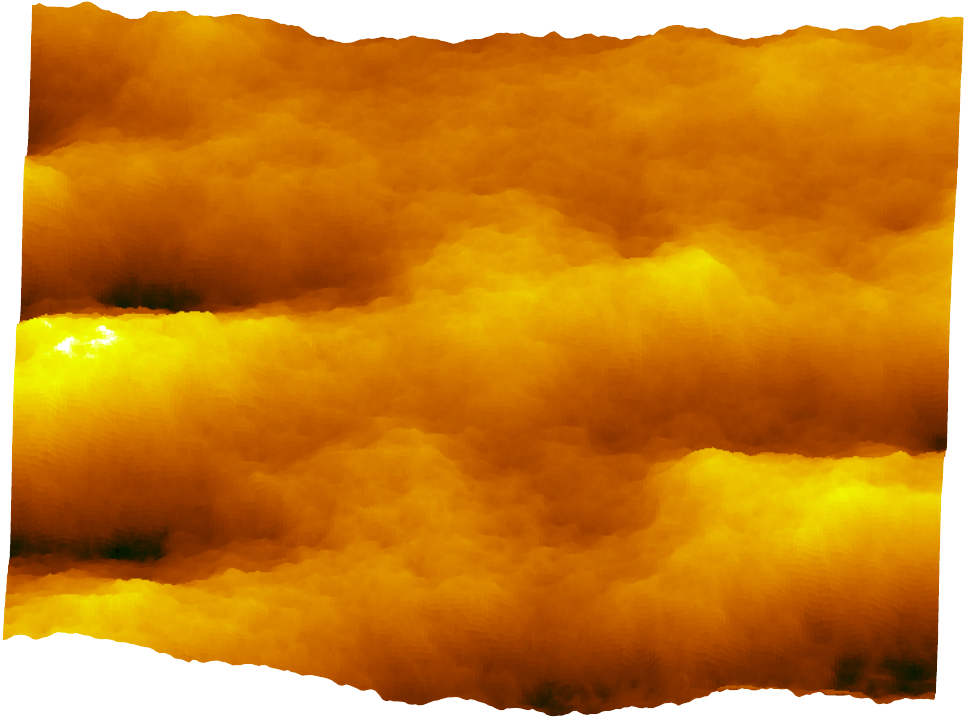}}
	\hfill
	\subfloat[UNet-opt\label{fig:cGANestimate4}]{%
		\includegraphics[trim={1cm 1cm 2cm 1cm},clip,width=0.17\linewidth,height=.05\textheight]{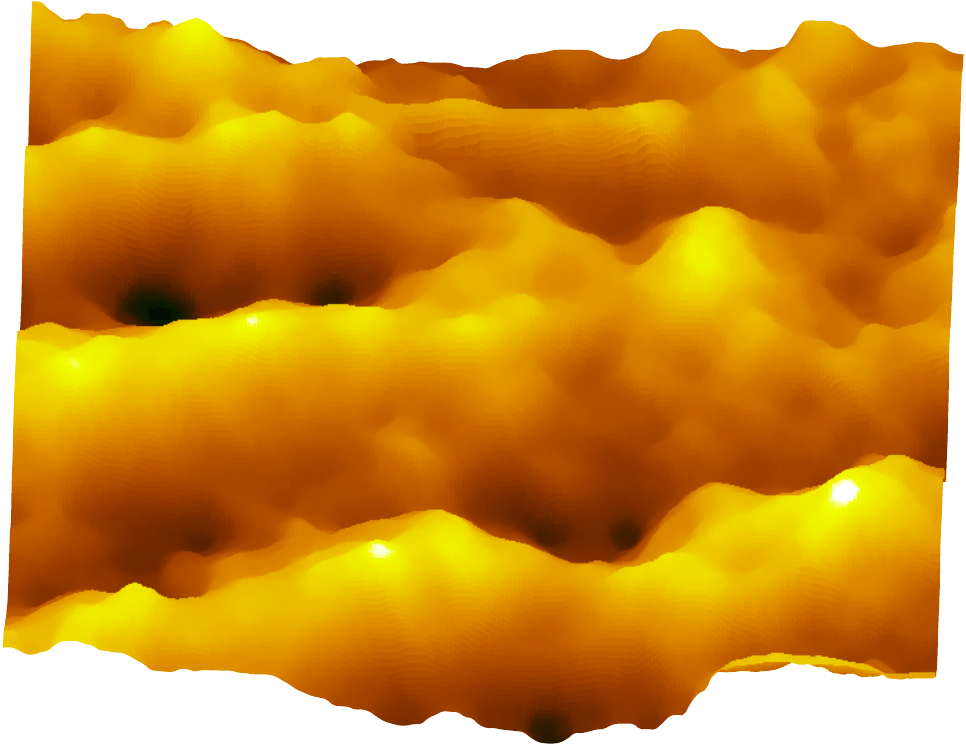}}\\
	
	\subfloat[$I_2$\label{fig:Intensity5}]{%
		\includegraphics[trim={1cm 1cm 1cm 1cm},clip,width=0.17\linewidth,height=.05\textheight]{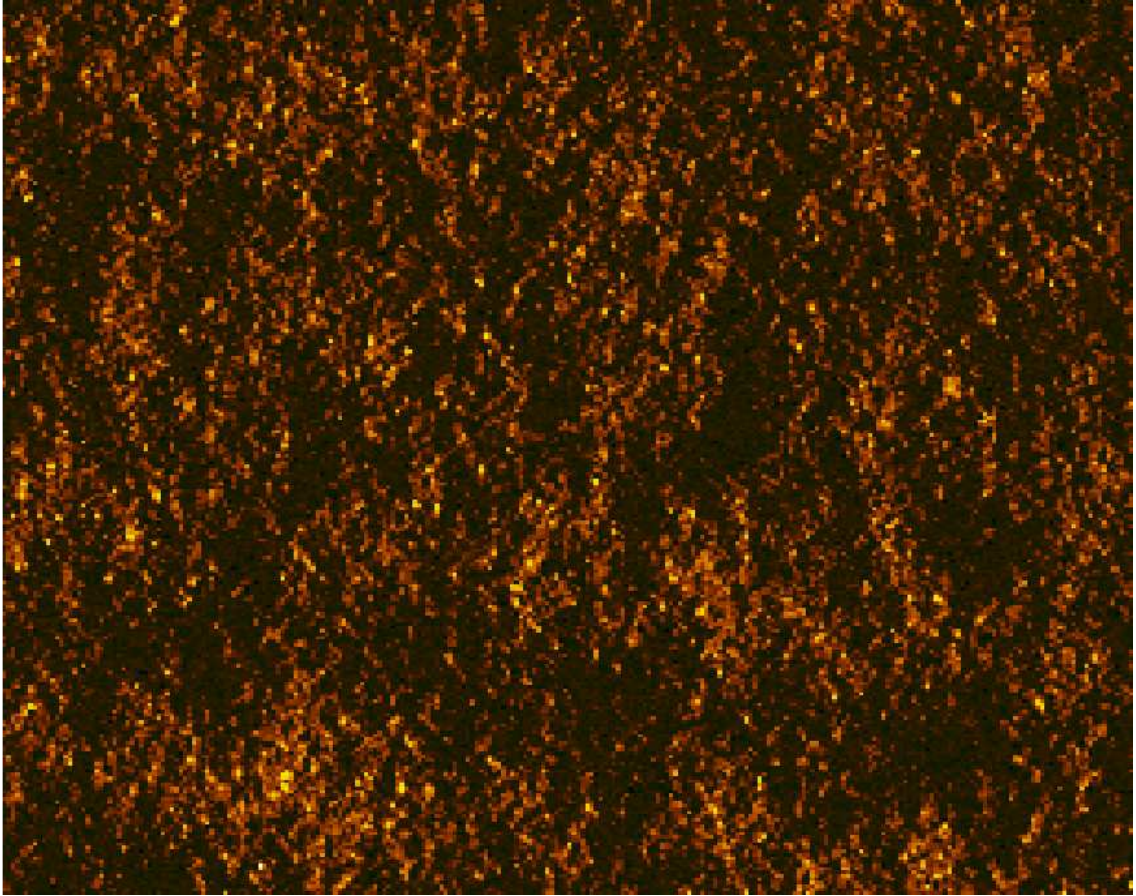}}
	\hfill
	\subfloat[$\mathcal{H}_2$\label{fig:Height5}]{%
		\includegraphics[trim={1cm 1cm 2cm 1cm},clip,width=0.17\linewidth,height=.05\textheight]{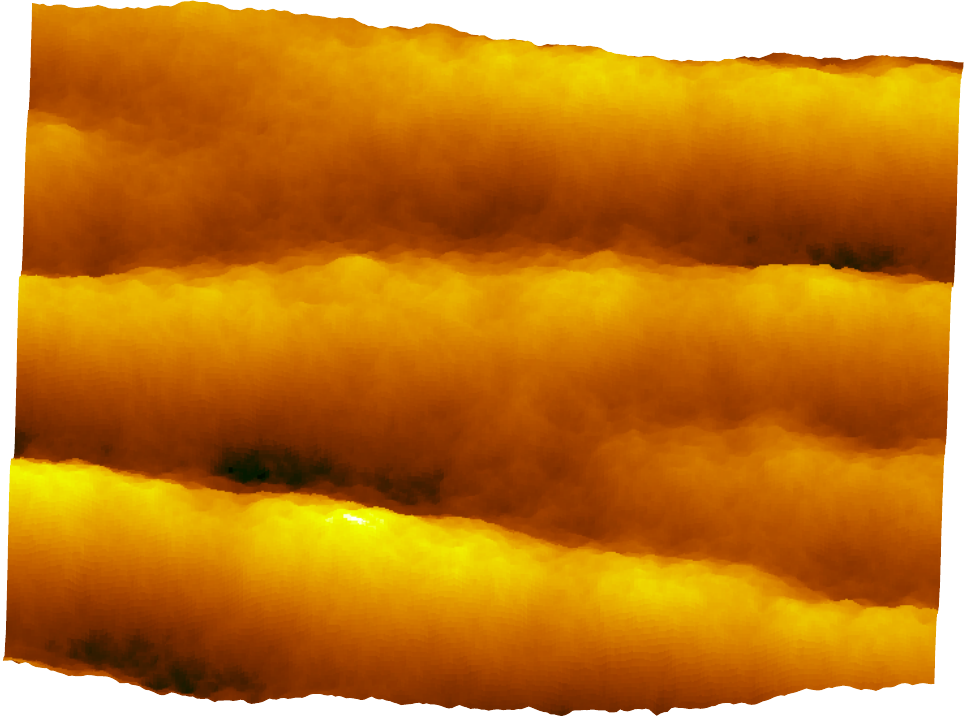}}
	\hfill
	\subfloat[GMRF\label{fig:GMRFestimate5}]{%
		\includegraphics[trim={1cm 1cm 2cm 1cm},clip,width=0.17\linewidth,height=.05\textheight]{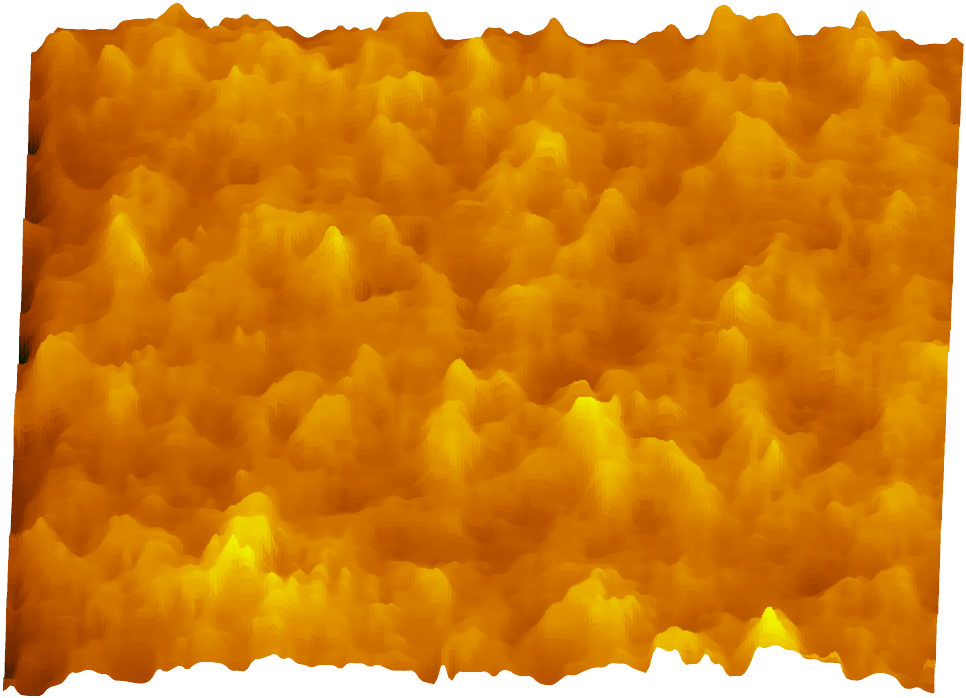}}
	\hfill
	\subfloat[pix2pix\label{fig:UNetestimate5}]{%
		\includegraphics[trim={1cm 1cm 2cm 1cm},clip,width=0.17\linewidth,height=.05\textheight]{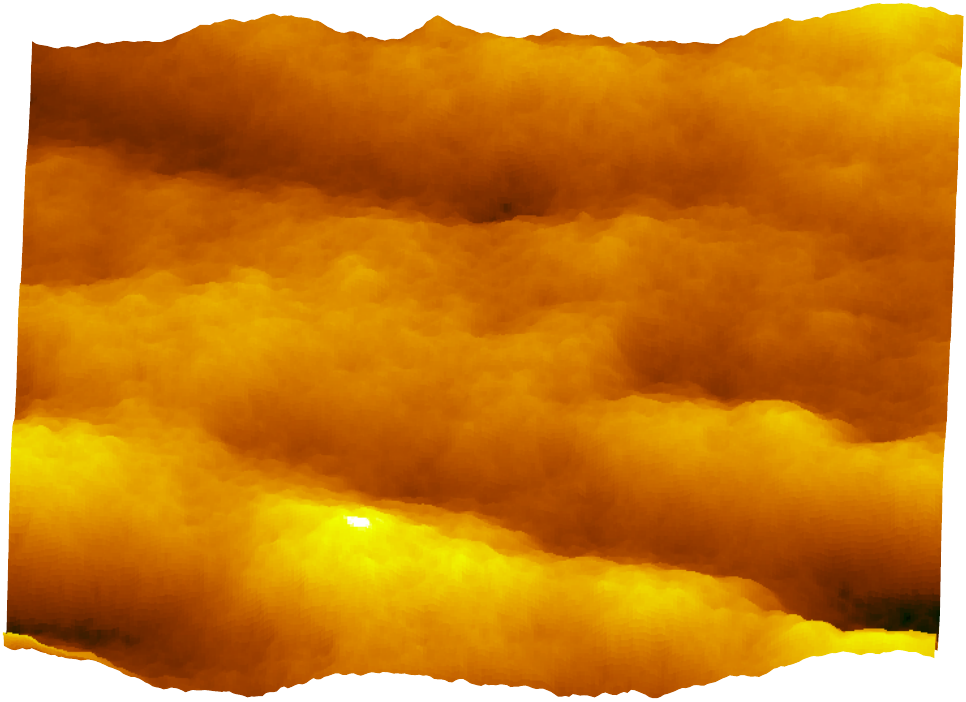}}
	\hfill
	\subfloat[UNet-opt\label{fig:cGANestimate5}]{%
		\includegraphics[trim={1cm 1cm 2cm 1cm},clip,width=0.17\linewidth,height=.05\textheight]{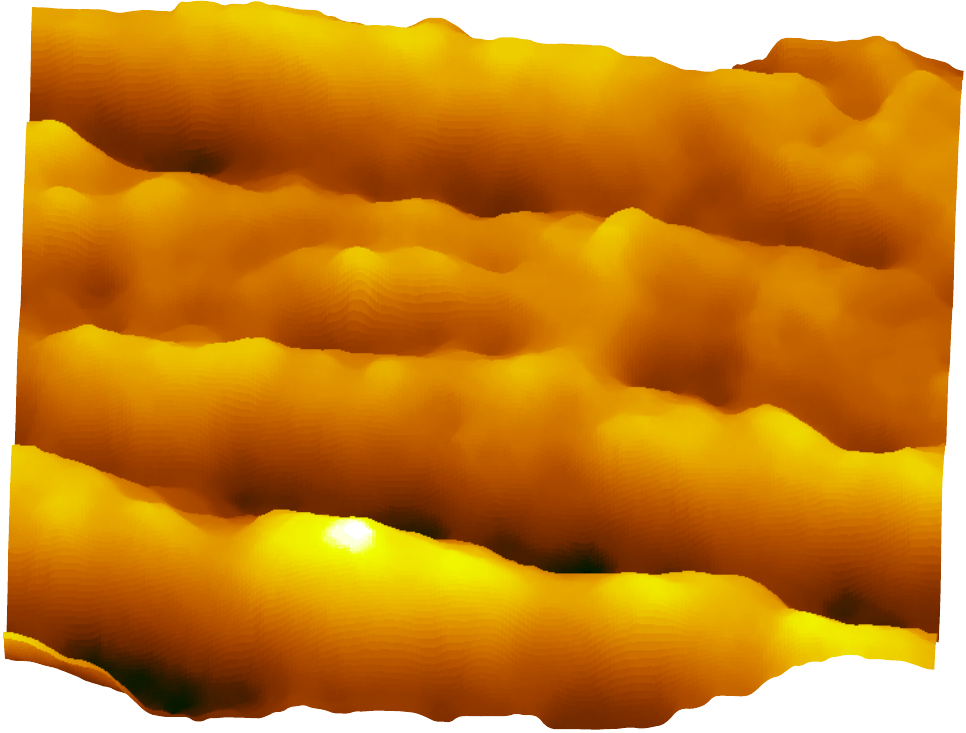}} \\
	
	\subfloat[$I_3$\label{fig:Intensity6}]{%
		\includegraphics[trim={1cm 1cm 1cm 1cm},clip,width=0.17\linewidth,height=.05\textheight]{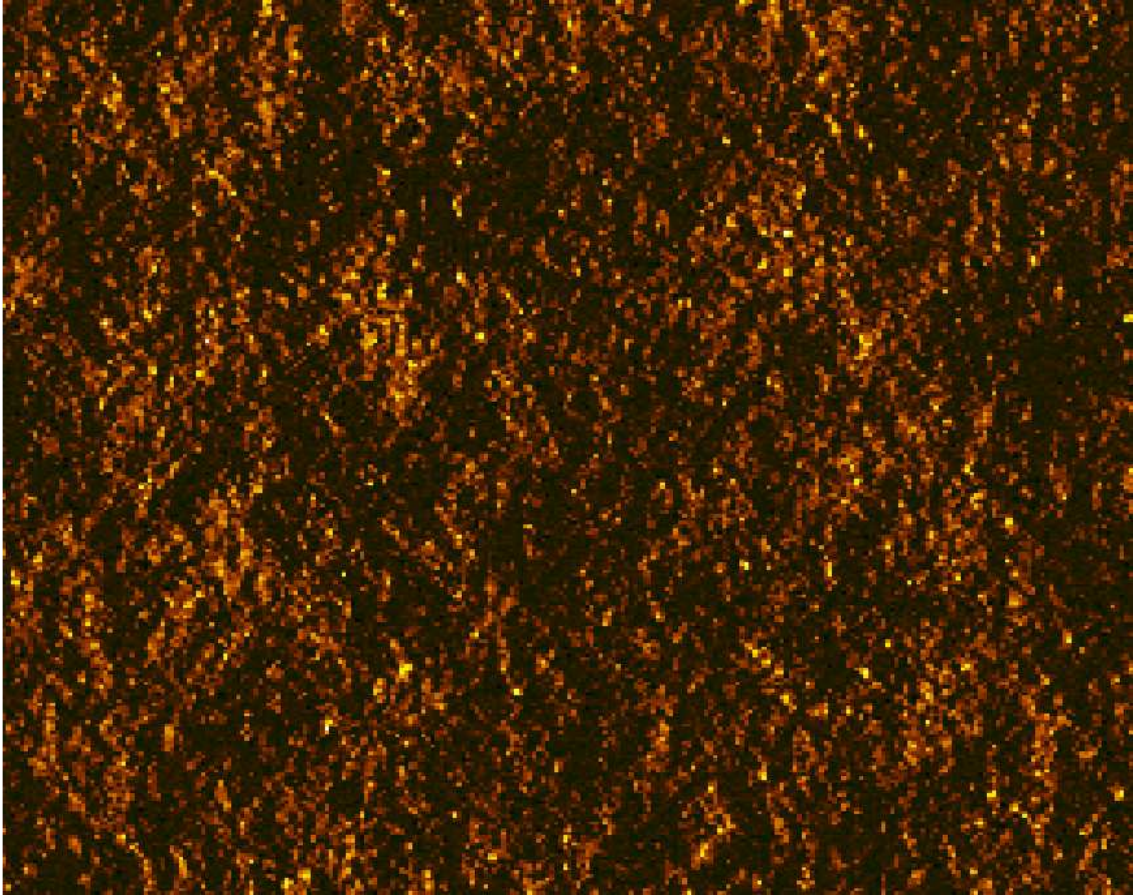}}
	\hfill
	\subfloat[$\mathcal{H}_3$\label{fig:Height6}]{%
		\includegraphics[trim={1cm 1cm 2cm 1cm},clip,width=0.17\linewidth,height=.05\textheight]{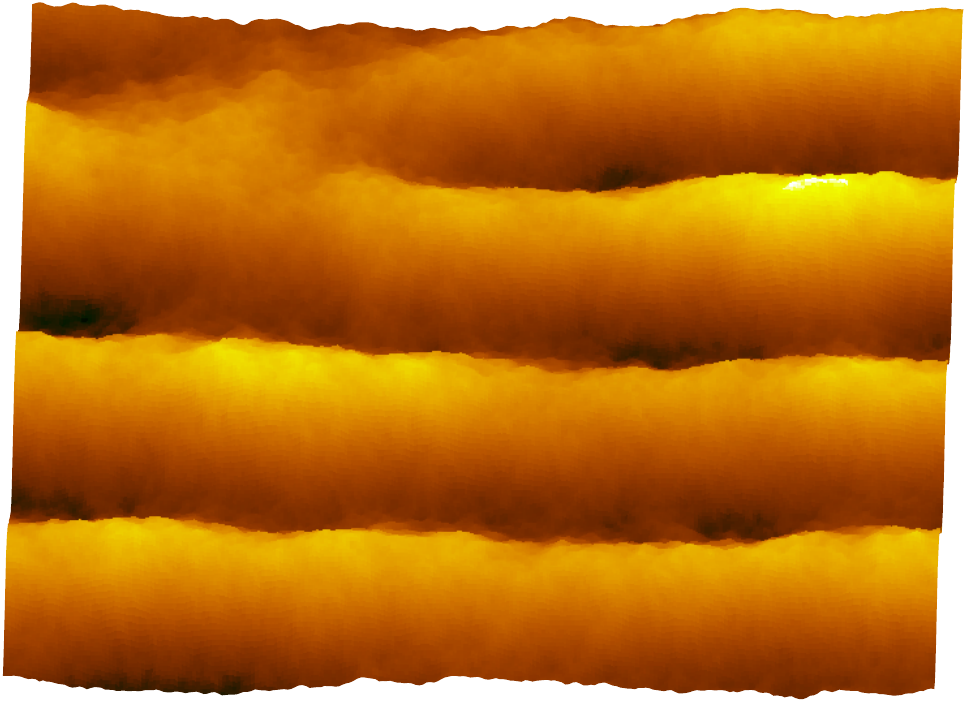}}
	\hfill
	\subfloat[GMRF\label{fig:GMRFestimate6}]{%
		\includegraphics[trim={1cm 1cm 2cm 1cm},clip,width=0.17\linewidth,height=.05\textheight]{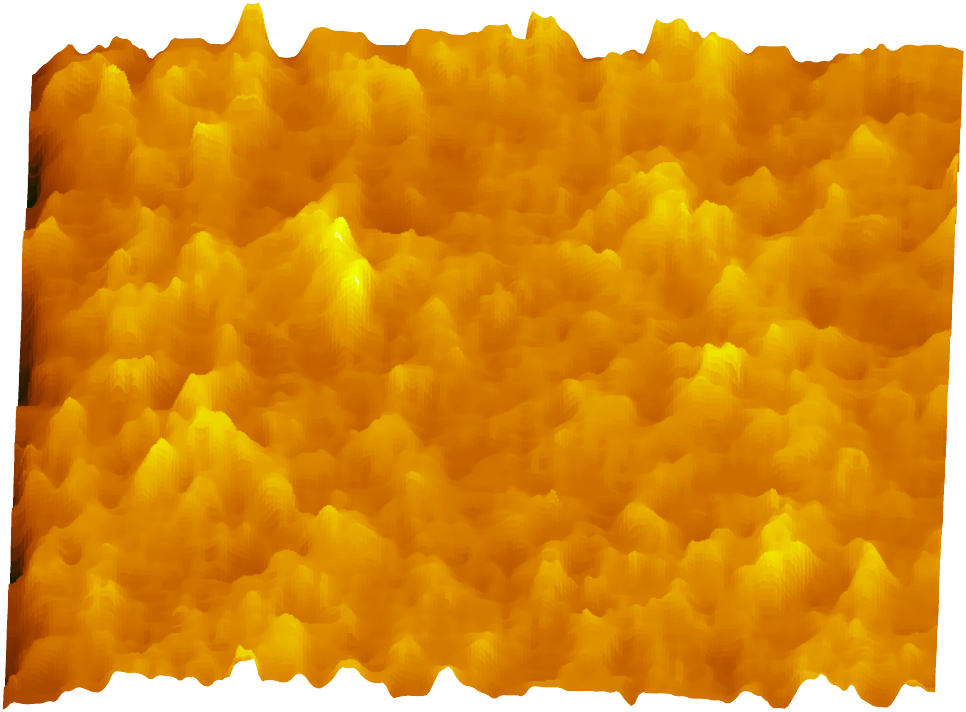}}
	\hfill
	\subfloat[pix2pix\label{fig:UNetestimate6}]{%
		\includegraphics[trim={1cm 1cm 2cm 1cm},clip,width=0.17\linewidth,height=.05\textheight]{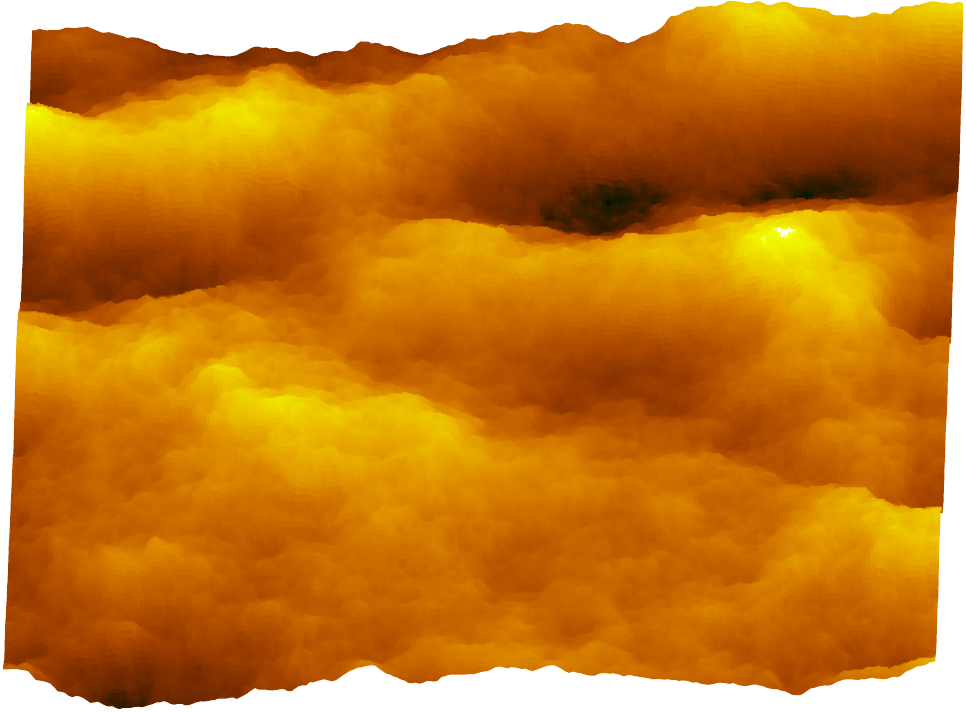}} 
	\hfill
	\subfloat[UNet-opt\label{fig:cGANestimate6}]{%
		\includegraphics[trim={1cm 1cm 2cm 1cm},clip,width=0.17\linewidth,height=.05\textheight]{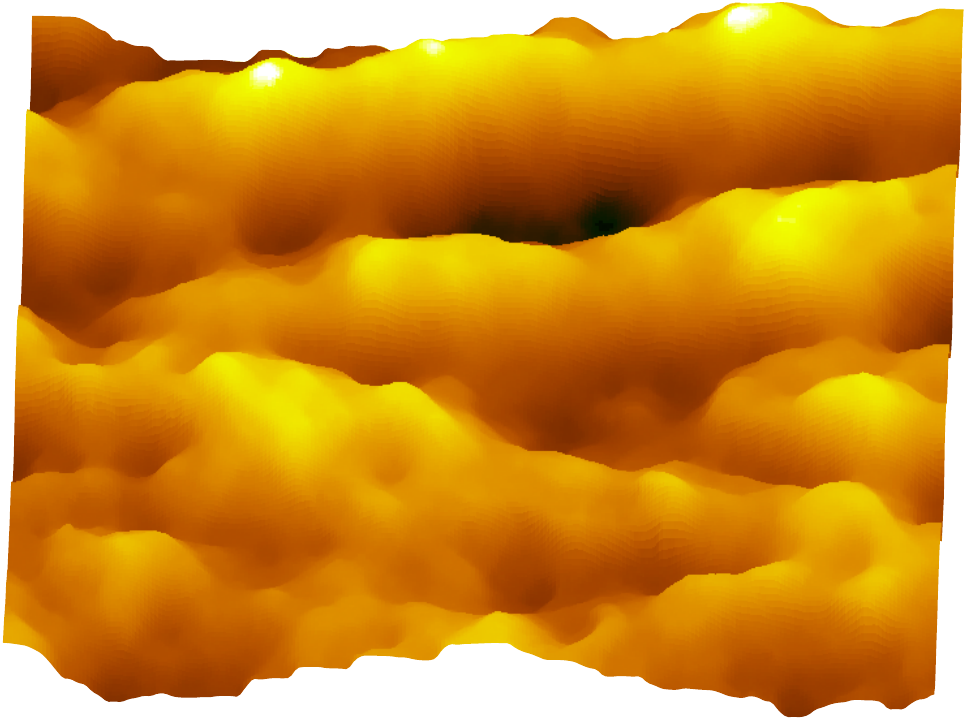}}\\
	\caption{Examples of seabed relief estimation on sand-ripple textures that have been beamformed from difficult cases. The top two seabed relief maps (b and g) are from five degree offsets of an orthogonal path while the bottom seabed relief map (l) is from an orthogonal path. Three intensity images (a,f,k) and their coregistered seabed relief maps (b,g,l) are shown.  GMRF estimates (c,h,m) do not learn this sophisticated special case like the UNet-opt estimates (e,j,o) and pix2pix estimates (d,i,n) do.}
	\label{fig:orthogonal-sand-ripples} 
\end{figure}

\begin{figure}[h!]
	\centering
	\subfloat[$I_1$\label{7a}]{%
		\includegraphics[trim={1cm 1cm 1cm 1cm},clip,width=0.17\linewidth,height=.05\textheight]{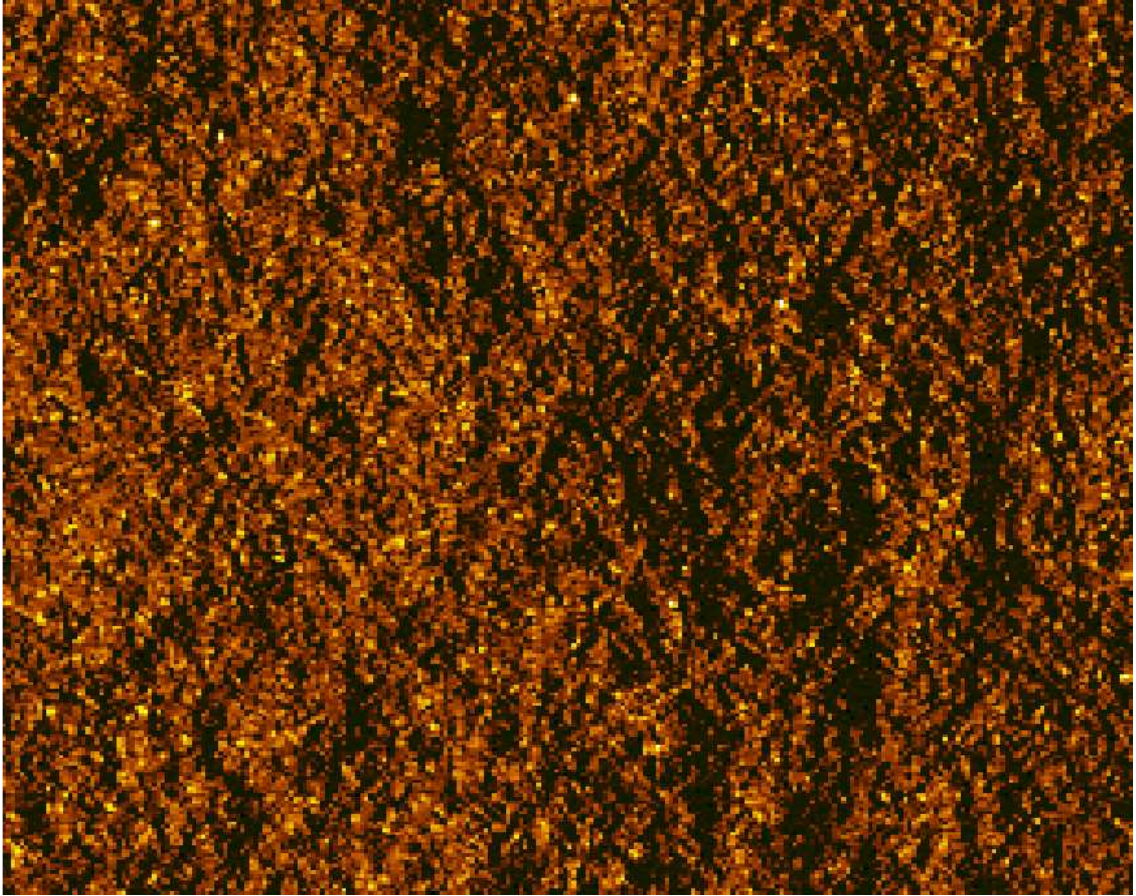}}
	\hfill
	\subfloat[$\mathcal{H}_1$\label{7b}]{%
		\includegraphics[trim={1cm 1cm 2cm 1cm},clip,width=0.17\linewidth,height=.05\textheight]{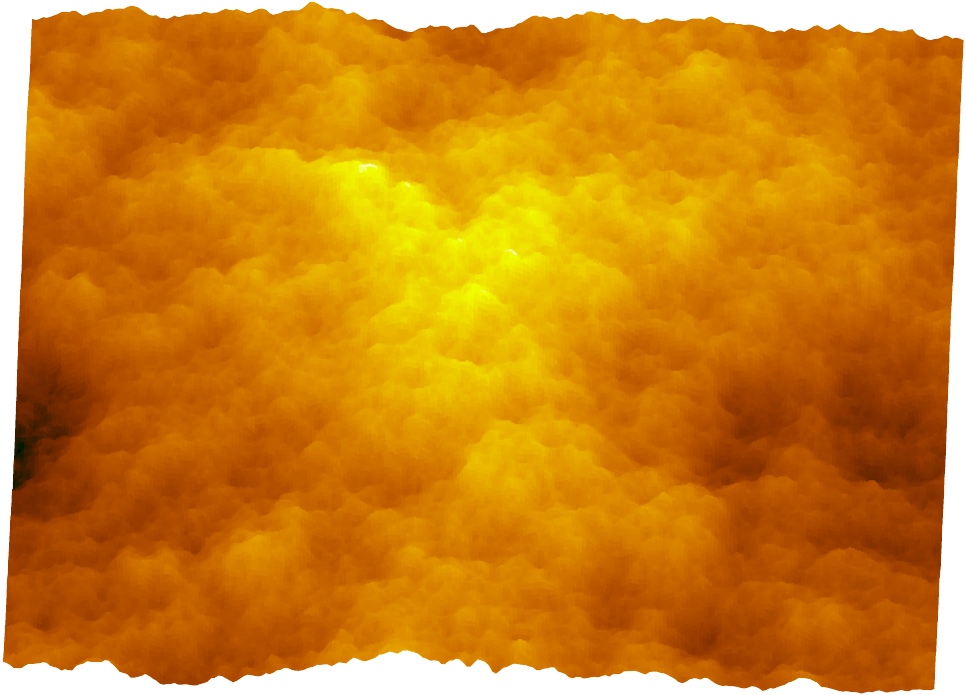}}
	\hfill
	\subfloat[GMRF\label{7c}]{%
		\includegraphics[trim={1cm 1cm 2cm 1cm},clip,width=0.17\linewidth,height=.05\textheight]{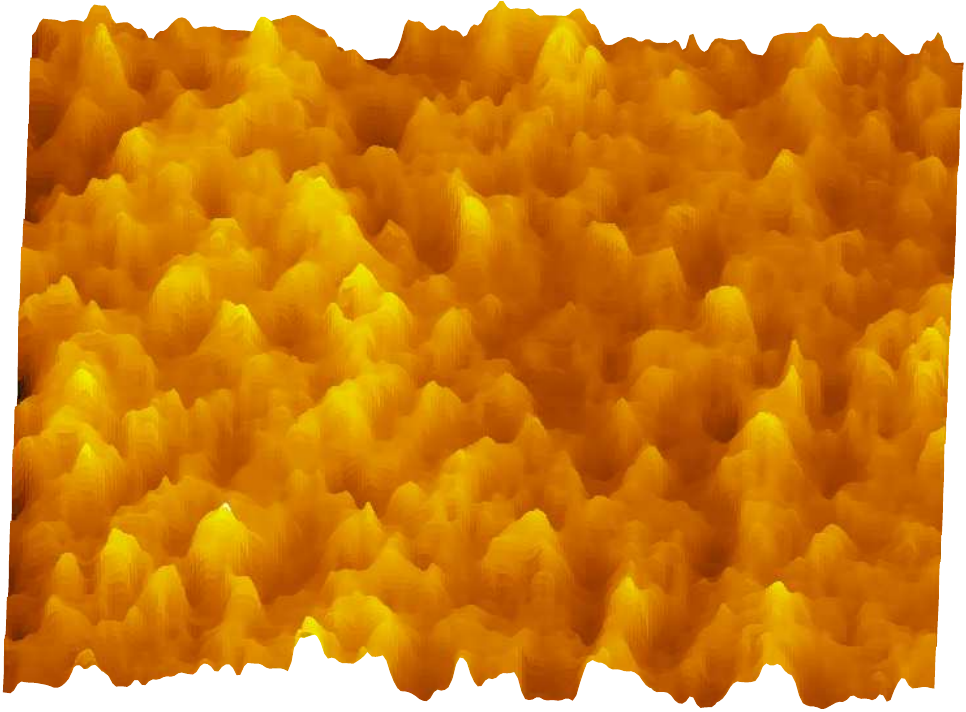}}
	\hfill
	\subfloat[pix2pix\label{7d}]{%
		\includegraphics[trim={1cm 1cm 2cm 1cm},clip,width=0.17\linewidth,height=.05\textheight]{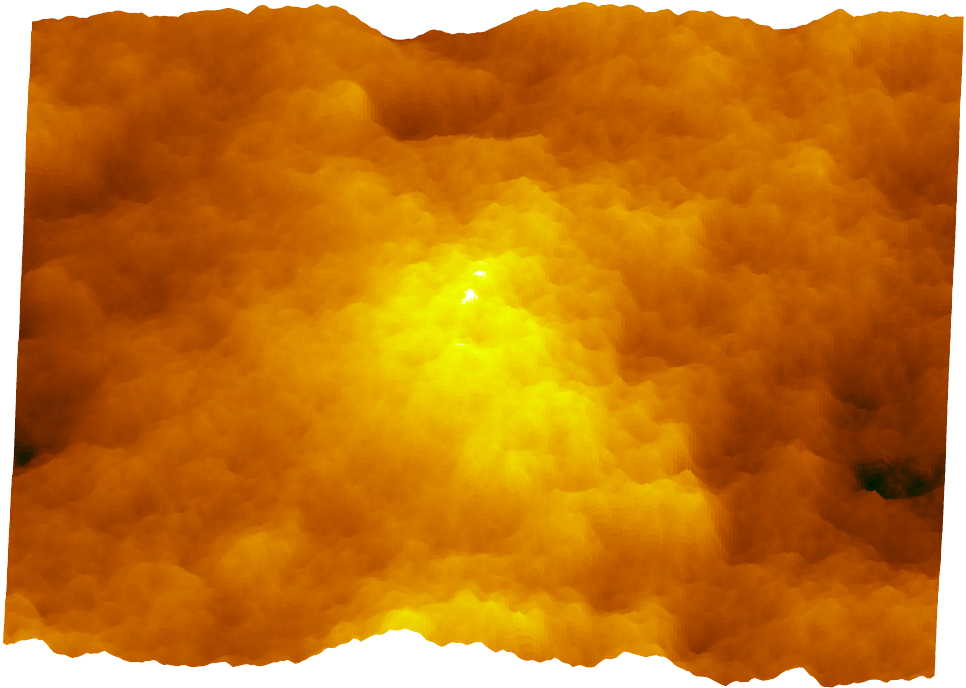}}  
	\hfill
	\subfloat[UNet-opt\label{7e}]{%
		\includegraphics[trim={1cm 1cm 2cm 1cm},clip,width=0.17\linewidth,height=.05\textheight]{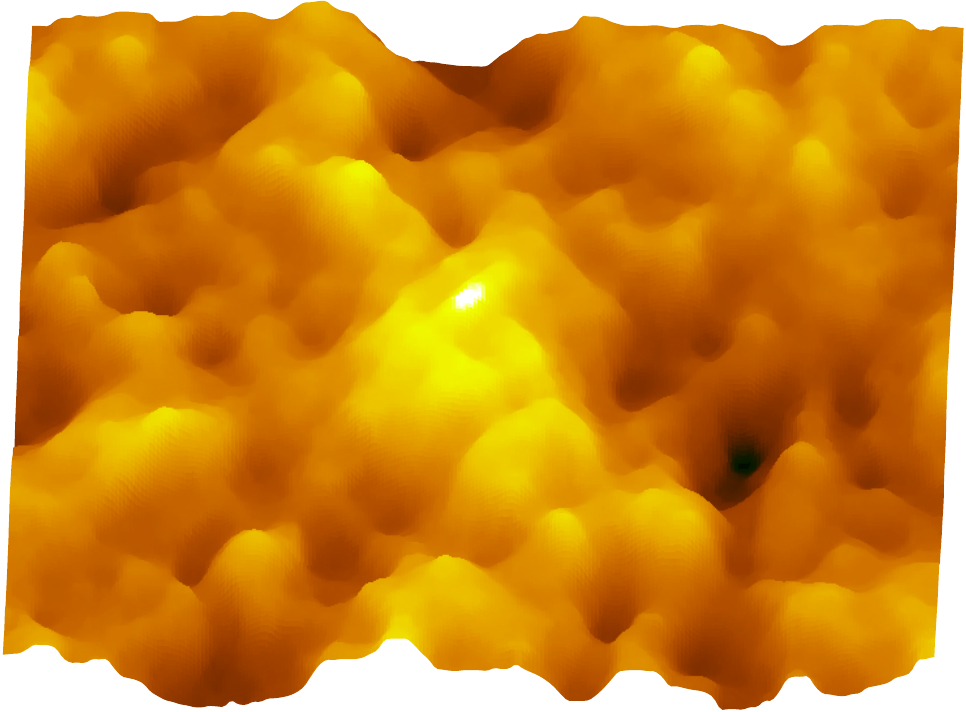}}\\
	
	\subfloat[$I_2$\label{8a}]{%
		\includegraphics[trim={1cm 1cm 1cm 1cm},clip,width=0.17\linewidth,height=.05\textheight]{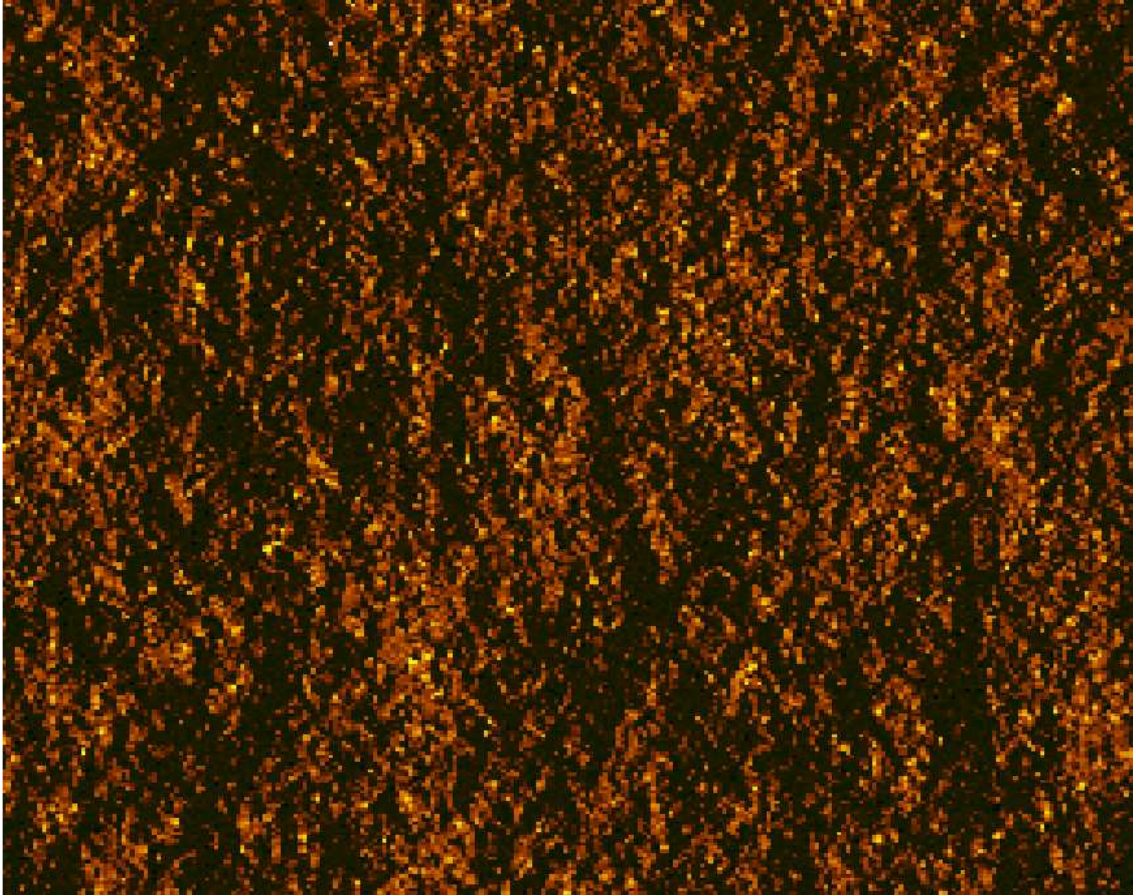}}
	\hfill
	\subfloat[$\mathcal{H}_2$\label{8b}]{%
		\includegraphics[trim={1cm 1cm 2cm 1cm},clip,width=0.17\linewidth,height=.05\textheight]{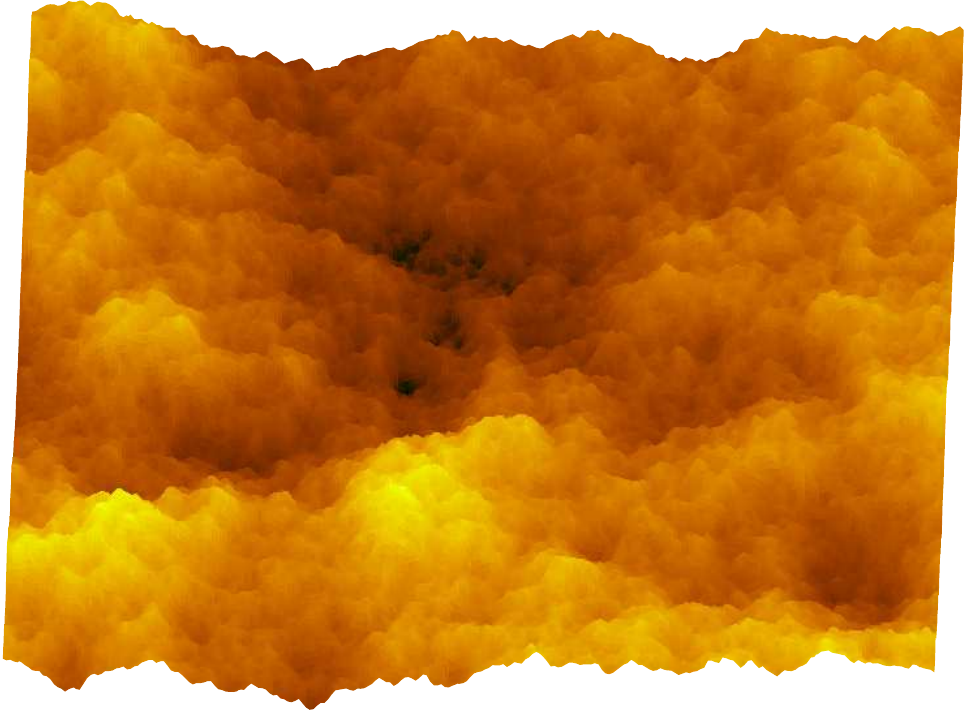}}
	\hfill
	\subfloat[GMRF\label{8c}]{%
		\includegraphics[trim={1cm 1cm 2cm 1cm},clip,width=0.17\linewidth,height=.05\textheight]{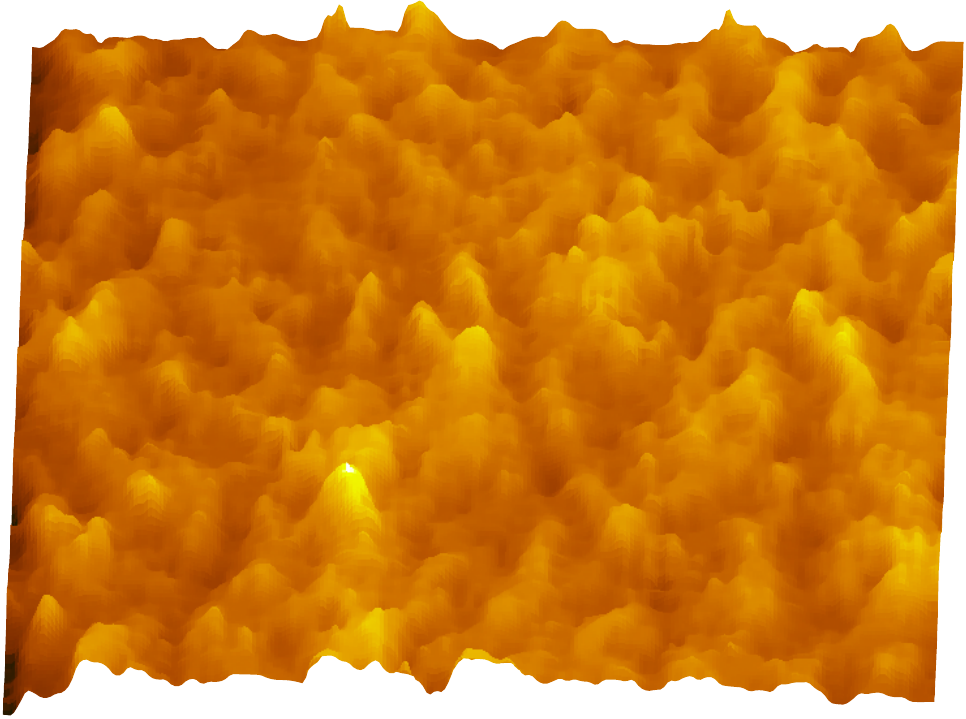}}
	\hfill
	\subfloat[pix2pix\label{8d}]{%
		\includegraphics[trim={1cm 1cm 2cm 1cm},clip,width=0.17\linewidth,height=.05\textheight]{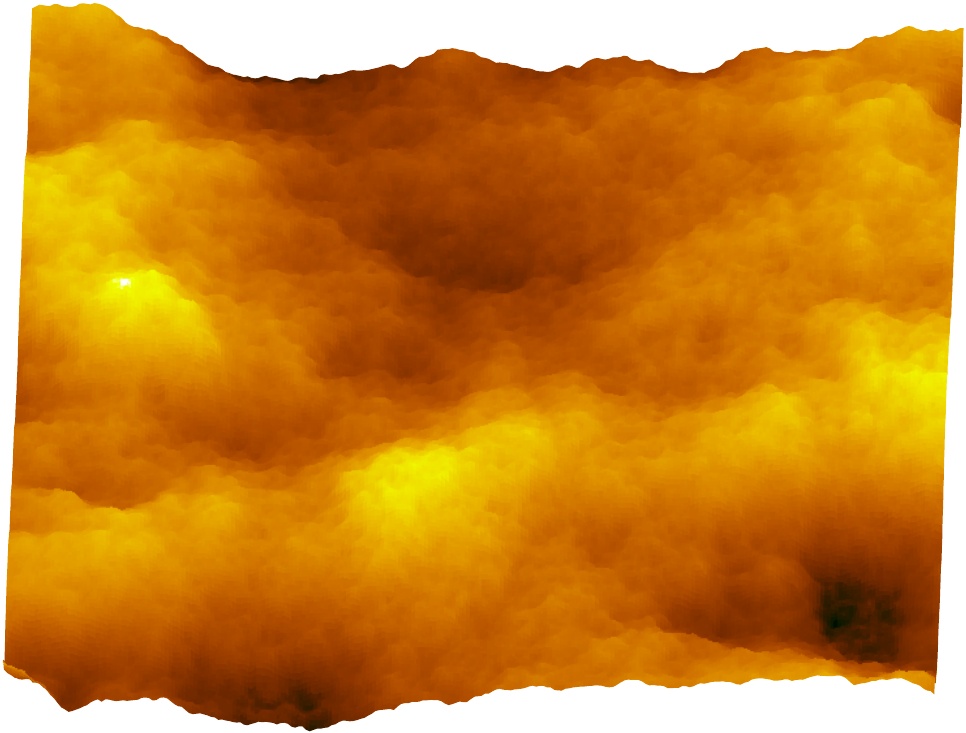}} 
	\hfill
	\subfloat[UNet-opt\label{8e}]{%
		\includegraphics[trim={1cm 1cm 2cm 1cm},clip,width=0.17\linewidth,height=.05\textheight]{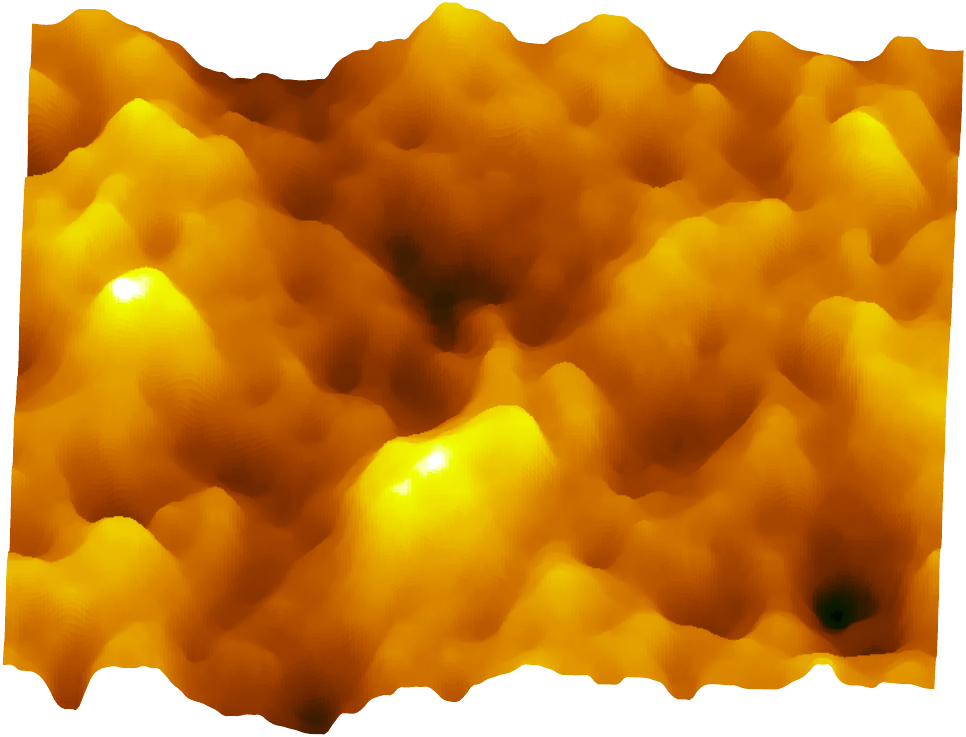}}\\
	
	\subfloat[$I_3$\label{9a}]{%
		\includegraphics[trim={1cm 1cm 1cm 1cm},clip,width=0.17\linewidth,height=.05\textheight]{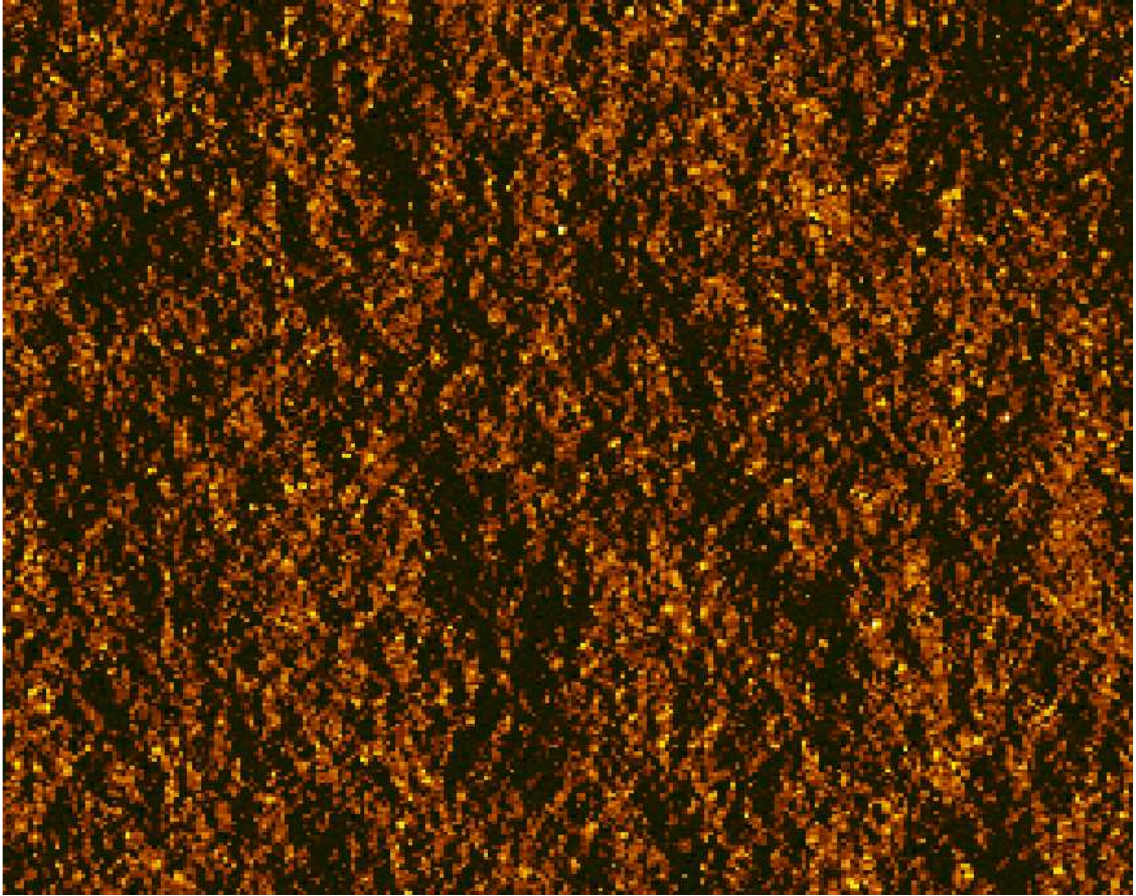}}
	\hfill
	\subfloat[$\mathcal{H}_3$\label{9b}]{%
		\includegraphics[trim={1cm 1cm 2cm 1cm},clip,width=0.17\linewidth,height=.05\textheight]{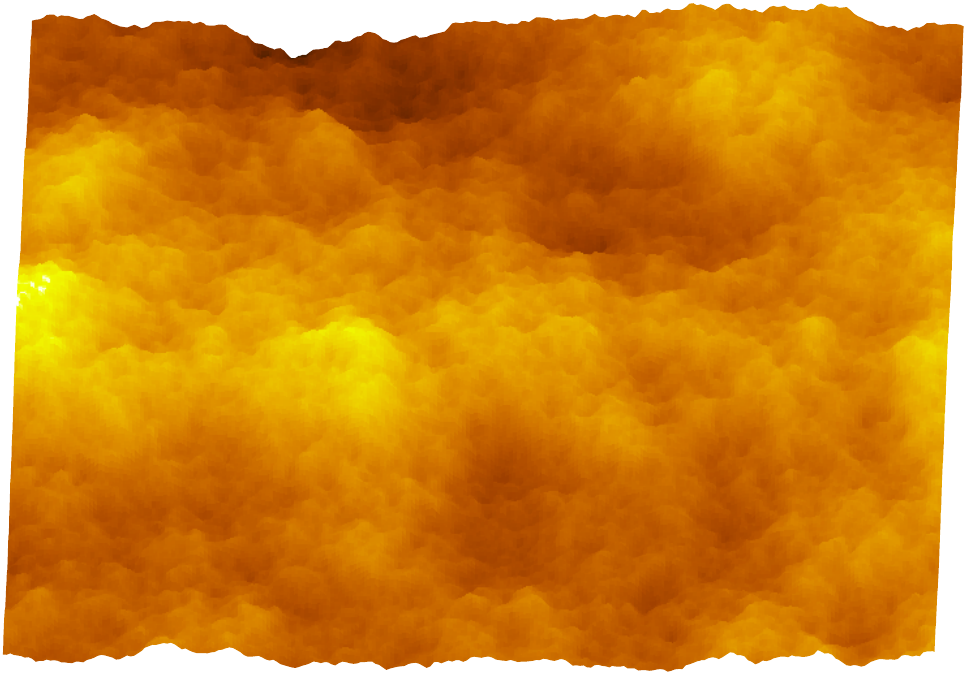}}
	\hfill
	\subfloat[GMRF\label{9c}]{%
		\includegraphics[trim={1cm 1cm 2cm 1cm},clip,width=0.17\linewidth,height=.05\textheight]{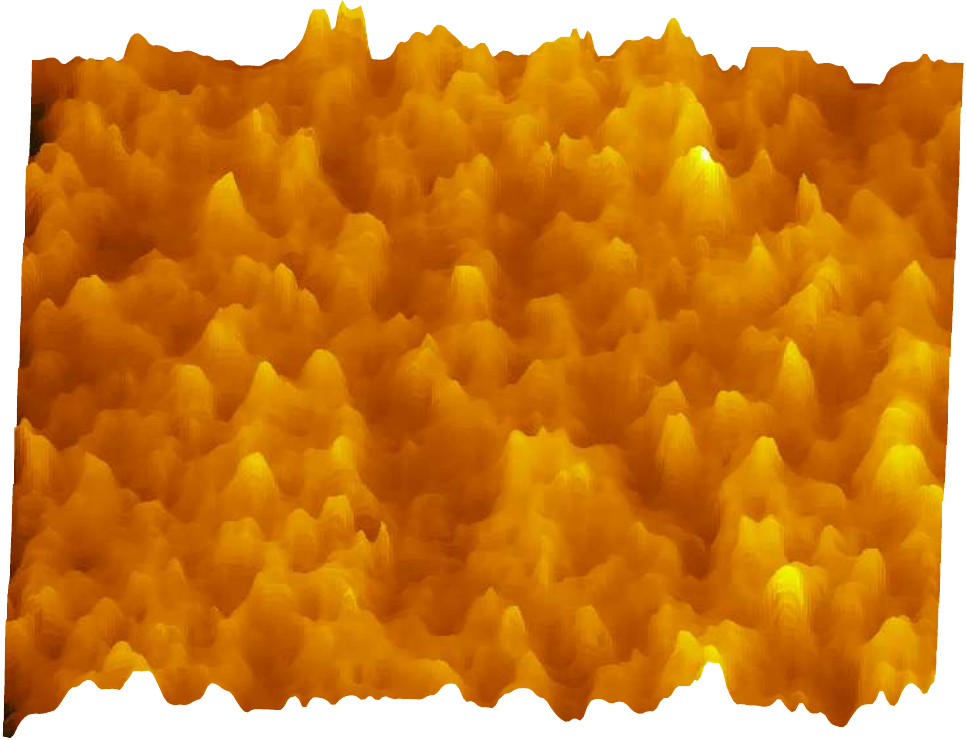}}
	\hfill
	\subfloat[pix2pix\label{9d}]{%
		\includegraphics[trim={1cm 1cm 2cm 1cm},clip,width=0.17\linewidth,height=.05\textheight]{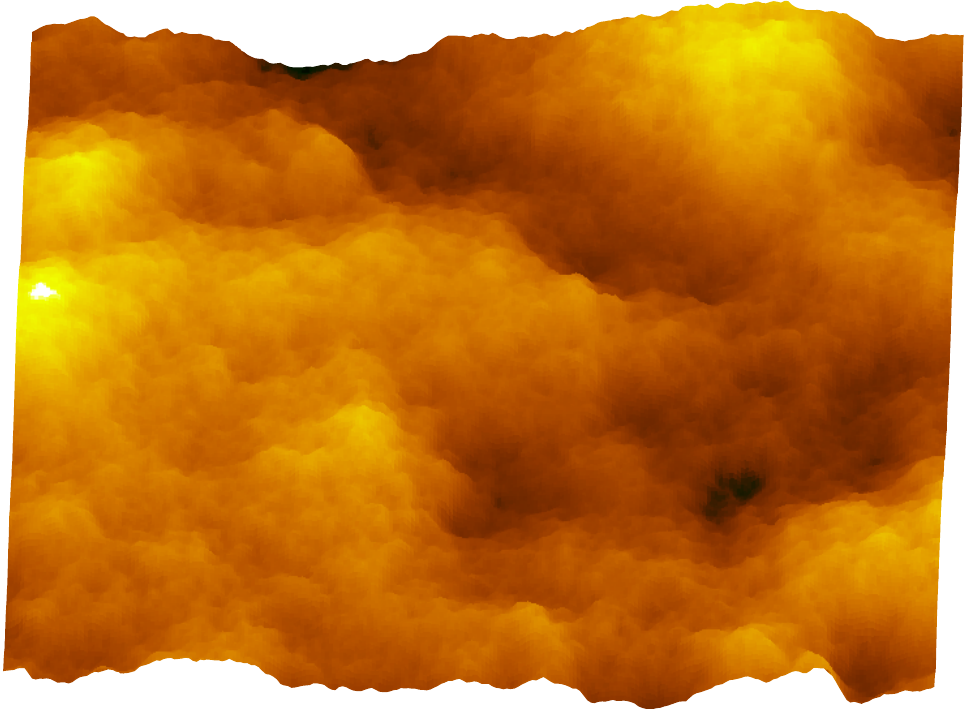}} 
	\hfill
	\subfloat[UNet-opt\label{9e}]{%
		\includegraphics[trim={1cm 1cm 2cm 1cm},clip,width=0.17\linewidth,height=.05\textheight]{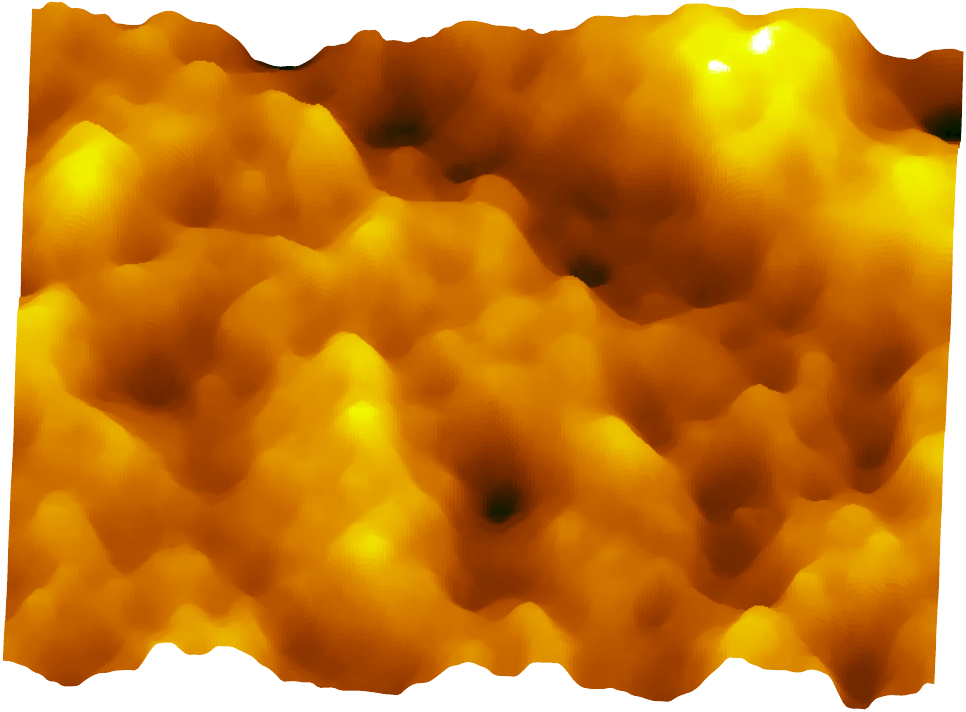}} \\
	\caption{Examples of seabed relief estimation on rough textures. Three intensity images (a,f,k) and their coregistered seabed relief maps (b,g,l) are shown. The pix2pix model (e,j,o) oversmoothes while the UNet-opt (d,i,n) model produces seabed relief maps that are visually closer to the desired seabed relief in each of the three examples. The GMRF model (c,h,m) underestimates the seabed relief in each example.}
	\label{fig:flat-text}
\end{figure}

On more complicated examples, such as intensity generated from orthogonal paths or five degree offsets of orthogonal paths of sand-ripple (Figure \ref{fig:orthogonal-sand-ripples}), the GMRF model is unable to reproduce the desired texture. Seabed relief values produced by the UNet-opt model qualitatively match the desired, while the pix2pix result is oversmoothed. Although this example may seem to be indicative of overtraining on sand-ripples, the UNet-opt model is still able to reproduce rough seabed relief maps (Figure \ref{fig:flat-text}). 

The seabed relief map produced by the pix2pix model matches the desired better than both the GMRF and UNet-opt models (Figure \ref{fig:flat-text}). However, the seabed relief map is a bit oversmoothed compared to the desired result. One may assume this is a result of the tradeoff between the tendency of the pix2pix model to oversmooth all textures. The UNet-opt incorrectly estimates the seabed relief for some portions of the roughness maps. This may be attributed to the imbalance of our dataset which consists of $40\%$ roughness maps and $60\%$ sand-ripple. 

Most examples in the dataset are similar to those shown in Figure \ref{fig:small-mix}. Qualitatively, pix2pix produces predictions that are similar to the desired output but are oversmoothed compared to the UNet-opt results. Like the textures shown in Figure \ref{fig:sand-ripples}, the GMRF results contain many small bumps on the peaks of the ripples. These examples demonstrate that a much simpler model (UNet-opt) can perform as well as or better than pix2pix on complex textures. 

The reliance on over a million parameters for the base UNet model improves upon the GMRF model by $20\%$ on the test set. Variations of the UNet architecture which require roughly $25\%$ and $12.5\%$ as many parameters achieve in the former case a reduction in error (UNet-opt) and the latter a slightly larger error. After a certain decrease in the number of parameters, the performance degrades for the case of the UNet-c2 model. The pix2pix model uses more parameters than the UNet-opt model due to the patchGAN discriminator containing over 250,000 parameters. However, training GAN models can be tedious and unstable, which in our case may be the reason pix2pix is outperformed by the simpler models on the simulated dataset. For example, it is difficult to know when training should stop because the loss curves are difficult to track. Additionally, GANs can succumb to issues such as mode collapse where the discriminator is stuck in a local minima and the generator produces a small set of samples which receieve low error \cite{thanhtung2020catastrophic}. 

\paragraph{Effects of bottleneck size} The bottleneck size in the UNet model controls the dimensionality of the compressed representation of the data. With too small of a bottleneck size, the decoder portion of the network is unable to recover the encoded information back to the original representation. Conversely, with too large of a bottleneck size, the network updates parameters which are unnecessary to recover the data appropriately and significant resources can be wasted. Between the UNet model, UNet-opt, and UNet-c2 models the bottleneck is changed from 256 to 128 to 64 respectively. 

With changing the bottleneck size, the average test error is similar between the first two models. As the bottleneck size is decreased further, the UNet-c2 model has a slight increase (Table \ref{tab:testPISAS}). This supports using a larger bottleneck size than 64 but not necessarily 256. With a small error decrease between the UNet and UNet-opt, reducing the number of parameters actually improved the performance.

\paragraph{Effects of network depth} Changing the network depth and widening each layer allows a comparison of similar numbers of parameters. Consequently, the number of fine-scaled representations that are passed through the skip connections is reduced. In the standard UNet architecture, four feature map outputs are copied from the encoder representation and concatenated onto the decoder representations. We reduce the number of decoder and encoder layers to six while keeping the number of parameters similar for the same bottleneck sizes. We test the effect of using one fewer skip connection.

Reducing network depth caused slight decreases in the performance of each model regardless of the bottleneck size (Table \ref{tab:testPISAS}). Based on the lowest error while relying on less parameters than the original model, we elected to use the UNet model with 128 bottleneck size (UNet-opt) for further experiments.

\subsection{Fine-tuning Experiments}
Our PISAS dataset contains low-resolution emulated realizations from simulated seabed relief maps. However, the PoSSM dataset was produced using a simulation tool that provides realistic intensity given an input seabed relief map. Therefore, we take our best performing model trained on the PISAS dataset and fine-tune with the same split of data on the PoSSM data. Low-fidelity samples are used to train the weights and high-fidelity samples are used to fine-tune them.

The base model for our fine-tuning experiments is the UNet-opt model trained on the PISAS dataset (UNet-opt-pt). Two fine-tuned models were trained given the UNet-op-pt weights as a starting point. Weights were updated by fine-tuning on the PoSSM data for 100 epochs using a learning rate of 0.1. Every parameter was updated in the first model (UNet-opt-ft). Only the parameters in the last convolutional layer were updated in the second model (UNet-opt-ftc). We compare the performance of these models on two collected SAS datasets.

\begin{figure}[h]
	\centering
	\subfloat[$V_1$\label{s2a}]{%
		\includegraphics[trim={1cm 1cm 1cm 1cm},clip,width=0.225\linewidth]{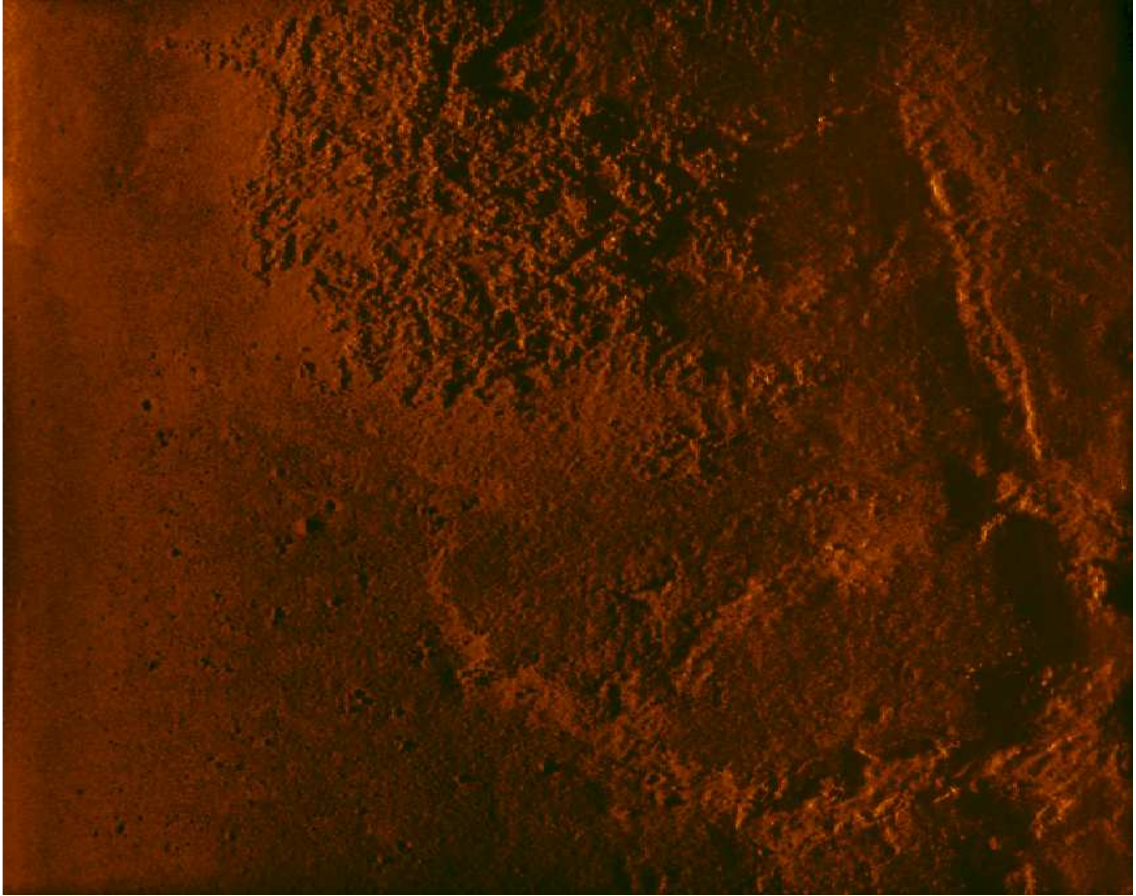}}
	\hfill
	\subfloat[GMRF\label{s2b}]{%
		\includegraphics[trim={1cm 1cm 2cm 1cm},clip,width=0.25\linewidth]{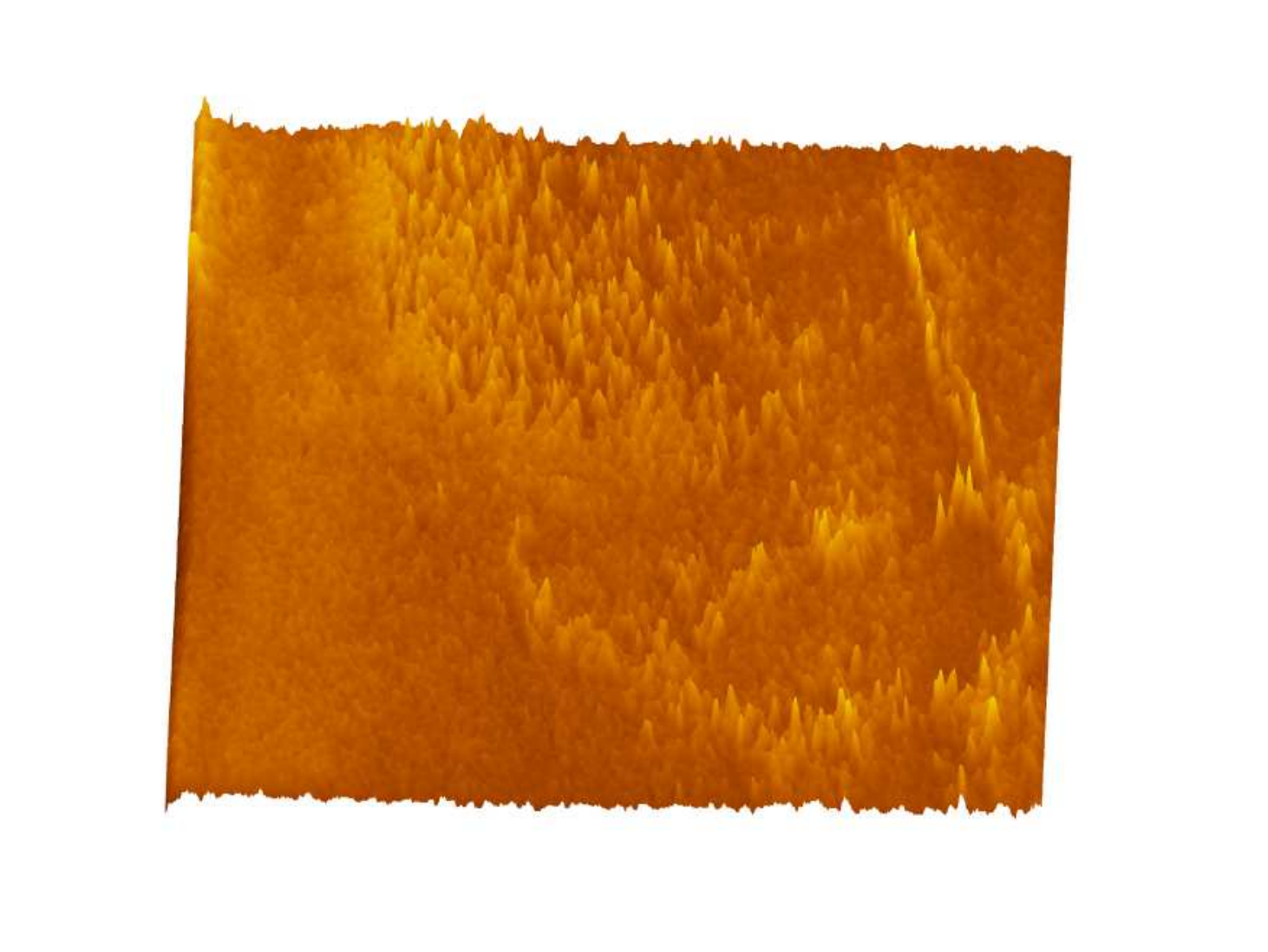}}
	\hfill
	\subfloat[UNet\label{s2c}]{%
		\includegraphics[trim={1cm 1cm 2cm 1cm},clip,width=0.25\linewidth]{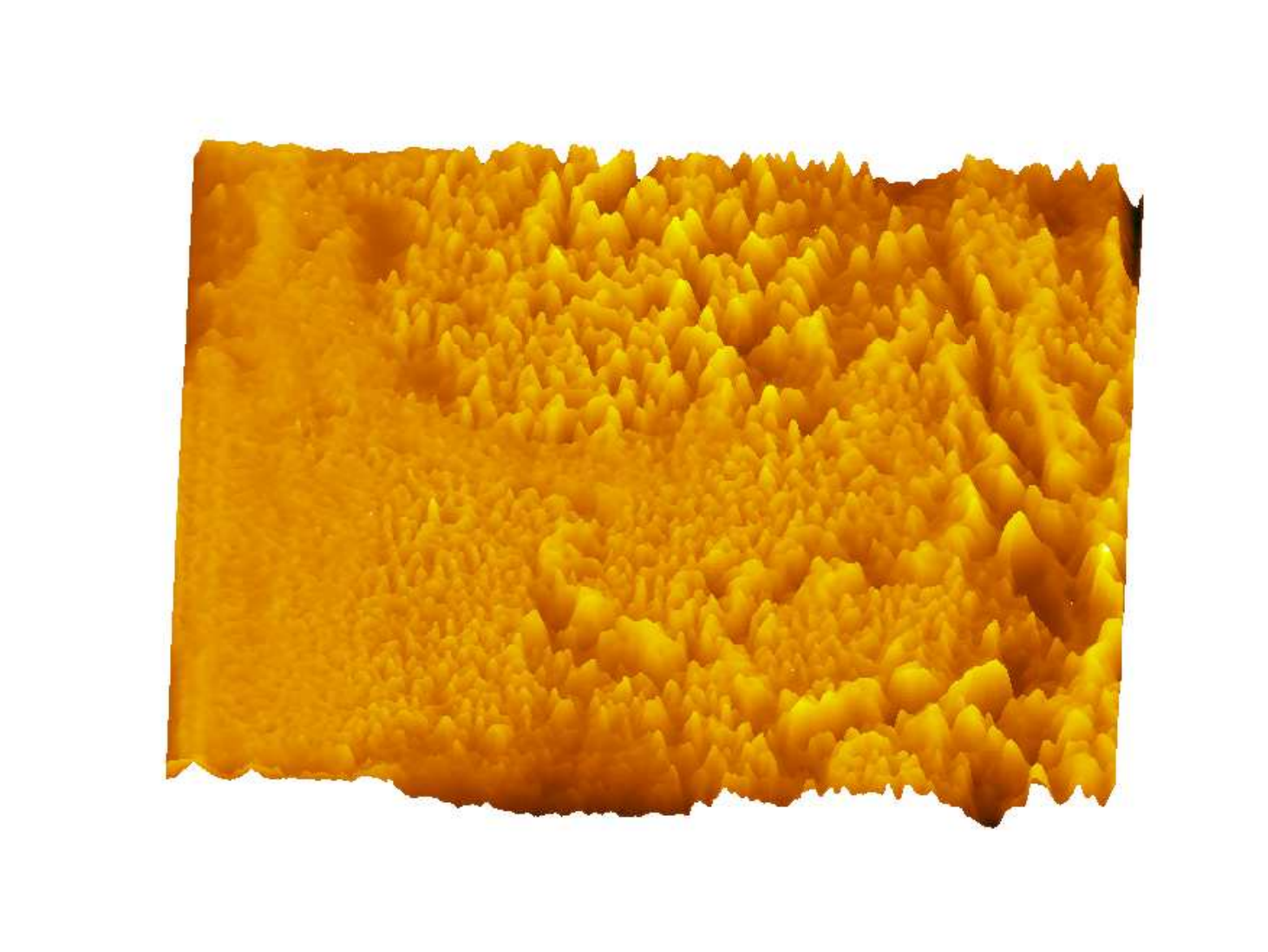}}
	\hfill
	\subfloat[UNet-opt-ftc\label{s2d}]{%
		\includegraphics[trim={1cm 1cm 2cm 1cm},clip,width=0.25\linewidth]{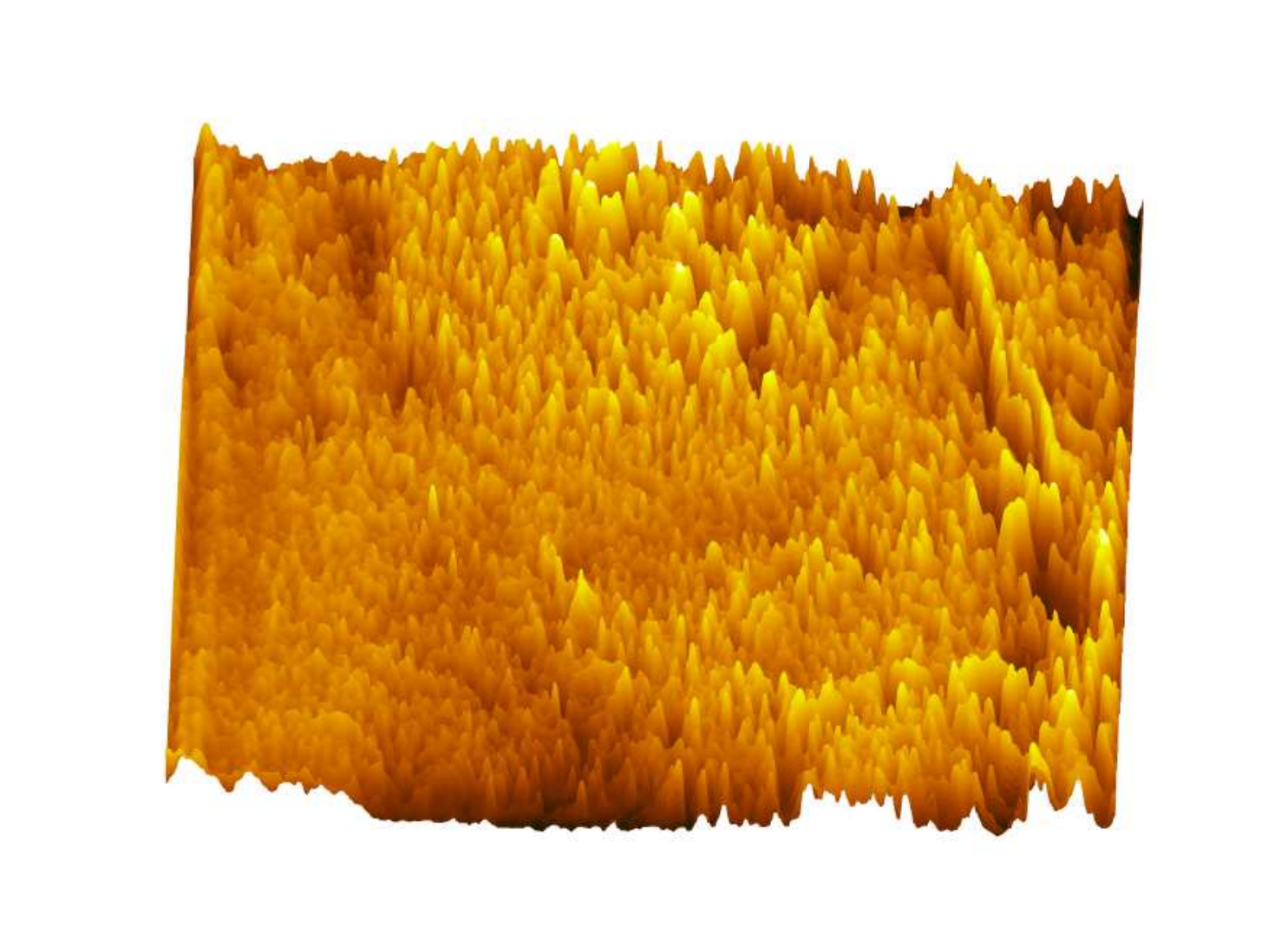}} \\
	
	\subfloat[$V_2$\label{s3a}]{%
		\includegraphics[trim={1cm 1cm 1cm 1cm},clip,width=0.225\linewidth]{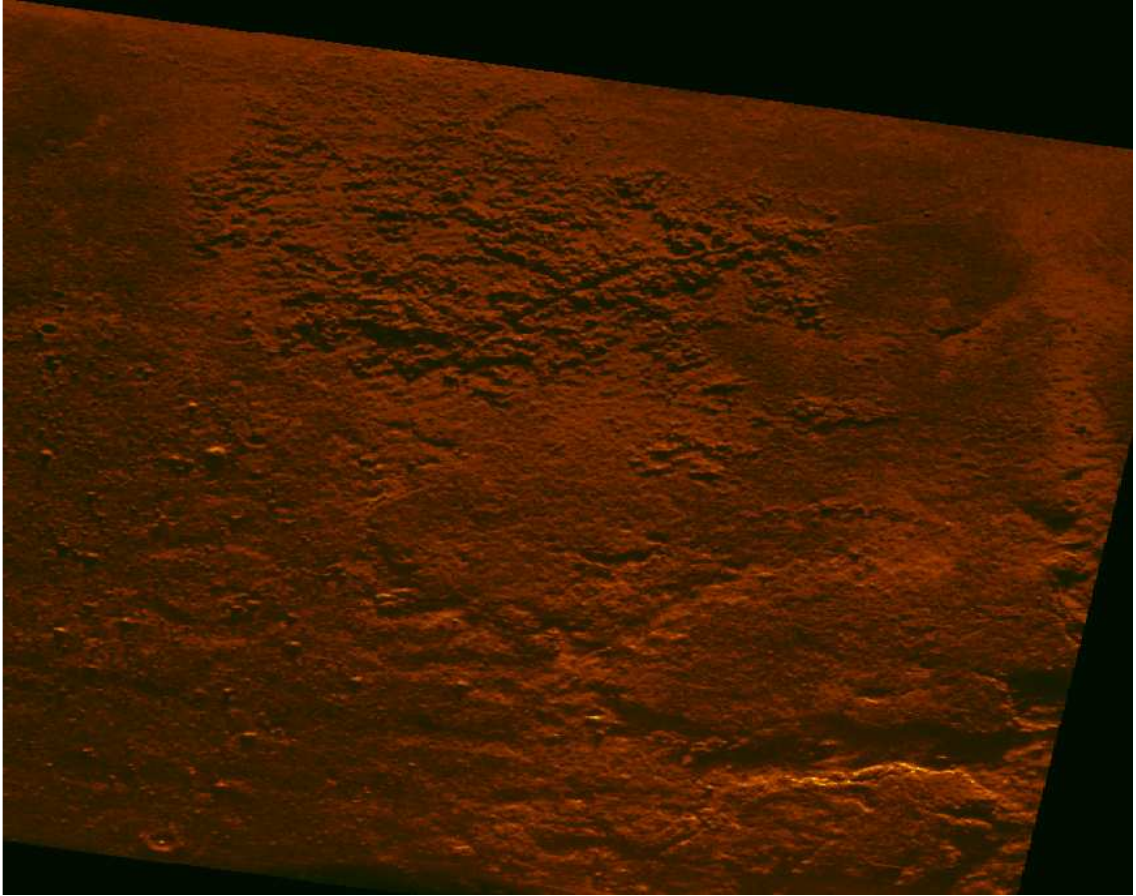}}
	\hfill
	\subfloat[GMRF\label{s3b}]{%
		\includegraphics[trim={1cm 1cm 2cm 1cm},clip,width=0.25\linewidth]{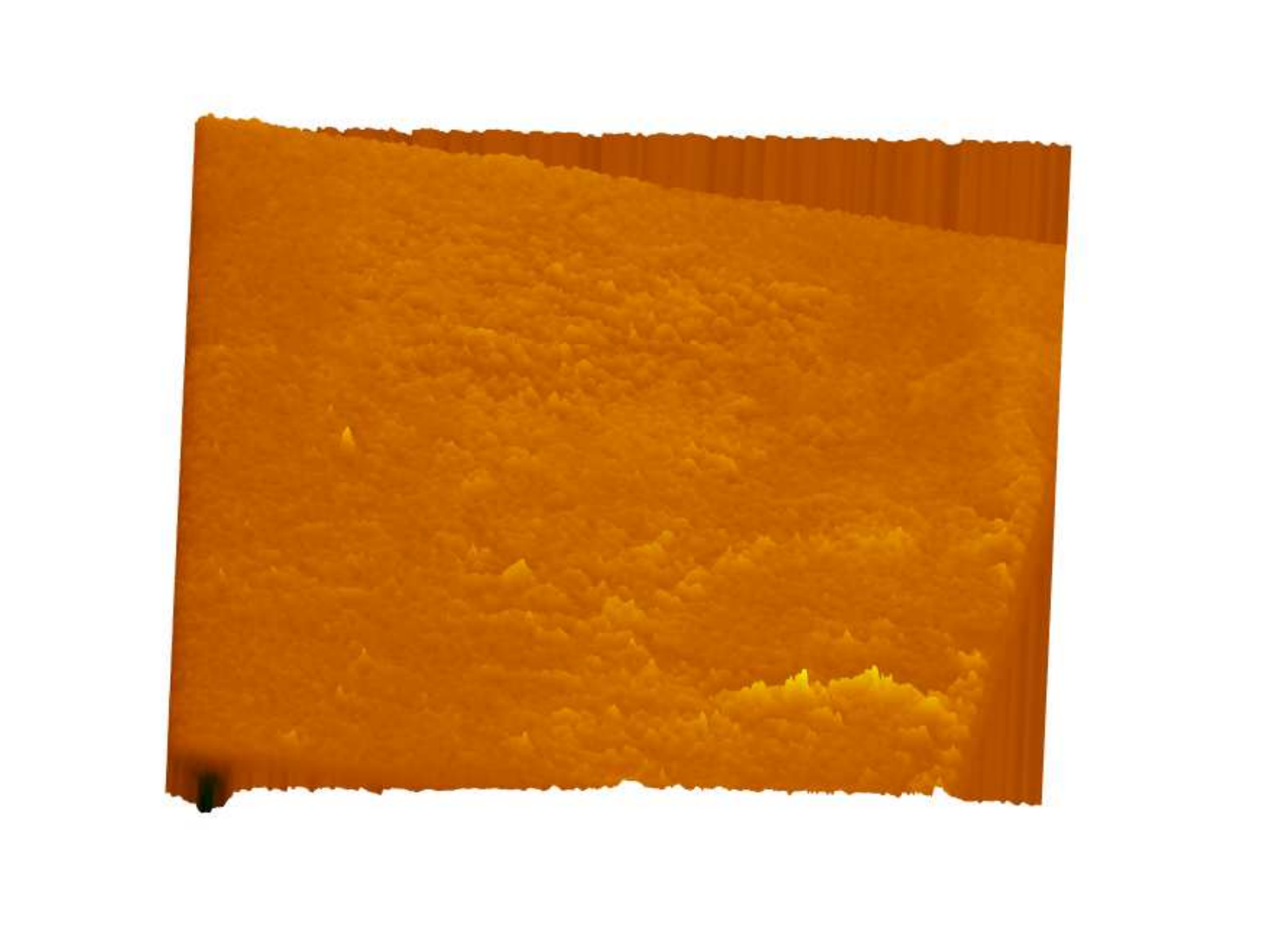}}
	\hfill
	\subfloat[UNet\label{s3c}]{%
		\includegraphics[trim={1cm 1cm 2cm 1cm},clip,width=0.25\linewidth]{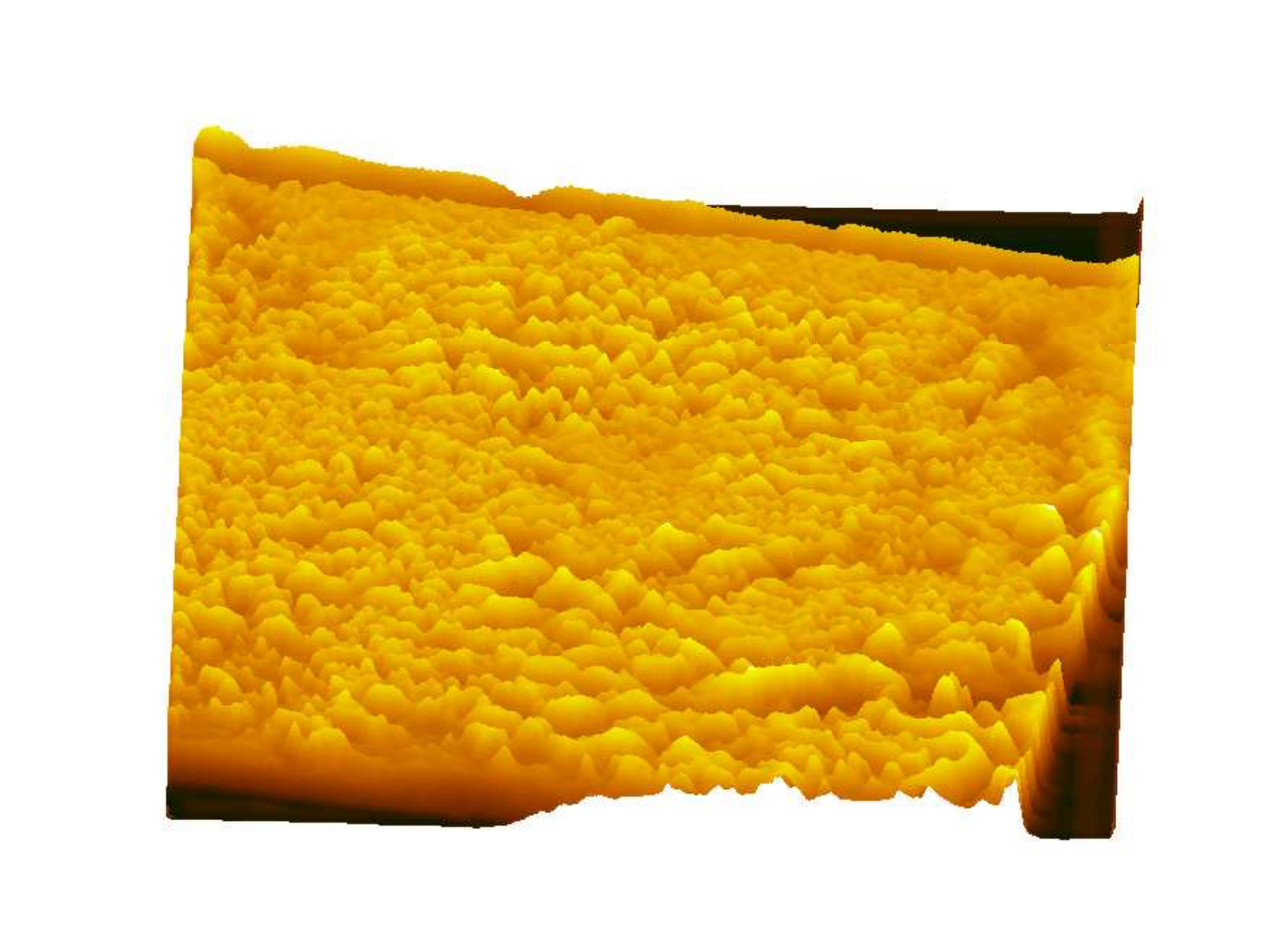}}
	\hfill
	\subfloat[UNet-opt-ftc\label{s3d}]{%
		\includegraphics[trim={1cm 1cm 2cm 1cm},clip,width=0.25\linewidth]{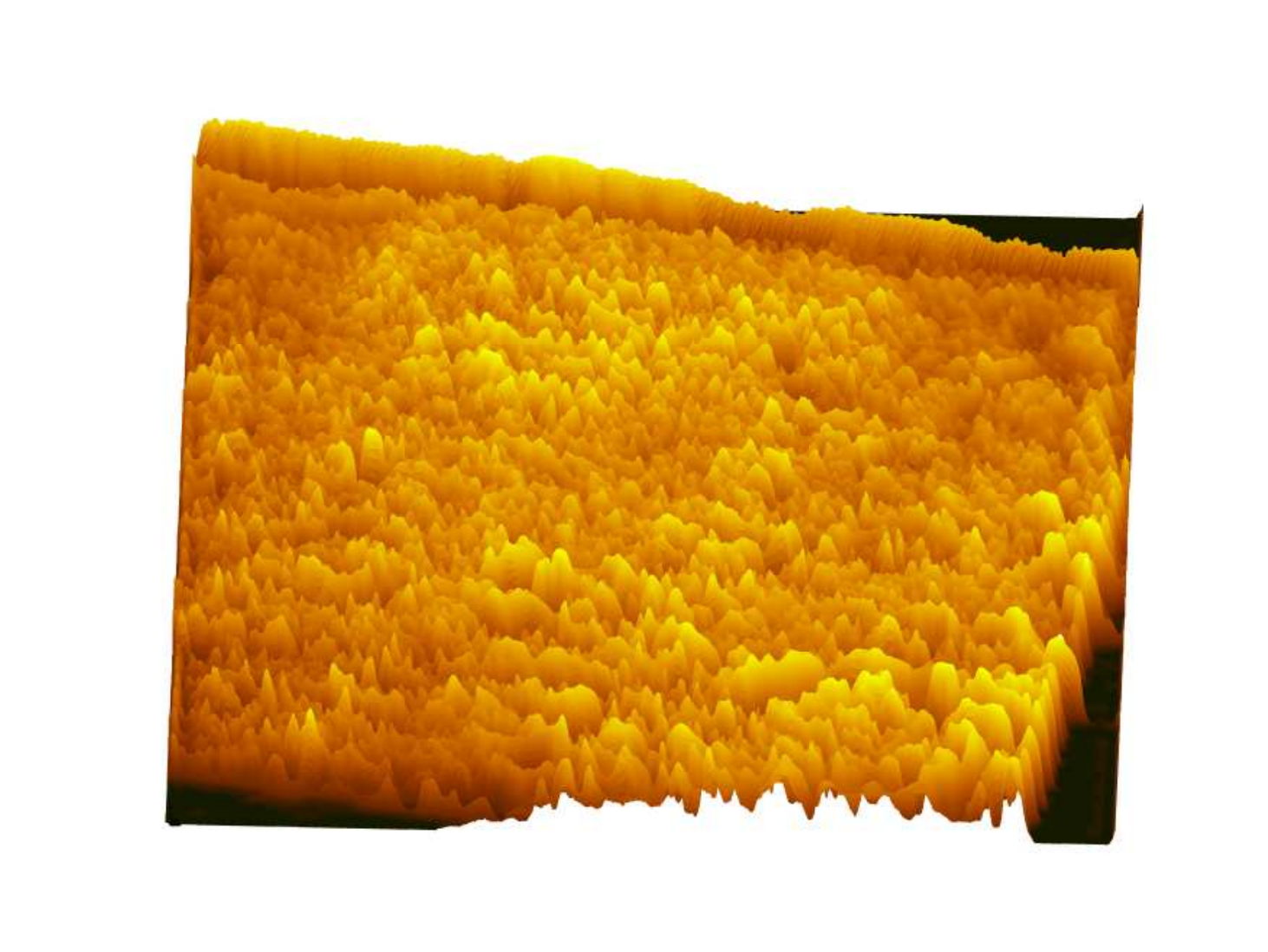}} \\
	\caption{Examples of seabed relief estimation on a scene containing rough, flat, and rocky areas. Views $V_1$ and $V_2$ refer to a pair of coregistered intensity images (a,e). No examples in our training dataset contain rocky textures. However, each model is able to produce seabed relief maps that have differences between the flat and rocky portions of the imagery. The GMRF model (b,f) and standard UNet oversmooth the seabed relief maps (c,g). The UNet-opt-ftc model produces seabed relief maps that are not oversmoothed and more similar across the looks (d,h).}
	\label{fig:scene1} 
\end{figure}

\begin{figure}[h]
	\centering
	\subfloat[$V_1$\label{s1a}]{%
		\includegraphics[trim={1cm 1cm 1cm 1cm},clip,width=0.225\linewidth]{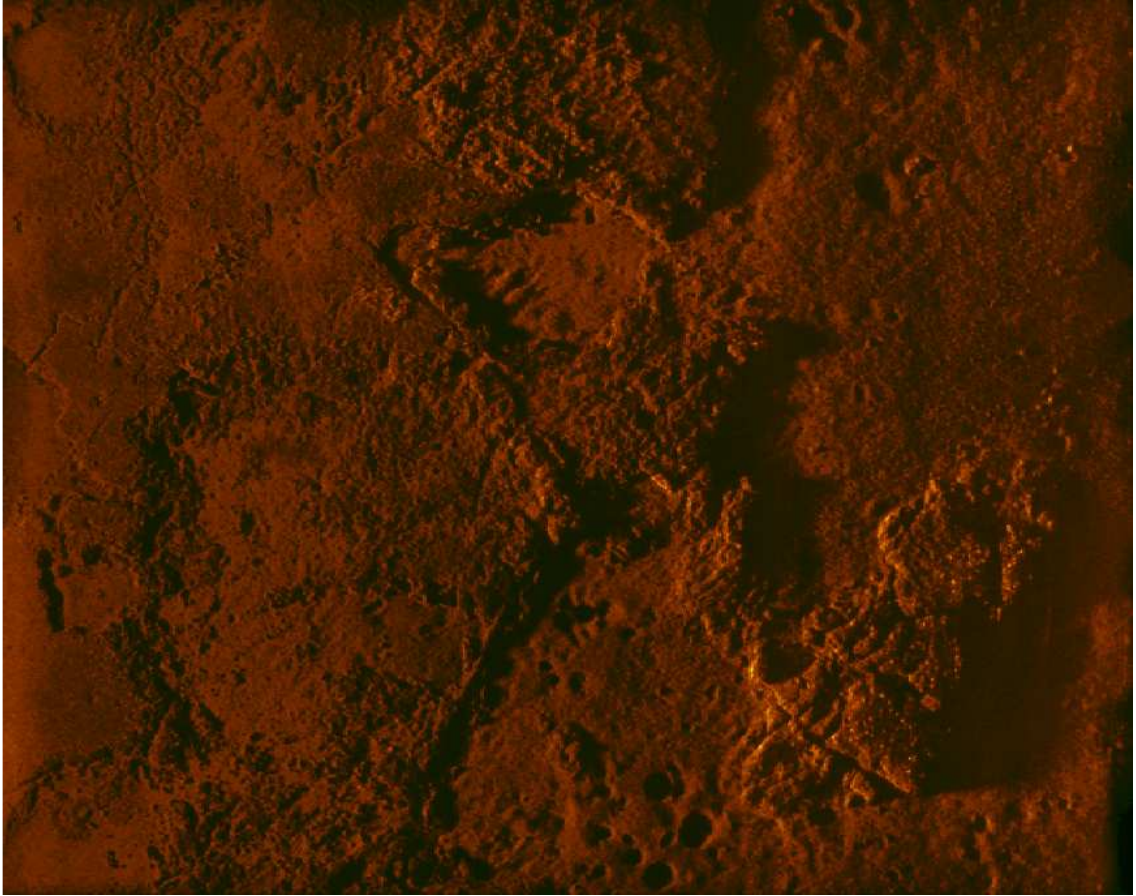}}
	\hfill
	\subfloat[GMRF\label{s1b}]{%
		\includegraphics[trim={1cm 1cm 2cm 1cm},clip,width=0.25\linewidth]{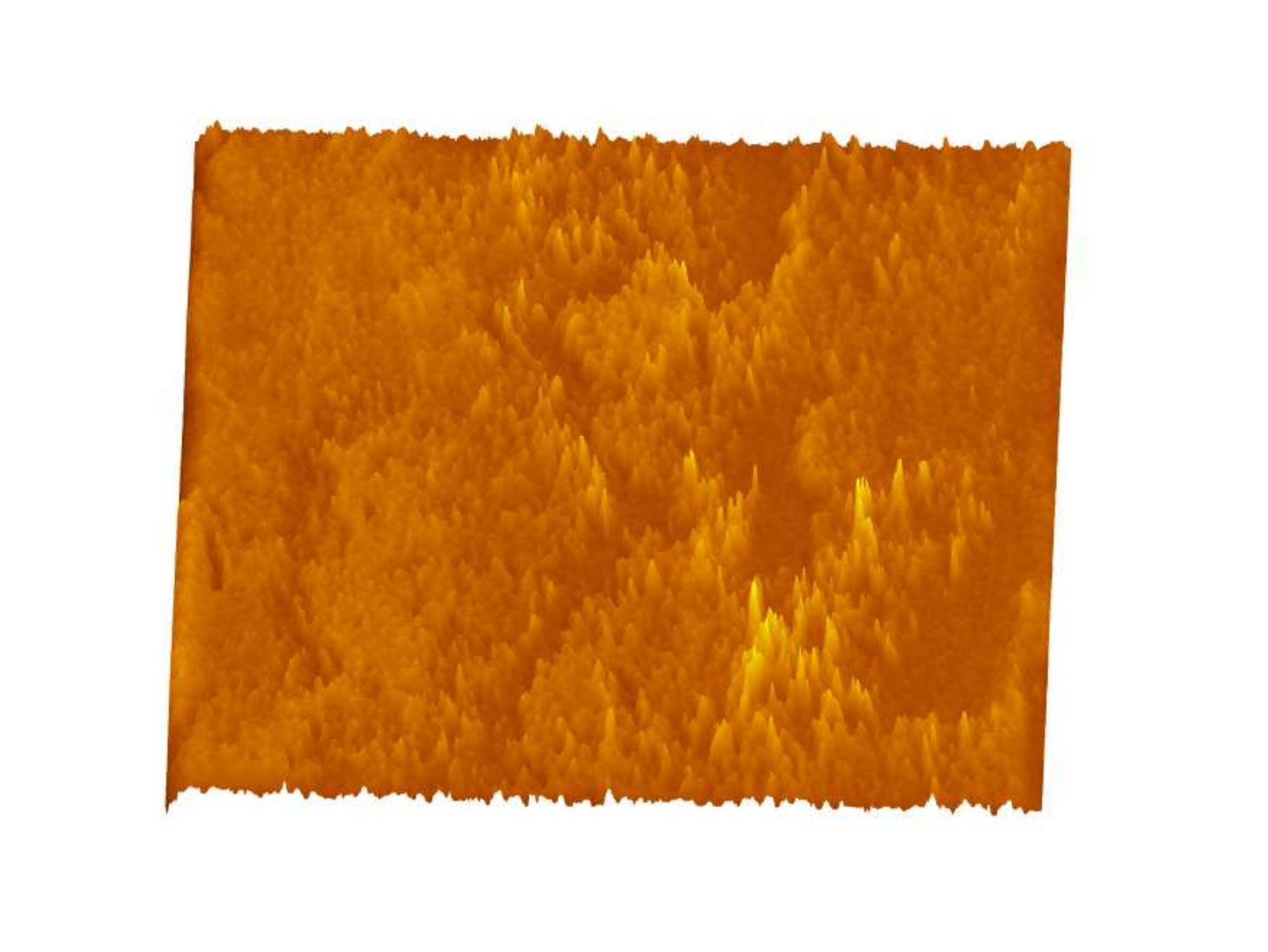}}
	\hfill
	\subfloat[UNet\label{s1c}]{%
		\includegraphics[trim={1cm 1cm 2cm 1cm},clip,width=0.25\linewidth]{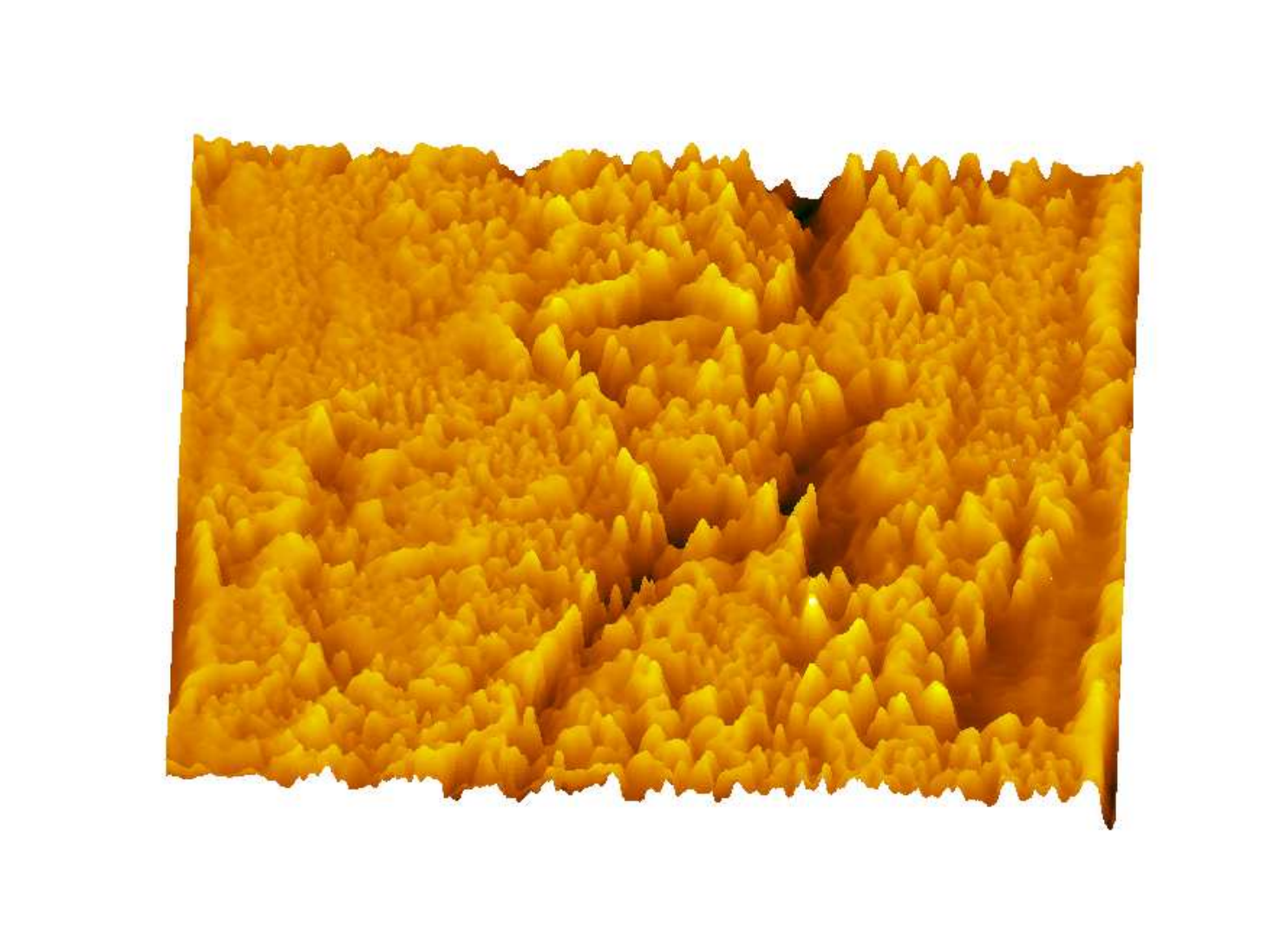}}
	\hfill
	\subfloat[UNet-opt-ftc\label{s1d}]{%
		\includegraphics[trim={1cm 1cm 2cm 1cm},clip,width=0.25\linewidth]{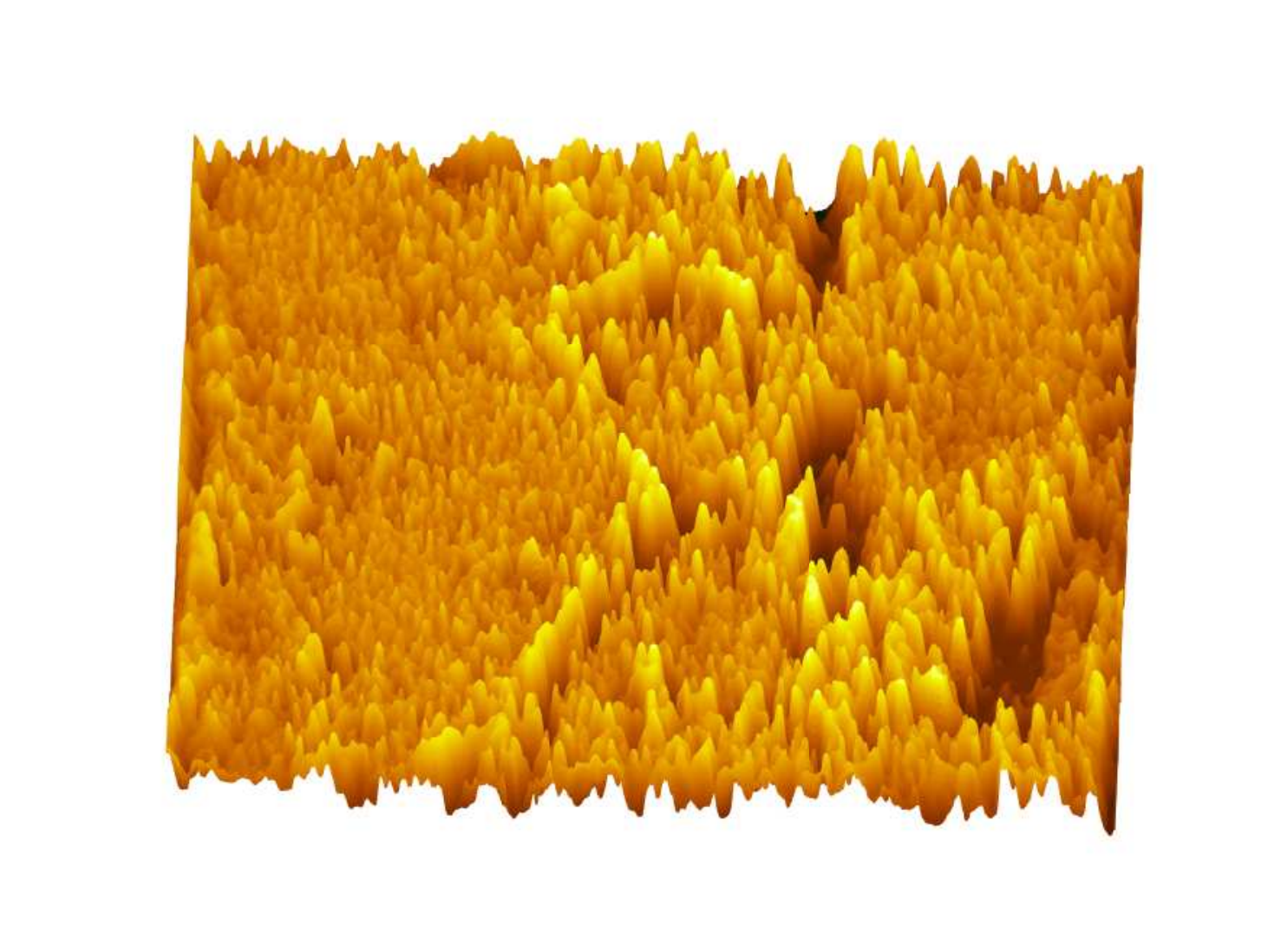}} \\
	
	\subfloat[$V_2$\label{s4a}]{%
		\includegraphics[trim={1cm 2.75cm 3.75cm 1cm},clip,width=0.225\linewidth]{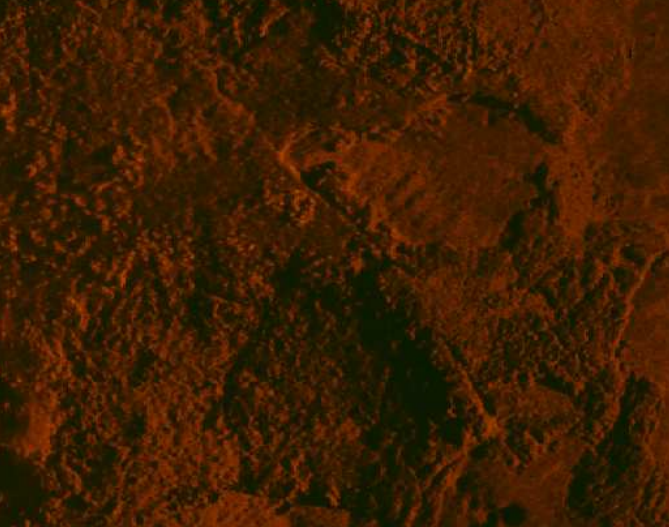}}
	\hfill
	\subfloat[GMRF\label{s4b}]{%
		\includegraphics[trim={1cm 3.75cm 4.75cm 1cm},clip,width=0.25\linewidth]{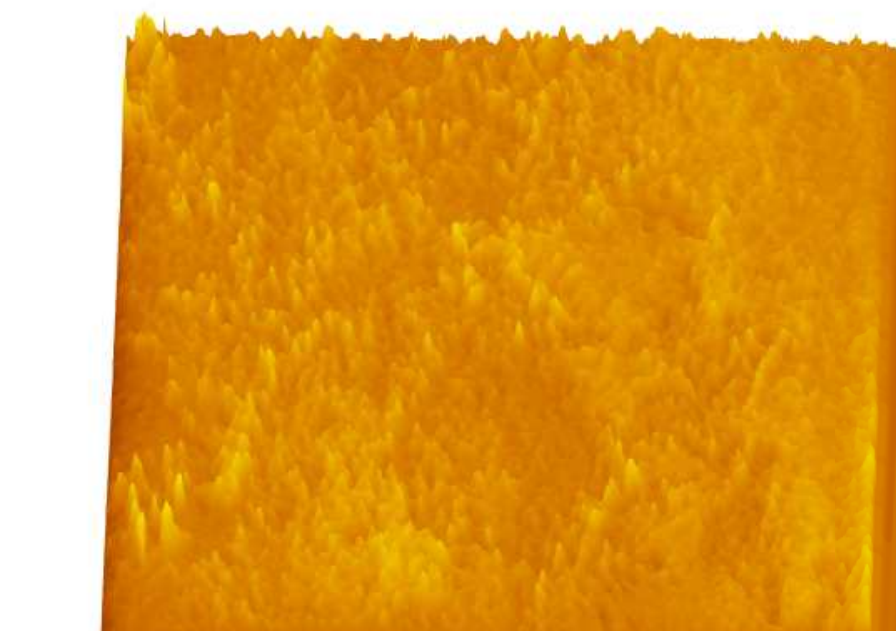}}
	\hfill
	\subfloat[UNet\label{s4c}]{%
		\includegraphics[trim={1cm 3.75cm 4.75cm 1cm},clip,width=0.25\linewidth]{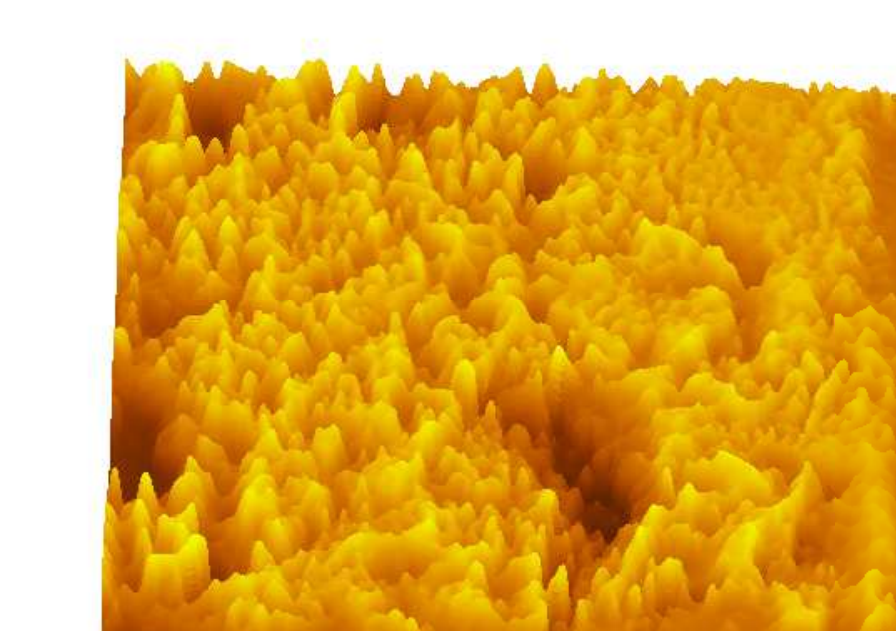}}
	\hfill
	\subfloat[UNet-opt-ftc\label{s4d}]{%
		\includegraphics[trim={1cm 3.75cm 4.75cm 1cm},clip,width=0.25\linewidth]{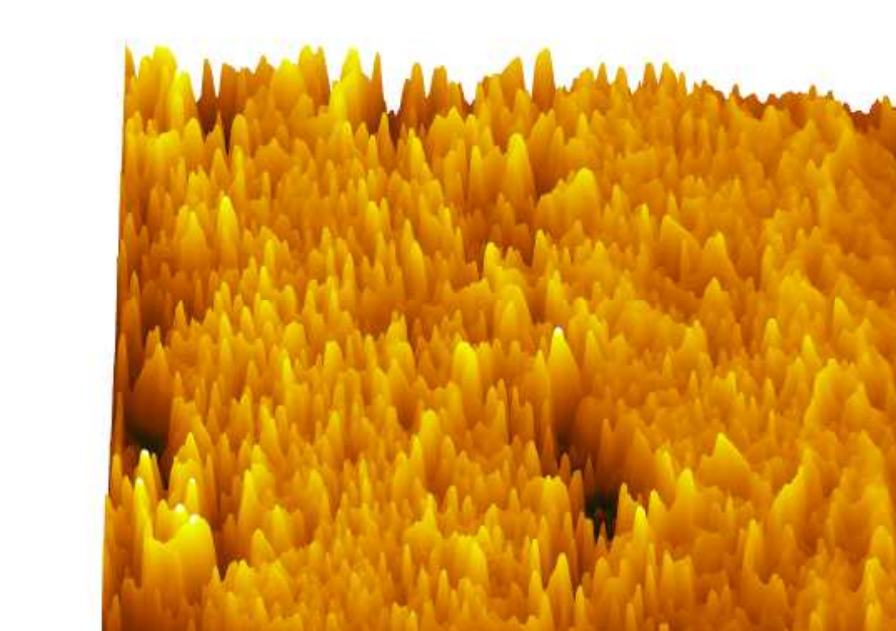}} \\
	\caption{Examples of seabed relief estimation on a scene containing rocky and flat textures. Views $V_1$ and $V_2$ refer to a pair of coregistered intensity images (a,e). The GMRF model (b,f) and standard UNet oversmooth the seabed relief maps (c,g). The UNet-opt-ftc model produces a seabed relief map that is not oversmoothed and more similar across the looks (d,h).}
	\label{fig:scene2} 
\end{figure}

\paragraph{Hand-aligned Multiaspect SAS Data}
Our hand-aligned multiaspect SAS dataset consists of five pairs of coregistered SAS images. There is no ground-truth seabed relief information for this dataset. We applied our fine-tuned models, as well as the standard UNet, and GMRF models to each intensity image and measured the $L_1$ error between coregistered pairs. The average and standard deviation across the dataset is used to evaluate the performance in comparison to raw intensity values (Table \ref{tab:testMASAS}). Qualitative examples of the predicted seabed relief from UNet are shown in \Cref{fig:scene1,fig:scene2}.

\begin{figure}[h]
	\centering
	\subfloat[$V_1$\label{sr1a}]{%
		\includegraphics[trim={1cm 1cm 1cm 1cm},clip,width=0.225\linewidth,height=1.5cm]{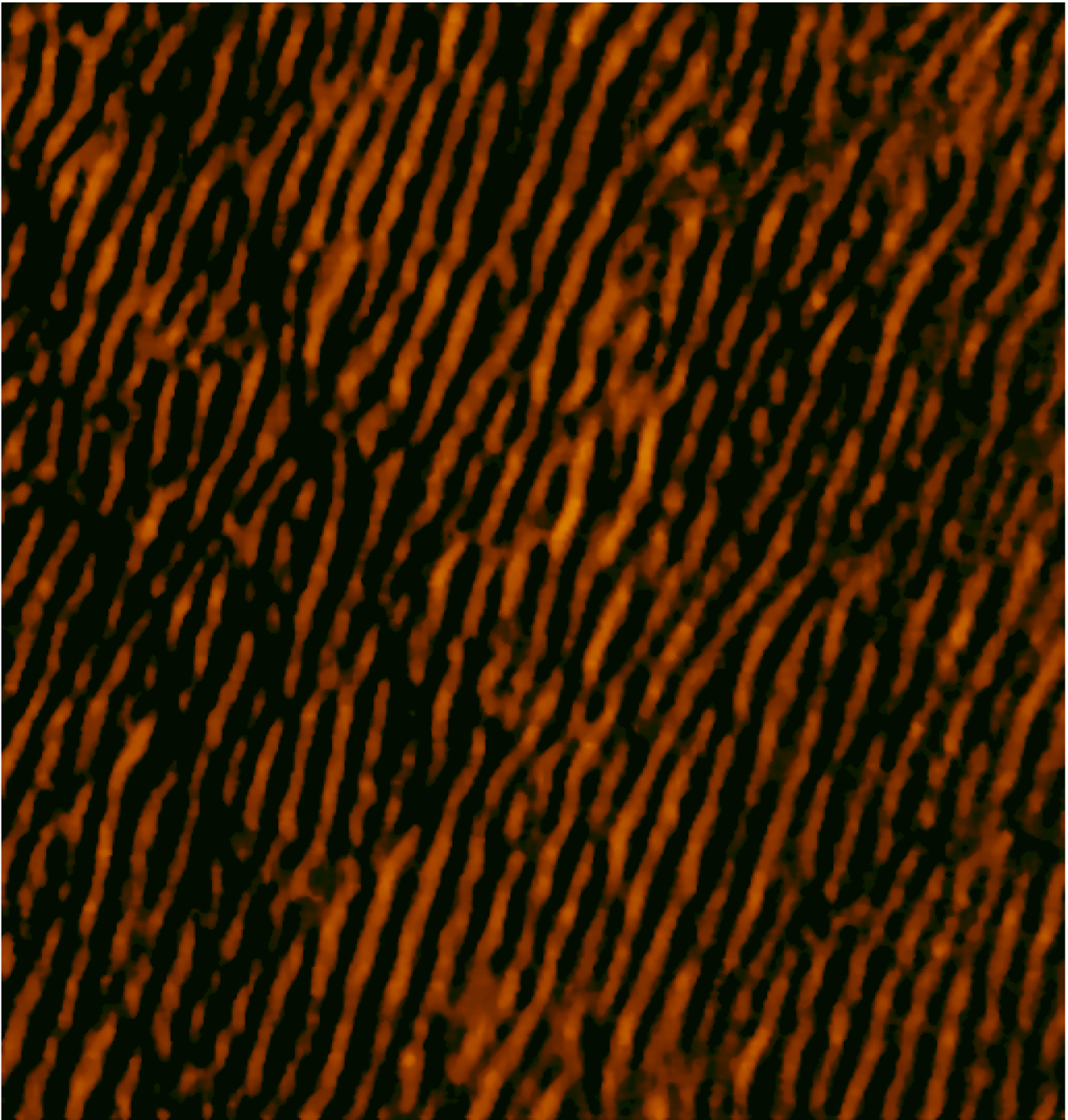}}
	\hfill
	\subfloat[GMRF\label{sr1b}]{%
		\includegraphics[trim={1cm 1cm 2cm 1cm},clip,width=0.25\linewidth]{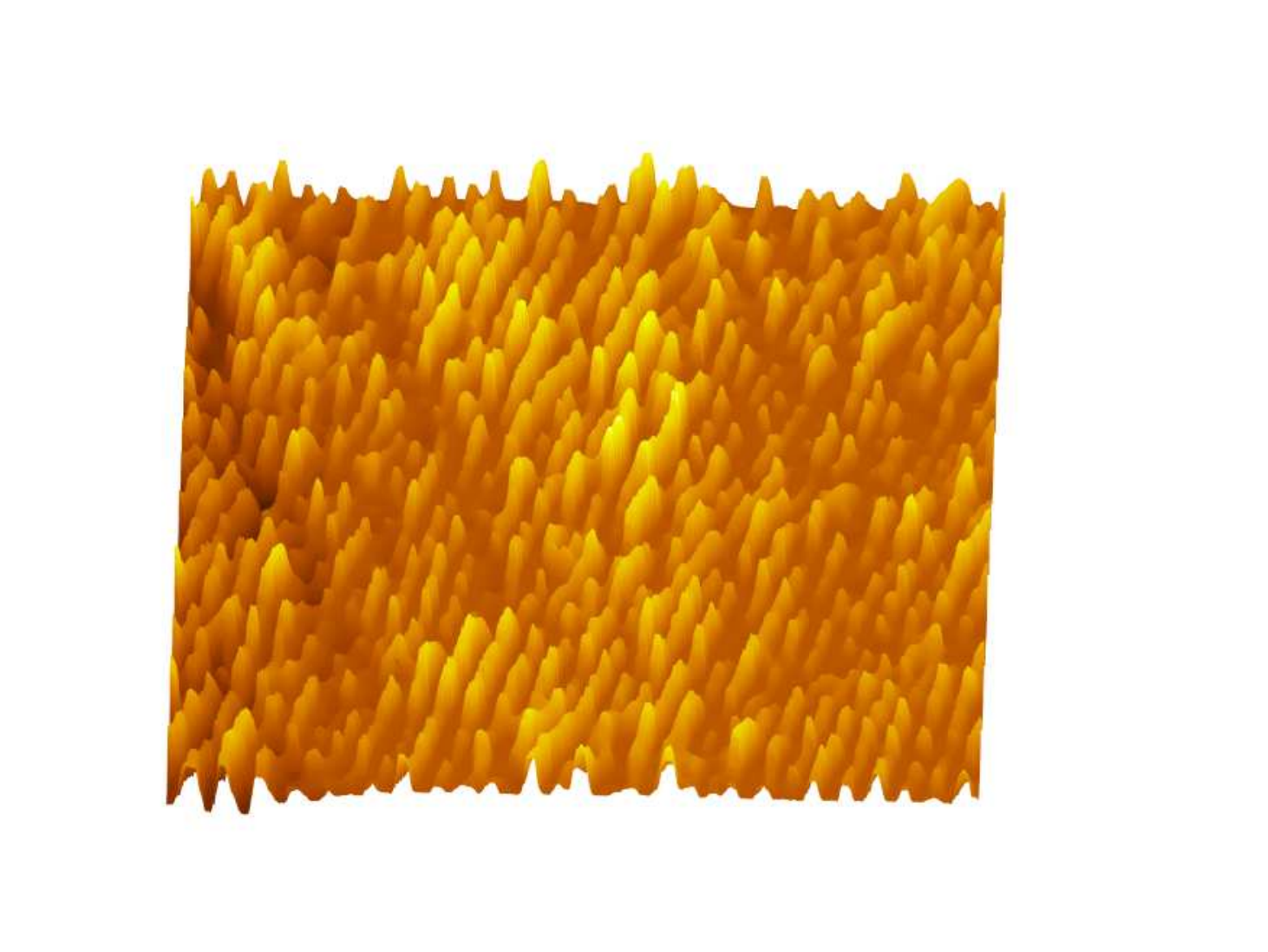}}
	\hfill
	\subfloat[UNet\label{sr1c}]{%
		\includegraphics[trim={1cm 1cm 2cm 1cm},clip,width=0.25\linewidth]{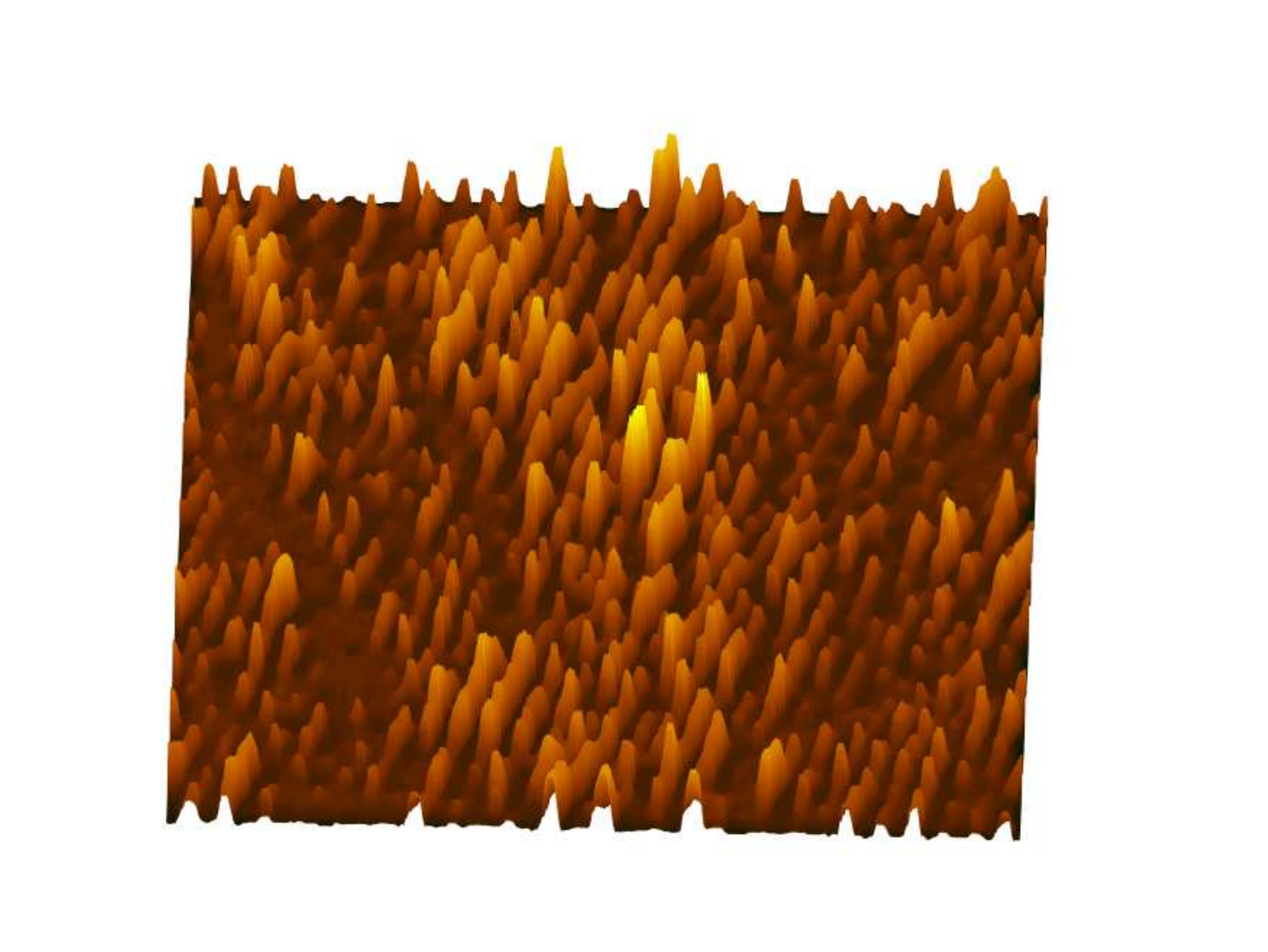}}
	\hfill
	\subfloat[UNet-opt-ftc\label{sr1d}]{%
		\includegraphics[trim={1cm 1cm 2cm 1cm},clip,width=0.25\linewidth]{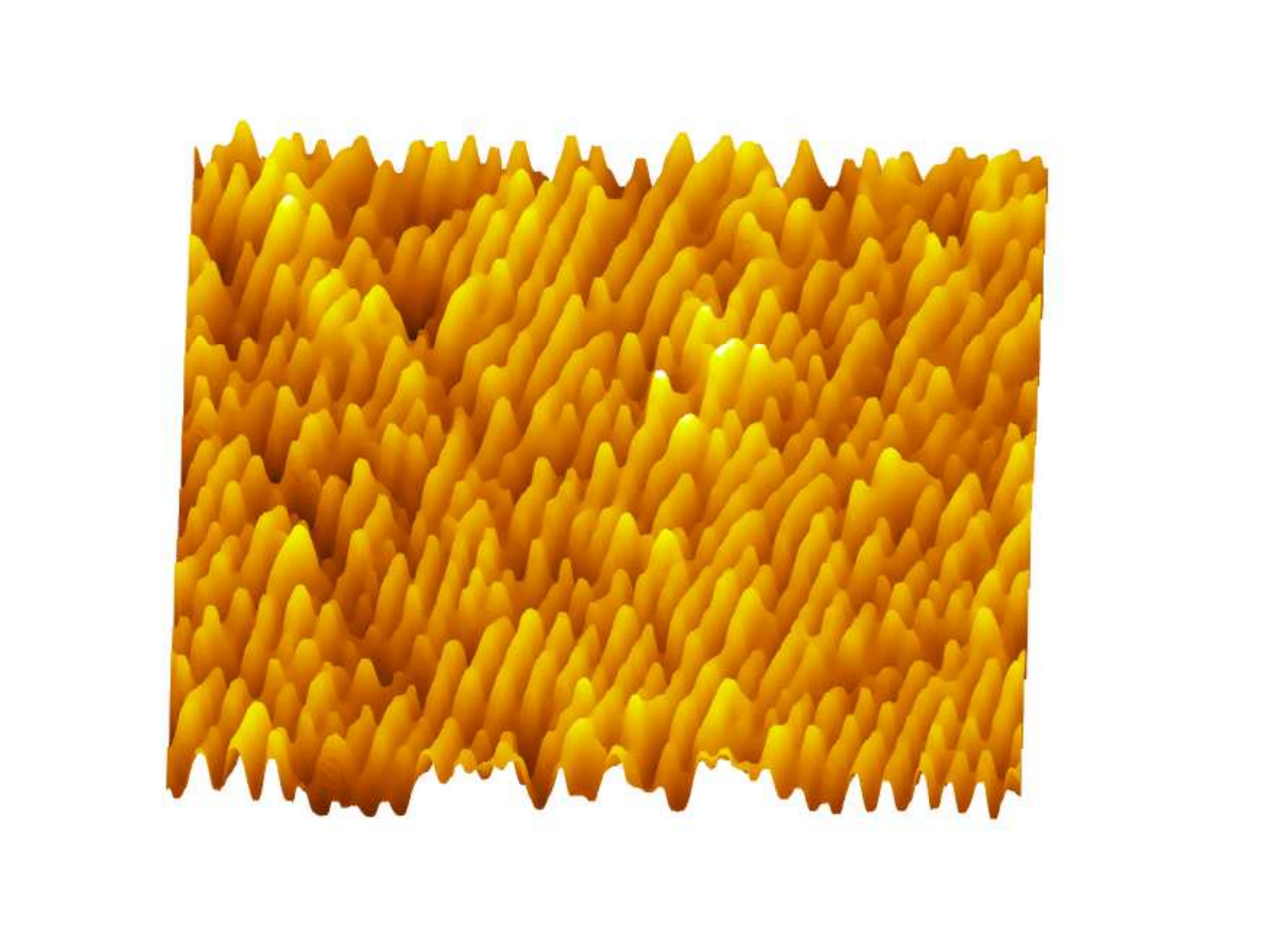}} \\
	
	\subfloat[$V_2$\label{sr4a}]{%
		\includegraphics[trim={1cm 1cm 2cm 1cm},clip,width=0.225\linewidth,height=1.5cm]{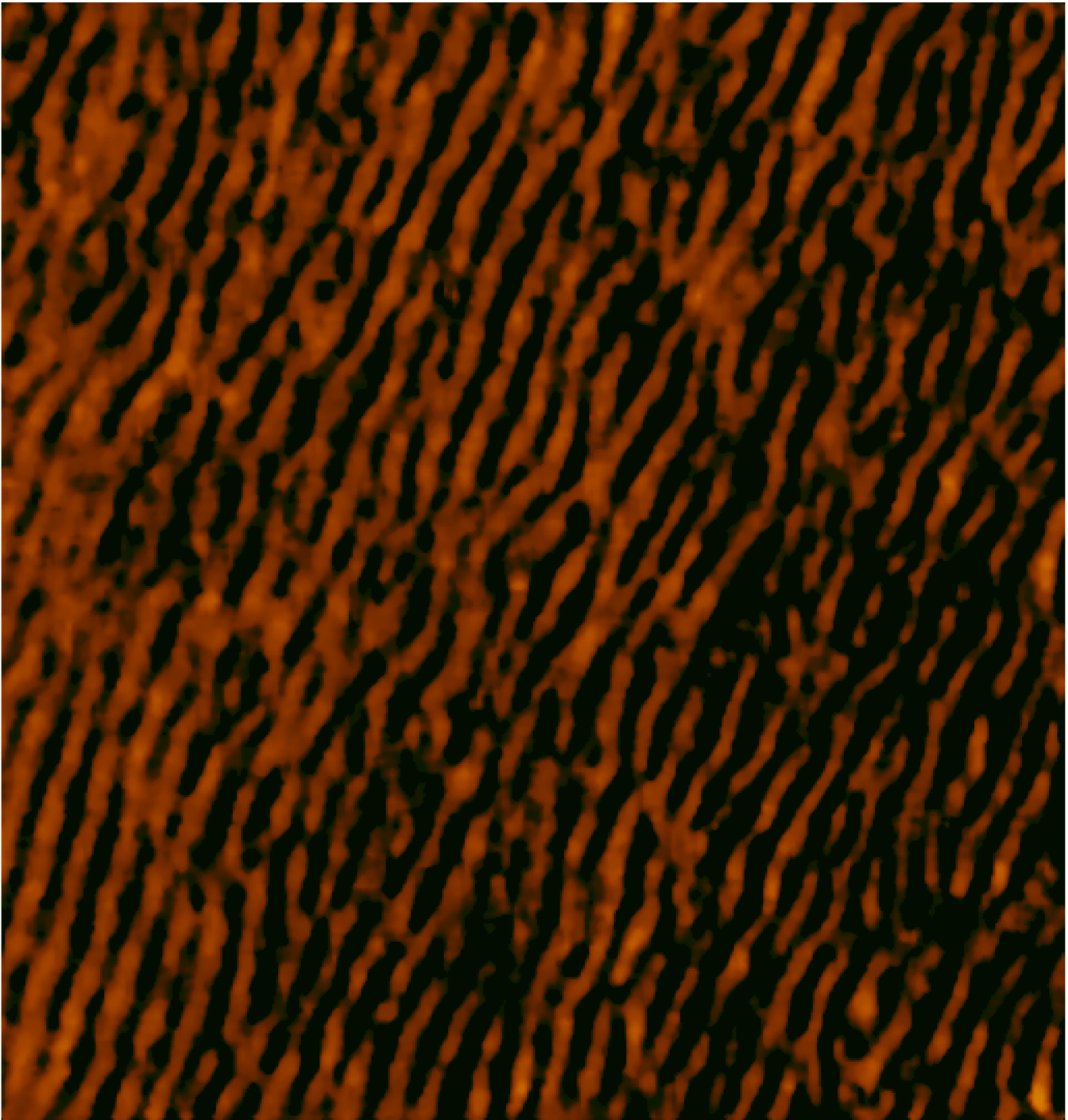}}
	\hfill
	\subfloat[GMRF\label{sr4b}]{%
		\includegraphics[trim={1cm 1cm 2cm 1cm},clip,width=0.25\linewidth]{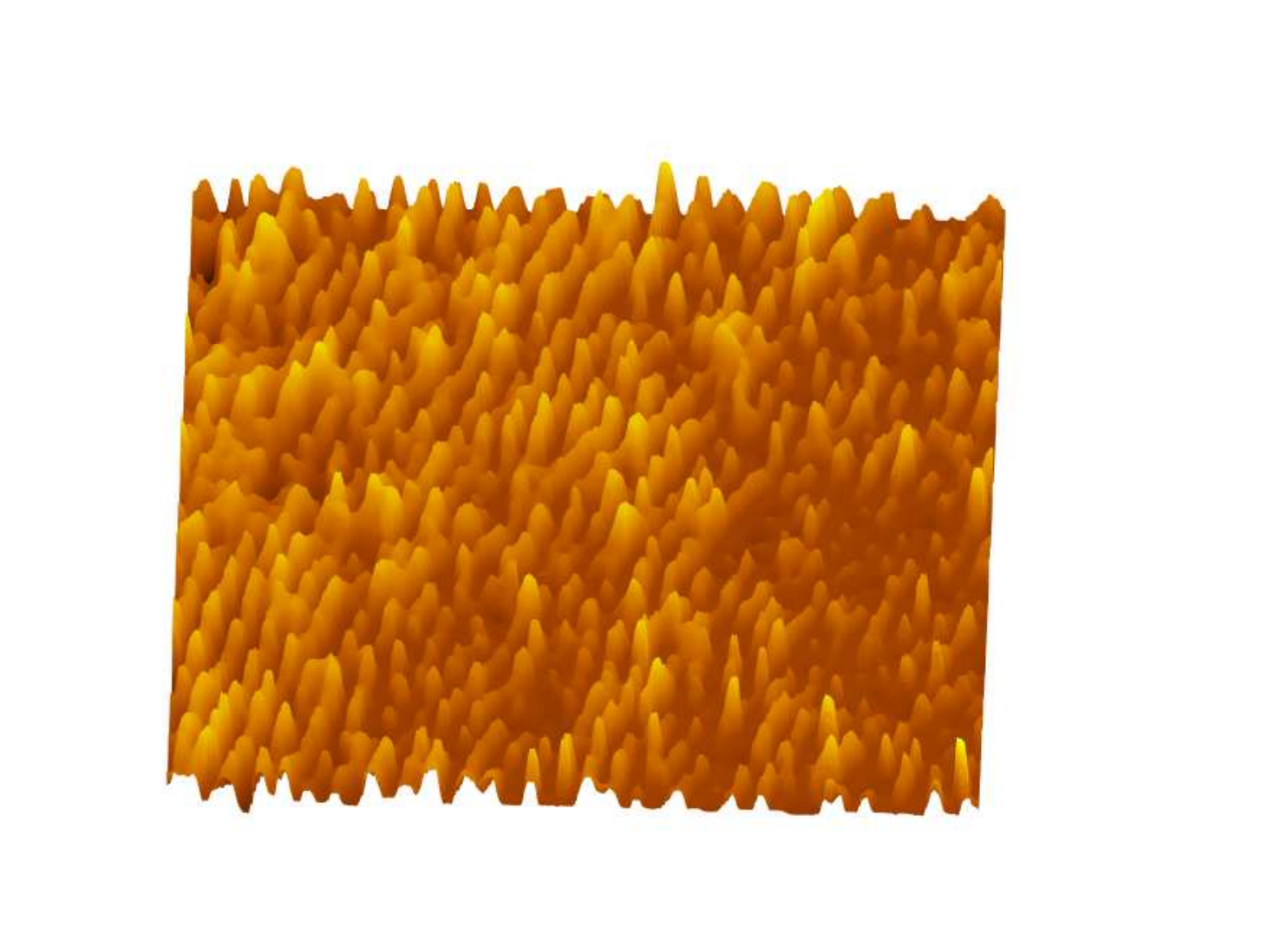}}
	\hfill
	\subfloat[UNet\label{sr4c}]{%
		\includegraphics[trim={1cm 1cm 2cm 1cm},clip,width=0.25\linewidth]{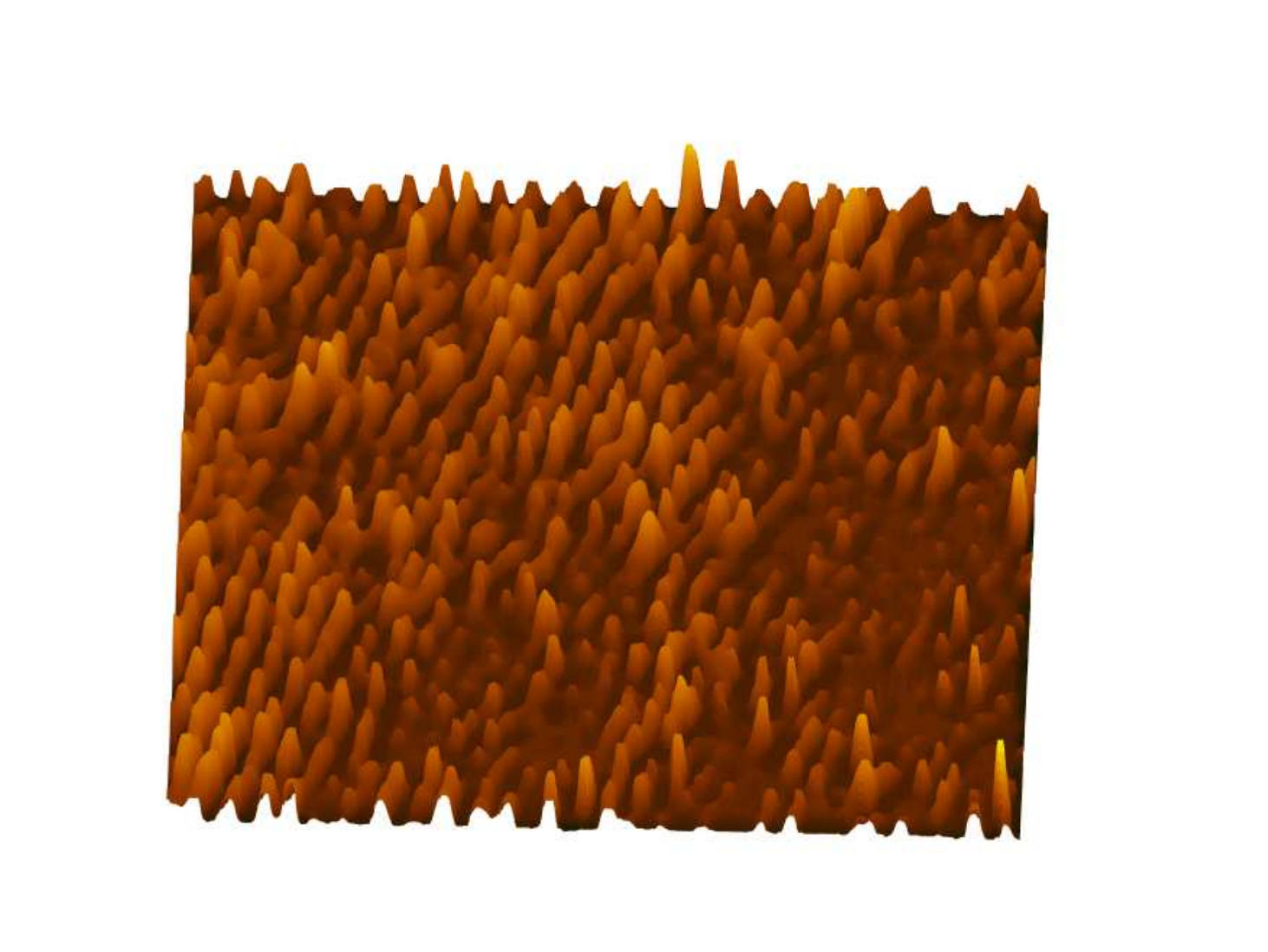}}
	\hfill
	\subfloat[UNet-opt-ftc\label{sr4d}]{%
		\includegraphics[trim={1cm 1cm 2cm 1cm},clip,width=0.25\linewidth]{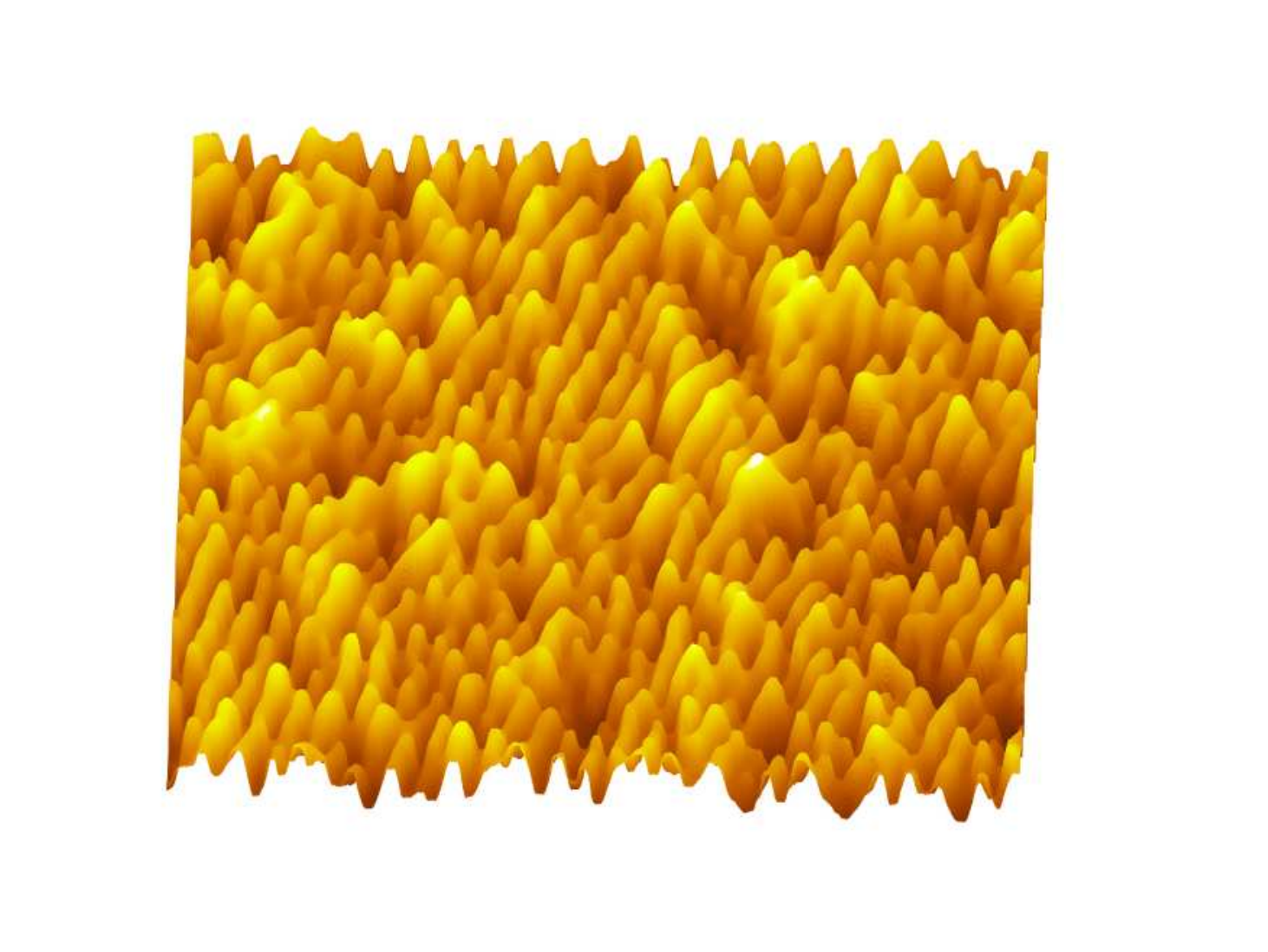}} \\
	\caption{Examples of seabed relief estimation on a pair of coregistered sand-ripple images. Views $V_1$ and $V_2$ refer to a pair of coregistered intensity images (a,e). The GMRF model (b,f) and standard UNet (c,g) produce noisy seabed relief maps. The UNet-opt-ftc model produces a seabed relief map that is not oversmoothed and more similar across the looks (d,h).}
	\label{fig:sceneSR} 
\end{figure}

The UNet-opt-ftc model produces more similar seabed relief estimations of sand-ripple captured from multiple aspects than the GMRF or standard UNet (Figure \ref{fig:sceneSR}). The GMRF and UNet estimates appear noisy in comparison to the UNet-opt-ftc. We also evaluated the methods on an interesting scene containing two looks of a flat, rough, and rocky textures (Figure \ref{fig:scene1}). Seabed relief maps produced by both the UNet and GMRF are mostly oversmoothed. Meanwhile, the UNet-opt-ftc model produces seabed relief maps that are the most similar and not overly smooth. In a rocky and flat scene (Figure \ref{fig:scene2}), the GMRF and UNet estimates are oversmoothed compared to the UNet-opt-ftc.

Regardless of the method used, each model produces estimated seabed relief with less error between looks than intensity (Table \ref{tab:testMASAS}). This supports using an estimated seabed relief space rather than raw intensity when relying on multiple looks of SAS data. Each variant of the UNet-opt model surpasses the performance of the GMRF and standard UNet. This can be better understood due to the UNet and GMRF oversmoothing the real SAS data in comparison to the UNet-opt-ftc (Figure \ref{fig:scene1}). Each fine-tuned model (UNet-opt-ft and UNet-opt-ftc) improves upon the pre-trained (UNet-opt-pt) in terms of the error between looks. This supports using the high-fidelity PoSSM data for fine-tuning the models. Overall, the best model (UNet-opt-ftc) is produced by pre-training on the PISAS data and fine-tuning the output layer on the PoSSM data.    

\begin{figure*}[h]
	\centering
	\subfloat[$\hat{H}_1$\label{er1}]{%
		\includegraphics[trim={1cm 1cm 1cm 1cm},clip,width=0.3\linewidth]{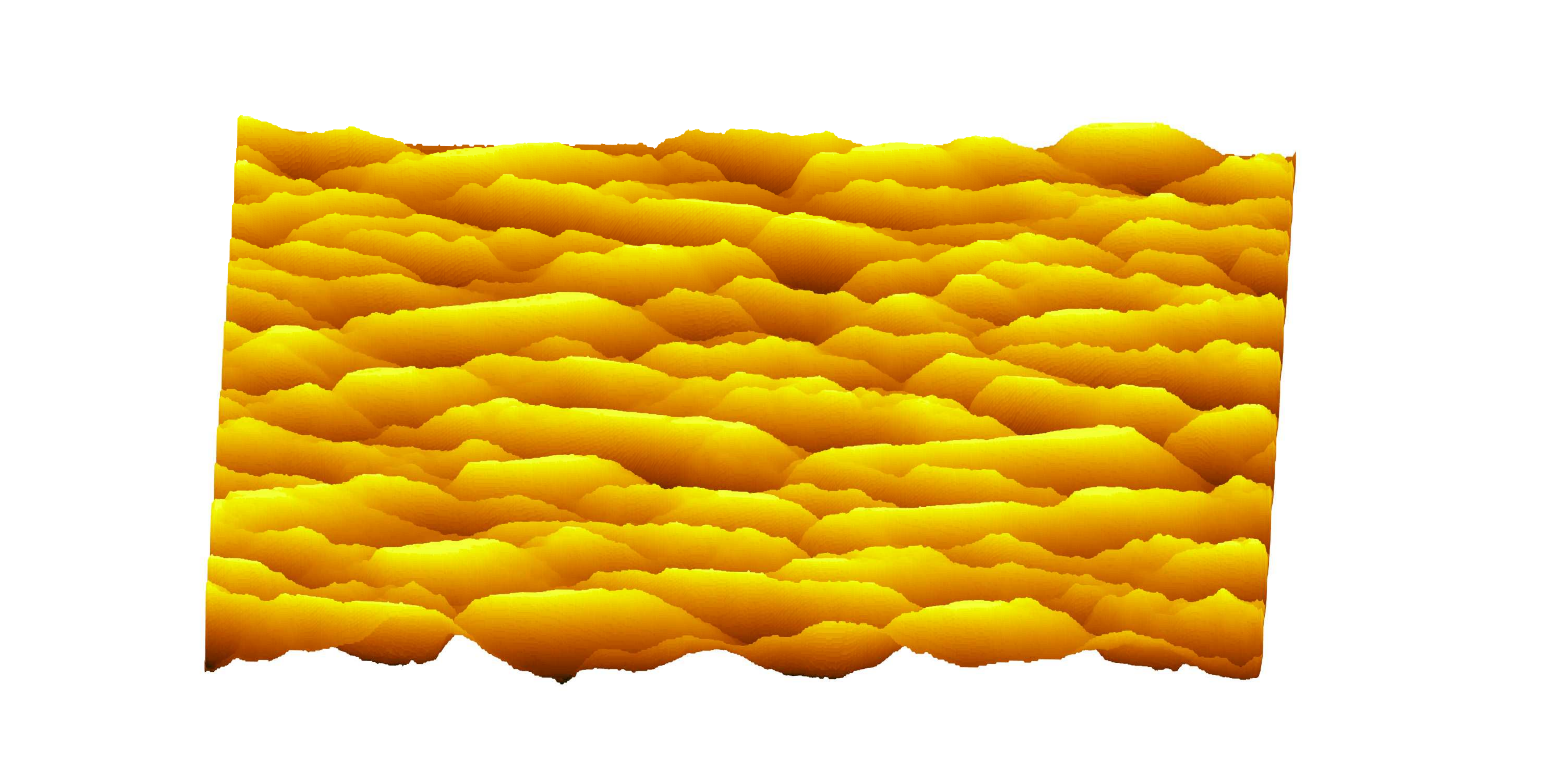}}\hfill
\subfloat[$H_2-H_1$\label{der1}]{%
	\includegraphics[trim={1cm 1cm 1cm 1cm},clip,width=0.3\linewidth]{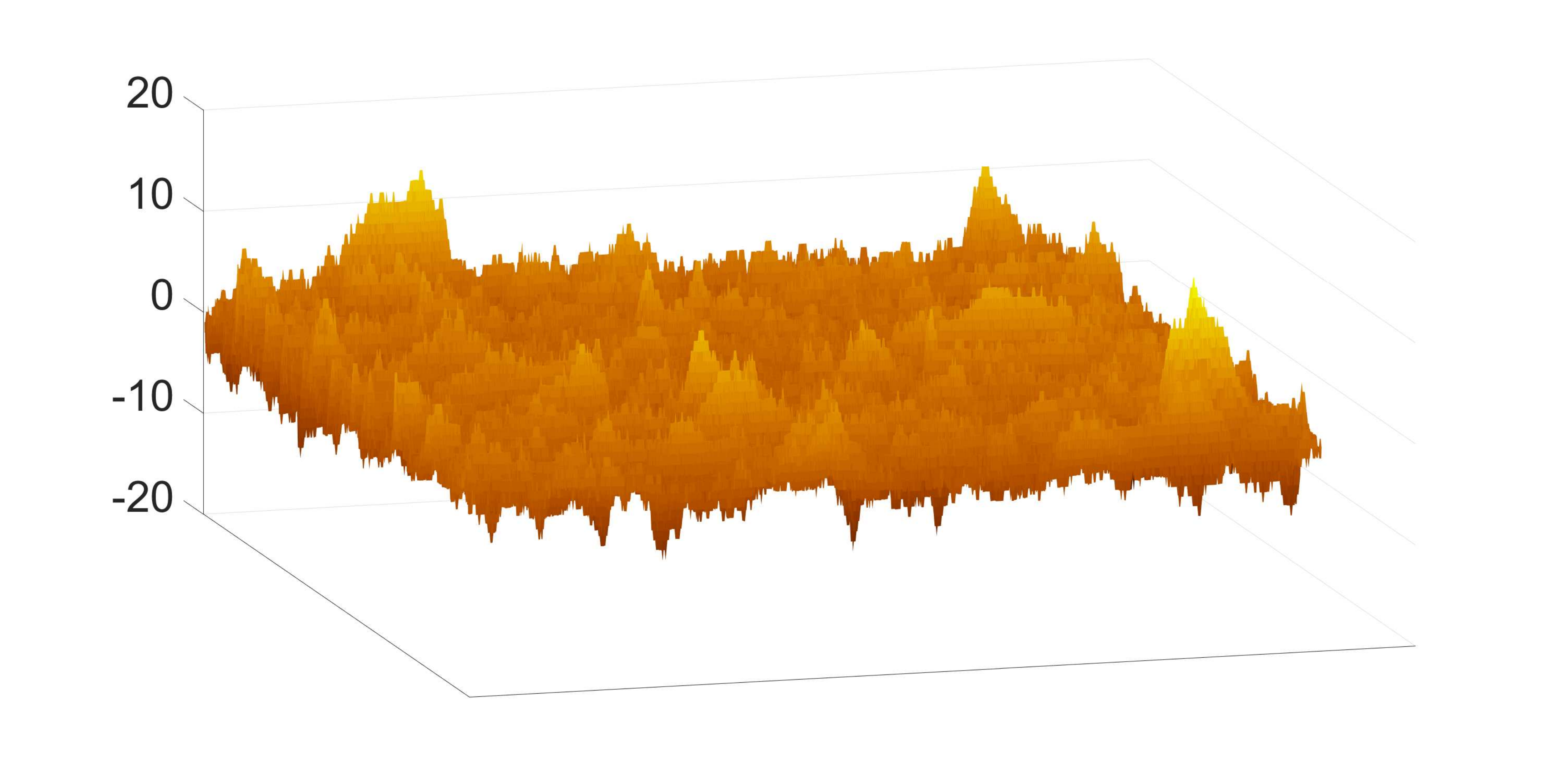}} \hfill
	\subfloat[$\hat{H}_2$\label{er2}]{%
	\includegraphics[trim={1cm 1cm 1cm 1cm},clip,width=0.3\linewidth]{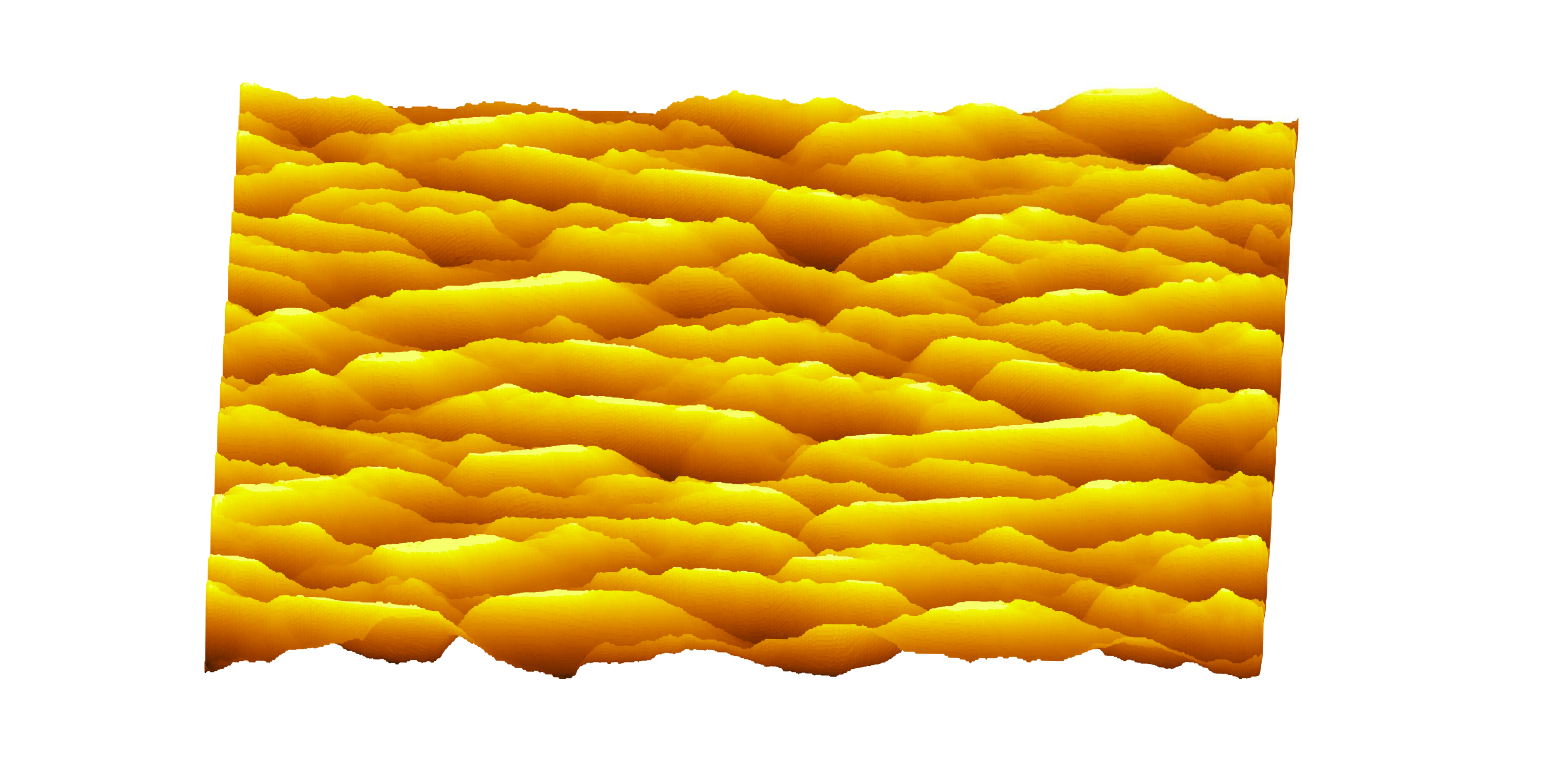}} \\
\subfloat[$\hat{H}_1$\label{er12}]{%
	\includegraphics[trim={1cm 1cm 1cm 1cm},clip,width=0.3\linewidth]{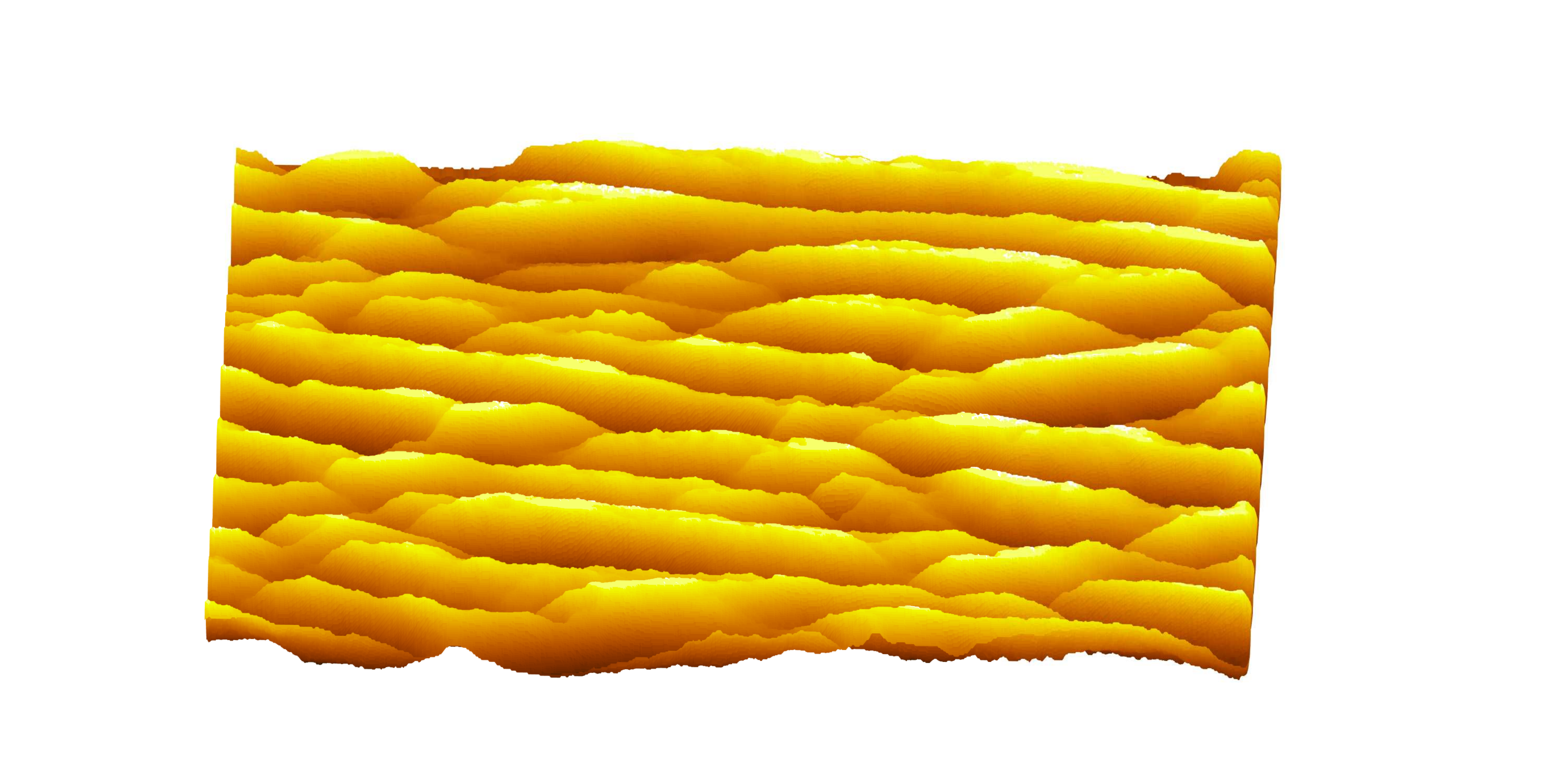}}\hfill
\subfloat[$H_2-H_1$\label{der12}]{%
	\includegraphics[trim={1cm 1cm 1cm 1cm},clip,width=0.3\linewidth]{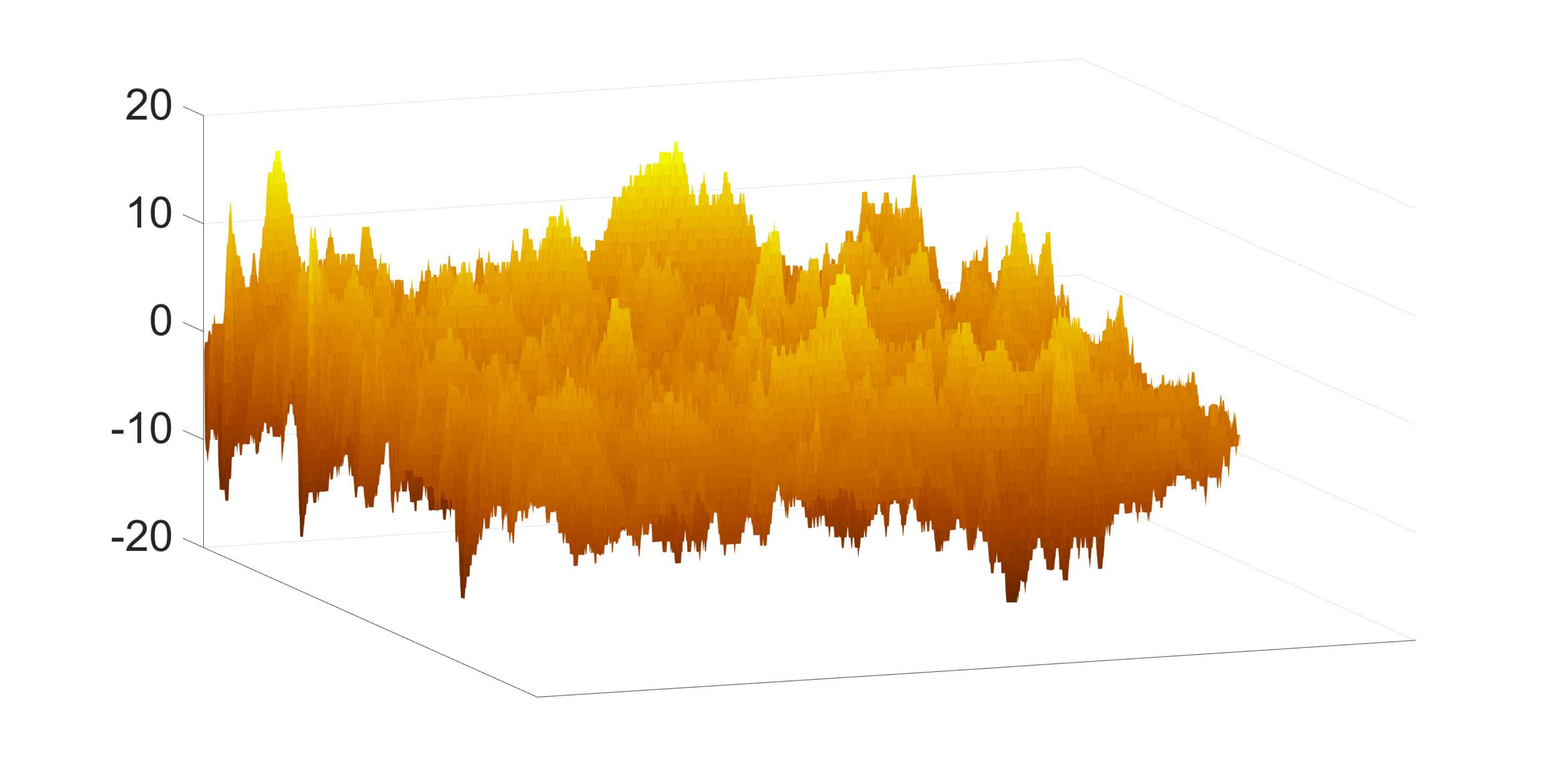}} \hfill
\subfloat[$\hat{H}_2$\label{er22}]{%
	\includegraphics[trim={1cm 1cm 1cm 1cm},clip,width=0.3\linewidth]{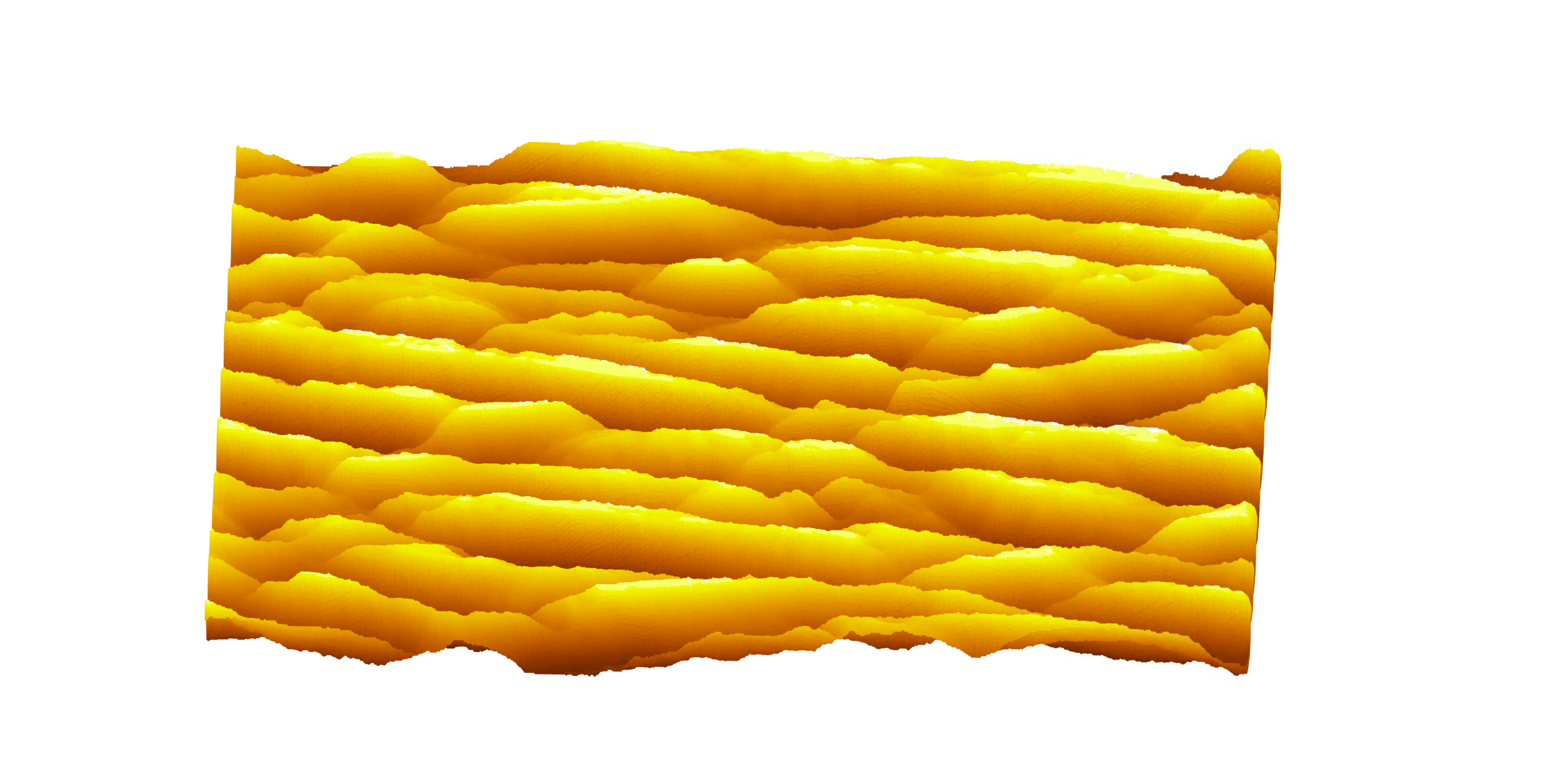}}\\
\subfloat[$\hat{H}_1$\label{er3}]{%
	\includegraphics[trim={1cm 1cm 1cm 1cm},clip,width=0.3\linewidth]{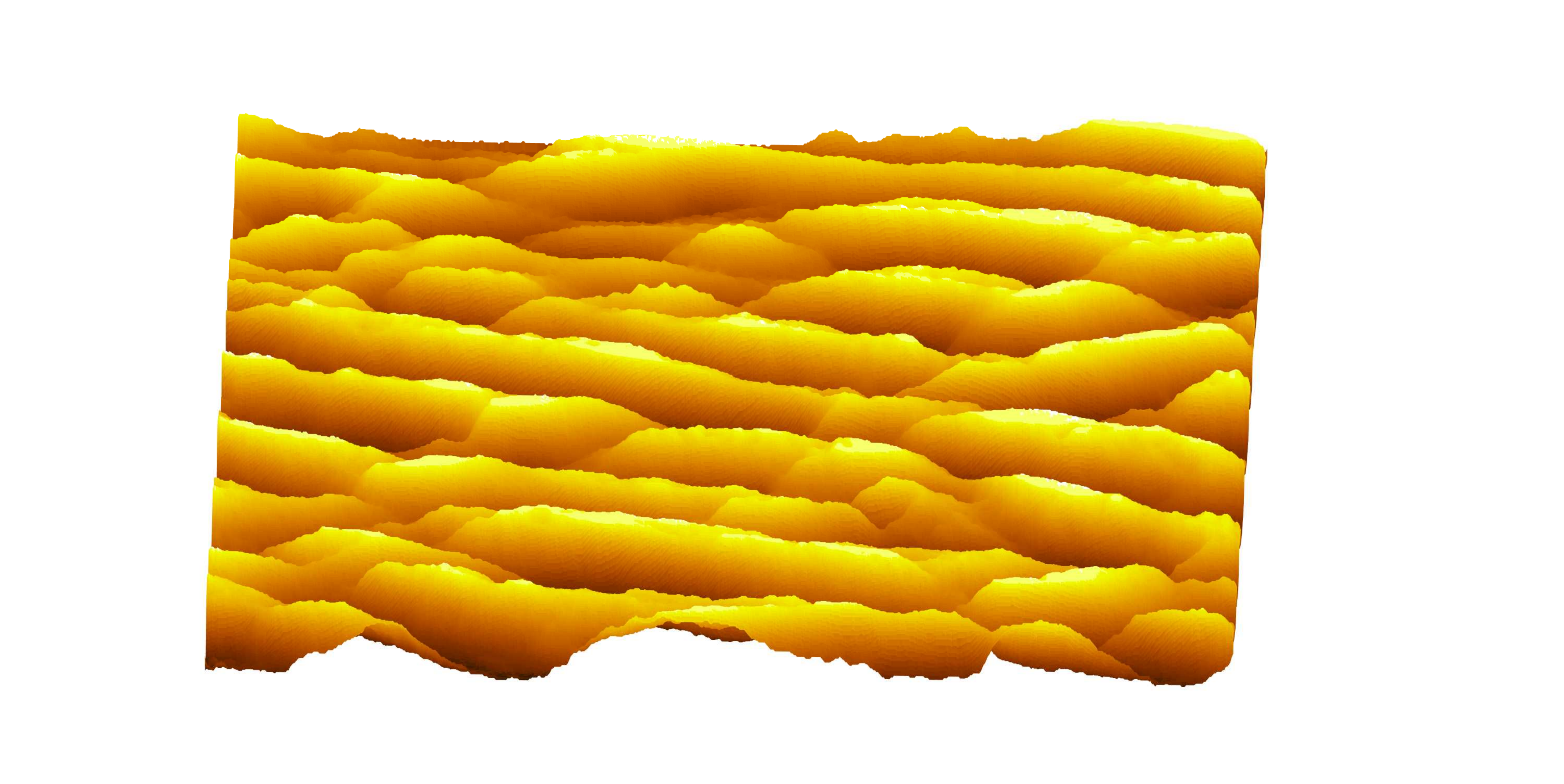}}\hfill
\subfloat[$H_2-H_1$\label{der22}]{%
	\includegraphics[trim={1cm 1cm 1cm 1cm},clip,width=0.3\linewidth]{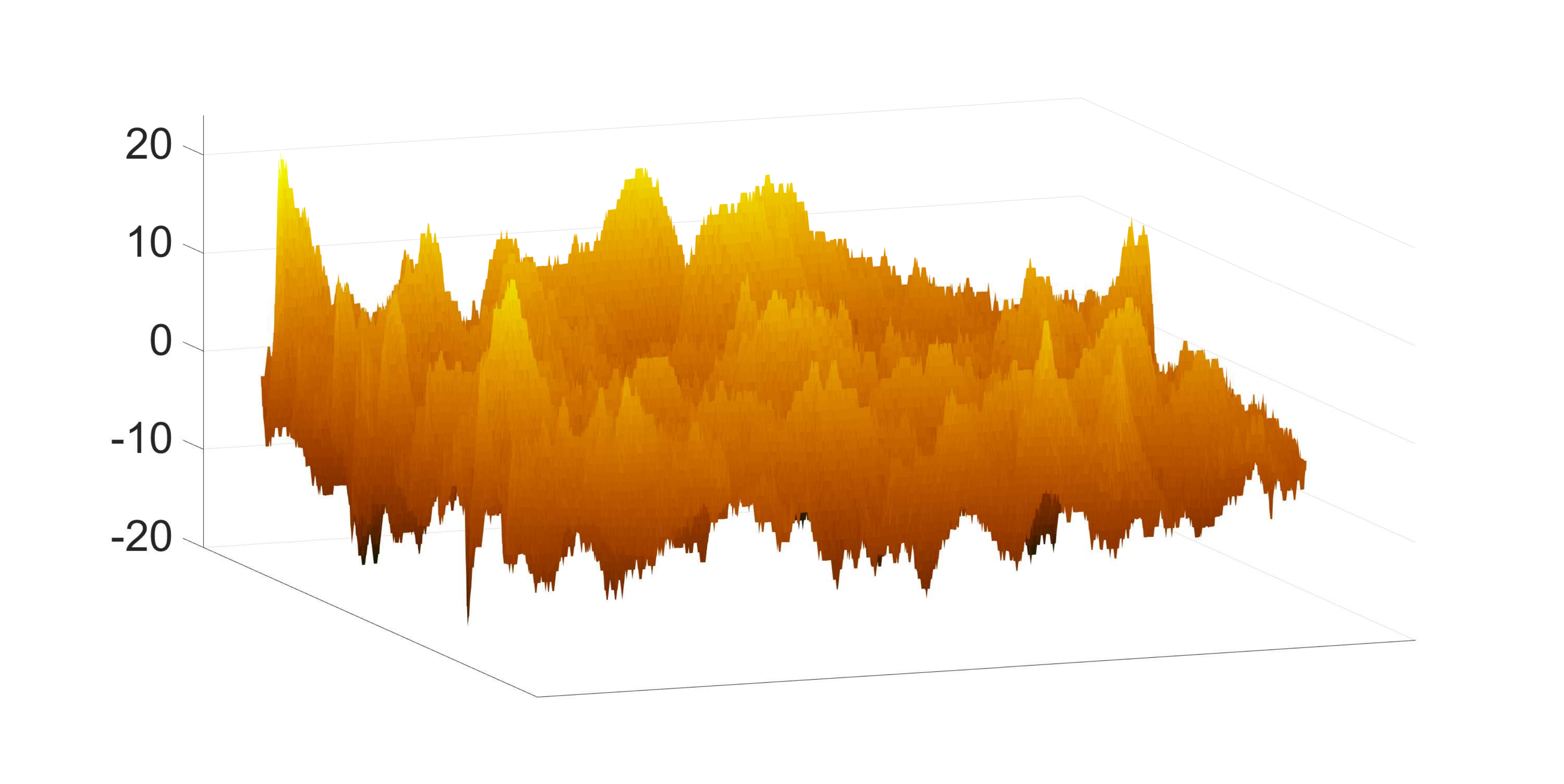}} \hfill
\subfloat[$\hat{H}_2$\label{er32}]{%
	\includegraphics[trim={1cm 1cm 1cm 1cm},clip,width=0.3\linewidth]{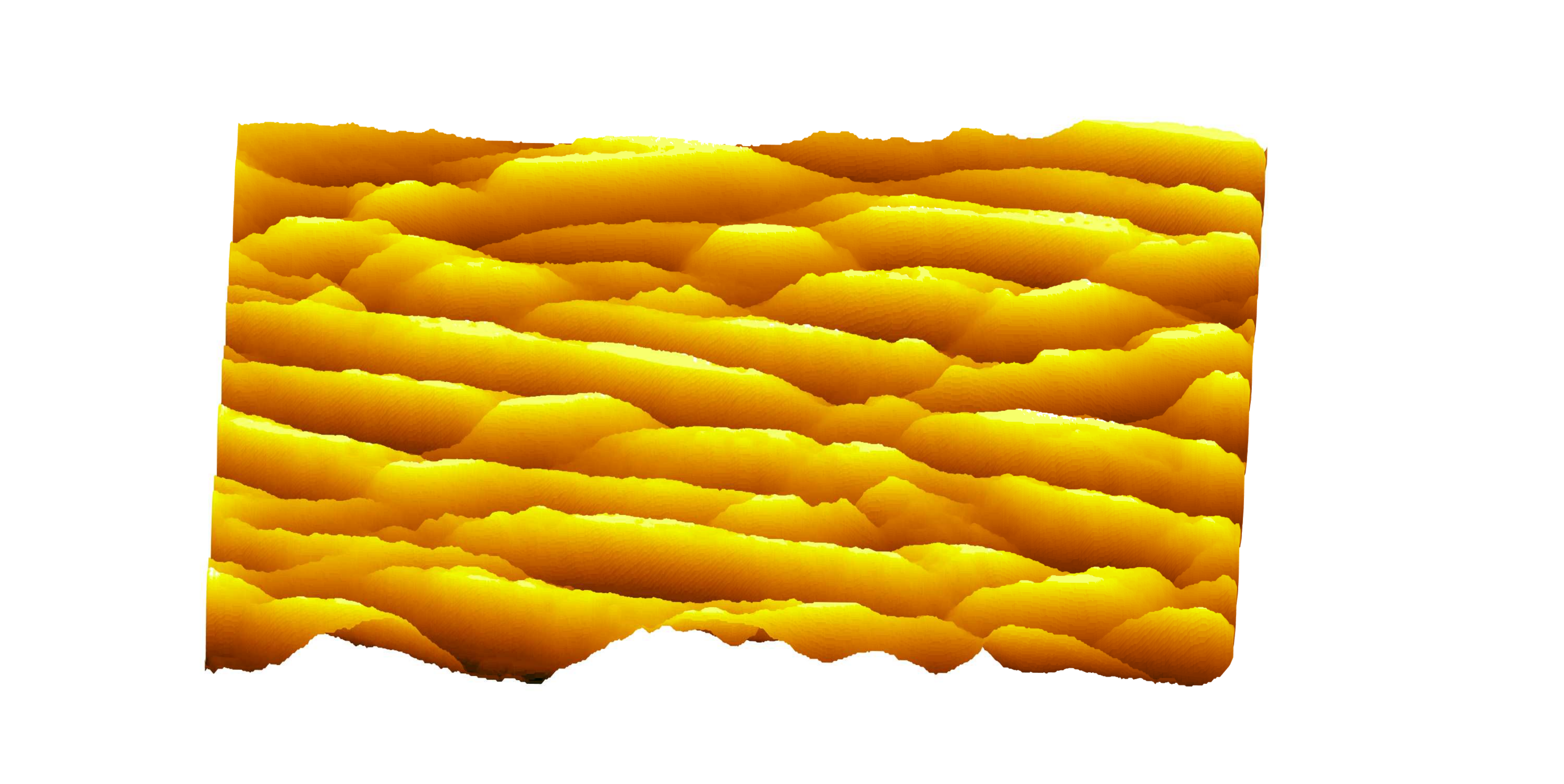}}\\
\subfloat[$\hat{H}_1$\label{er14}]{%
	\includegraphics[trim={1cm 1cm 1cm 1cm},clip,width=0.3\linewidth]{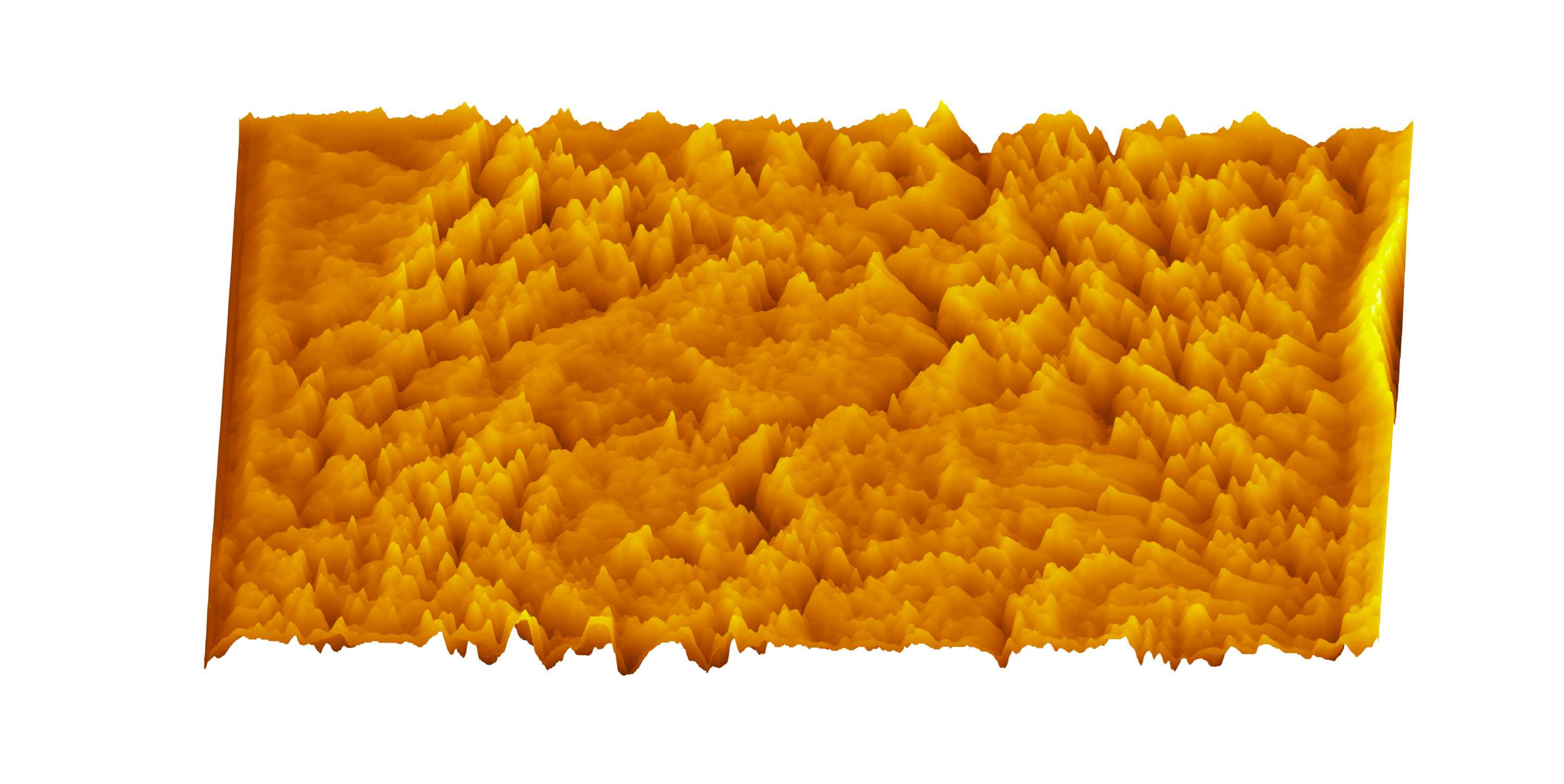}}\hfill
\subfloat[$H_2-H_1$\label{der14}]{%
	\includegraphics[trim={1cm 1cm 1cm 1cm},clip,width=0.3\linewidth]{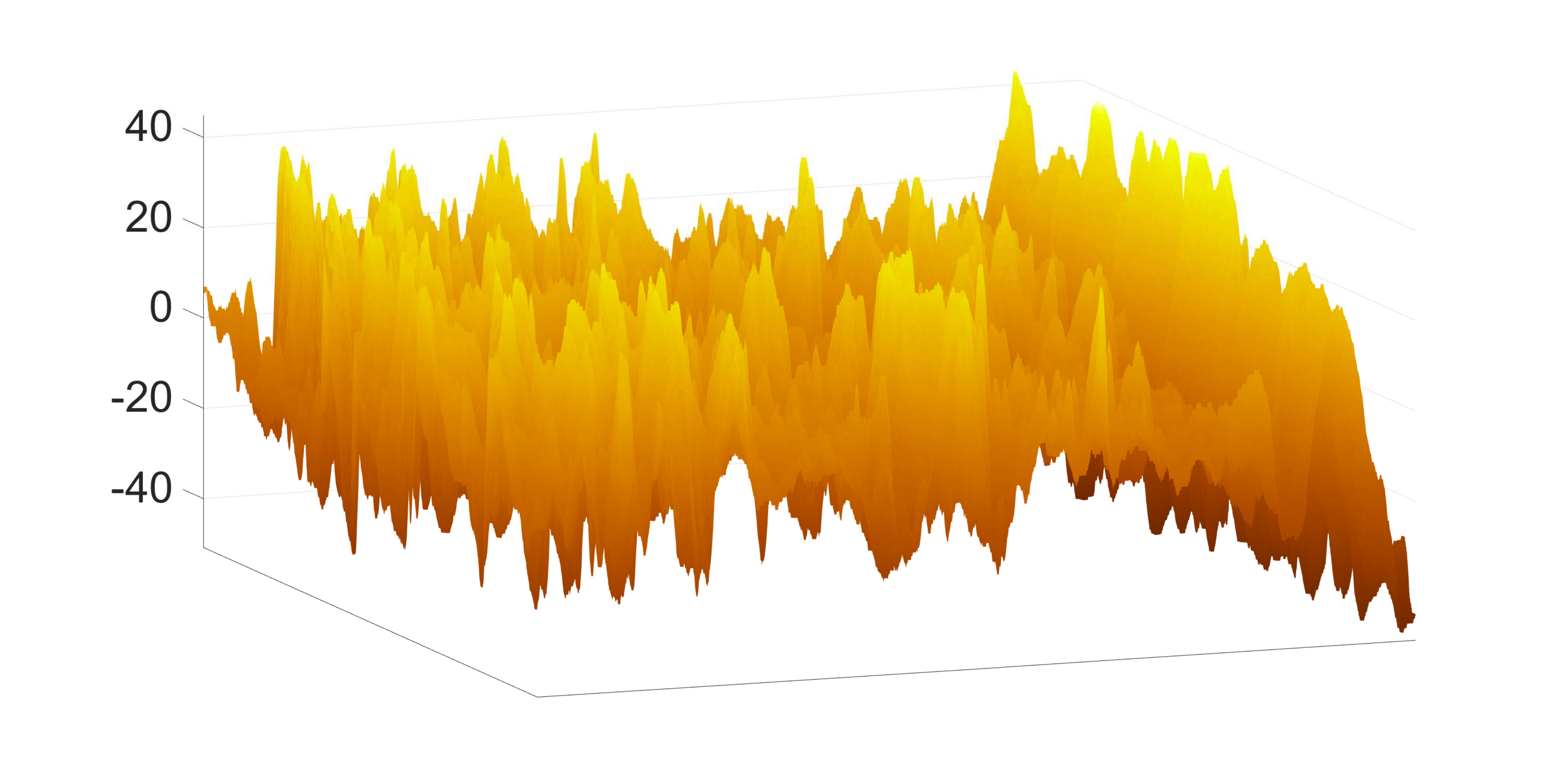}} \hfill
\subfloat[$\hat{H}_2$\label{er24}]{%
	\includegraphics[trim={1cm 1cm 1cm 1cm},clip,width=0.3\linewidth]{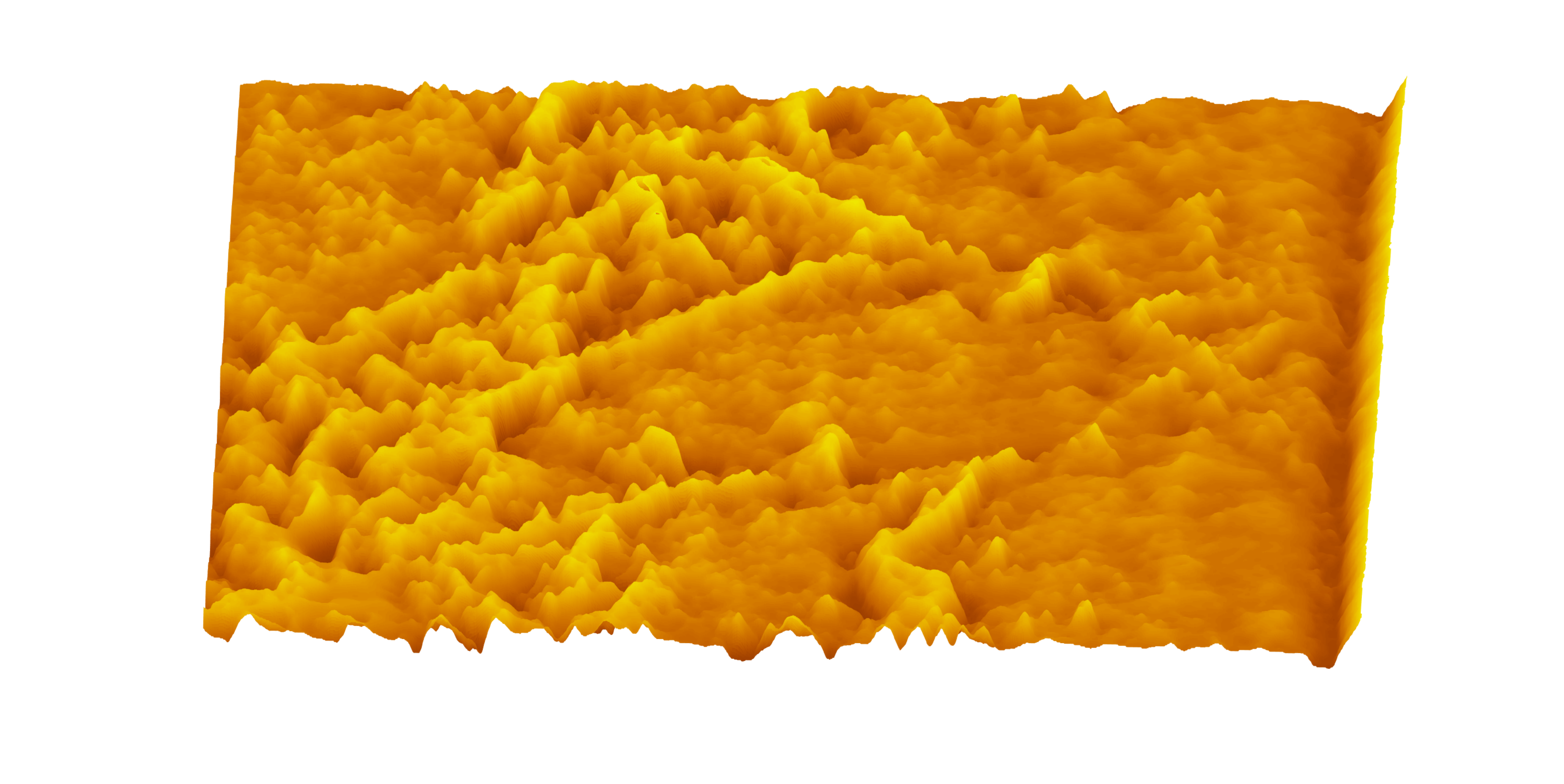}}\\
	\subfloat[$\hat{H}_1$\label{er13}]{%
	\includegraphics[trim={1cm 1cm 1cm 1cm},clip,width=0.3\linewidth]{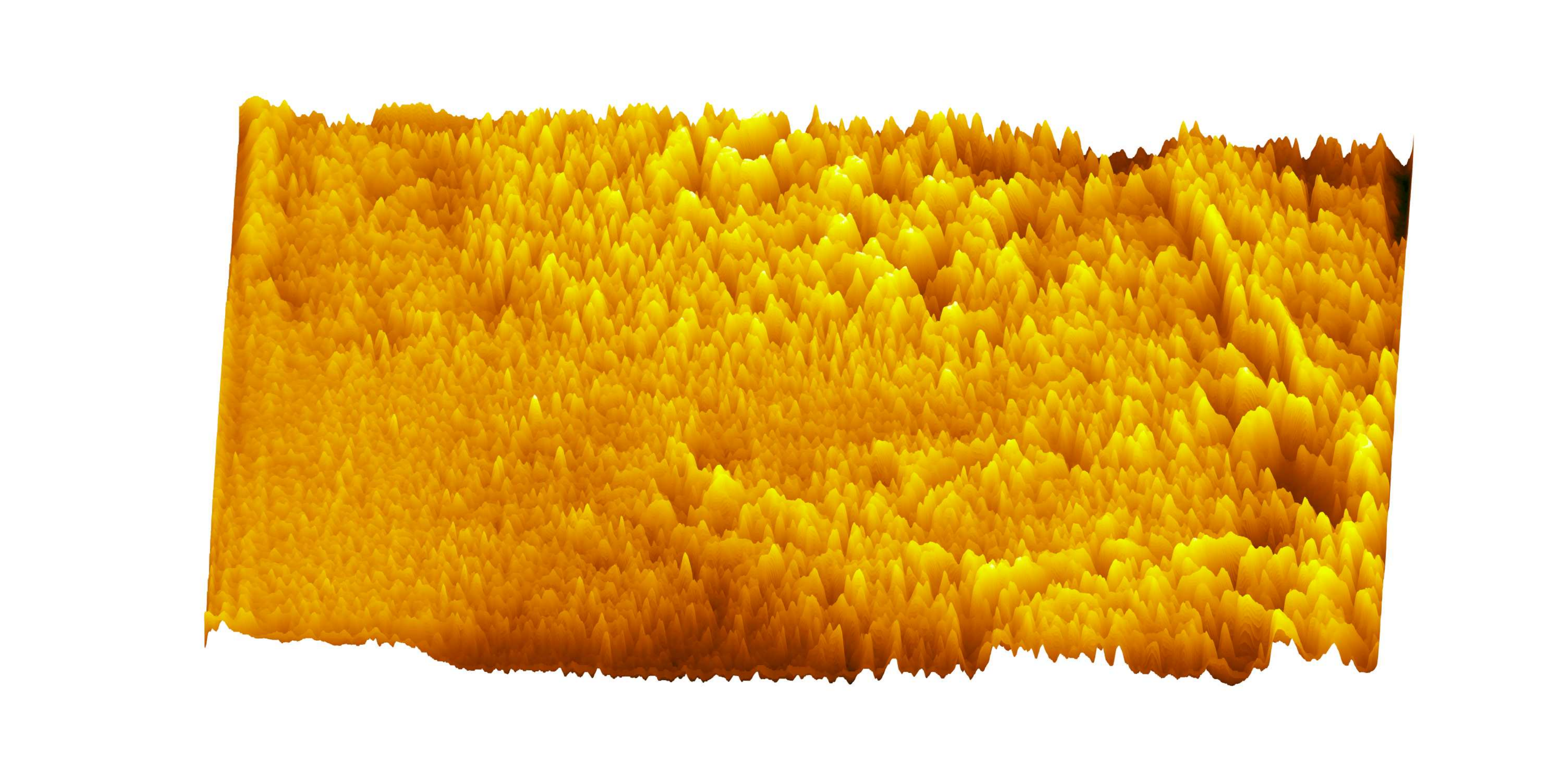}}\hfill
\subfloat[$H_2-H_1$\label{der13}]{%
	\includegraphics[trim={1cm 1cm 1cm 1cm},clip,width=0.3\linewidth]{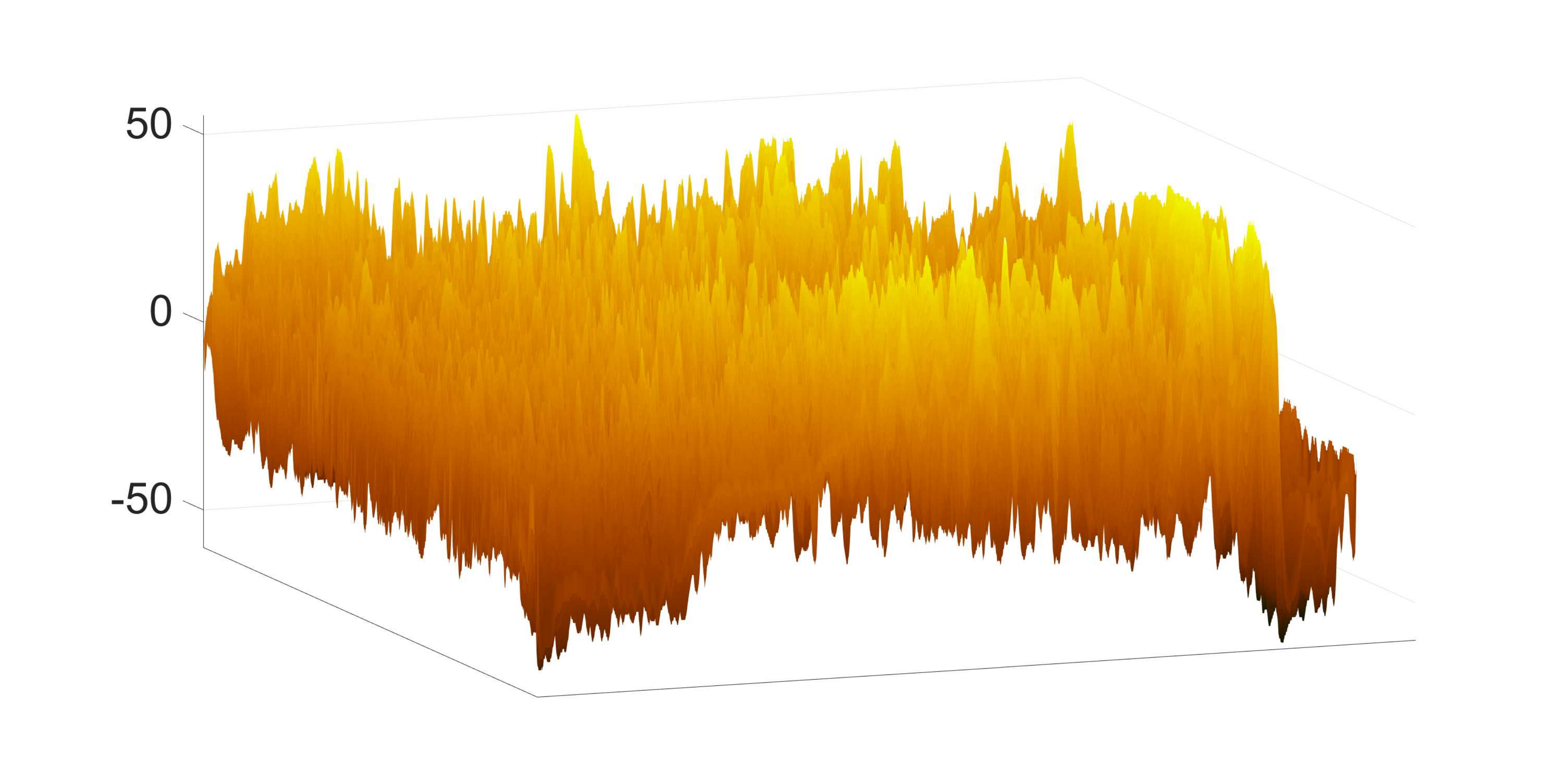}} \hfill
\subfloat[$\hat{H}_2$\label{er23}]{%
	\includegraphics[trim={1cm 1cm 1cm 1cm},clip,width=0.3\linewidth]{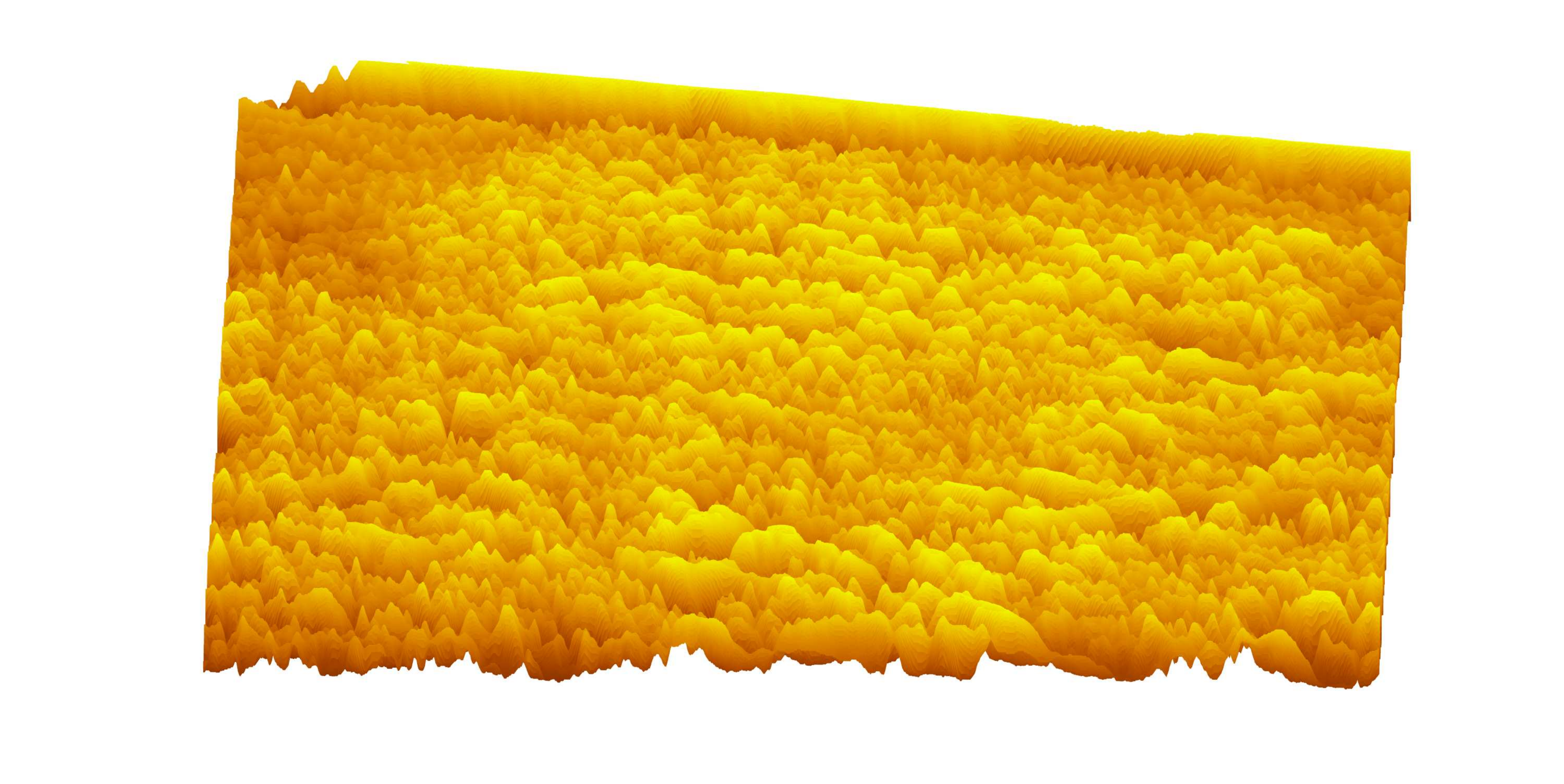}}\\
	\caption{Examples of difference maps with various magnitudes of $L_1$ error between desired and predicted seabed relief estimates on pairs of coregistered sand-ripple images. The residuals (b,e,h,k,n) have $L_1$ error 1, 2.5, 5, 15,  and 30 respectively.}
	\label{fig:scaleError} 
\end{figure*}

To demonstrate the differences between the magnitudes of the errors in Table \ref{tab:testMASAS}, we show the difference maps between pairs of looks with various $L_1$ errors in Figure \ref{fig:scaleError}.
	
		\begin{table}[t]
		\centering
		\caption{\label{tab:testMASAS} $L_1$ error between the desired and predicted for the hand-aligned SAS dataset for each model. The data each model was trained on is listed next to the model name. Models that were pre-trained with PISAS and fine-tuned with PoSSM are describes as name (PoSSM). Each model acheieves smaller error on the multi-aspect dataset than using the raw intensity. Each UNet model outperforms the GMRF model and the UNet-opt variants do better than all other models. Fine-tuning the output convolutional layer (UNet-opt-ftc) achieves the best performance.}
		\begin{tabular}{lc}
			\multicolumn{2}{c}{Multi-Aspect SAS} \\
			Model           & Error           \\ \hline
			\rowcolor[HTML]{E0E0E0} 
			Intensity (Baseline)          & $35.53\pm4.96$  \\
		    \rowcolor[HTML]{FFFFFF} 
			GMRF (PISAS) &  $32.49\pm12.45$ \\
			\rowcolor[HTML]{E0E0E0} 
			UNet (PISAS)               & $27.37\pm6.80$  \\
			\rowcolor[HTML]{FFFFFF} 
			UNet-opt-pt (PISAS)   & $6.89\pm1.52$ \\
			\rowcolor[HTML]{E0E0E0} 
			UNet-opt-ft (PoSSM)               & $6.51\pm1.33$  \\
			\rowcolor[HTML]{FFFFFF} 
			UNet-opt-ftc (PoSSM)   & $\bm{5.10\pm0.90}$ \\
		\end{tabular}
	\end{table}

\begin{figure}[ht]
			\centering
		\subfloat[$V_1$\label{cs1a}]{%
			\includegraphics[trim={1cm 1cm 1cm 1cm},clip,width=0.225\linewidth]{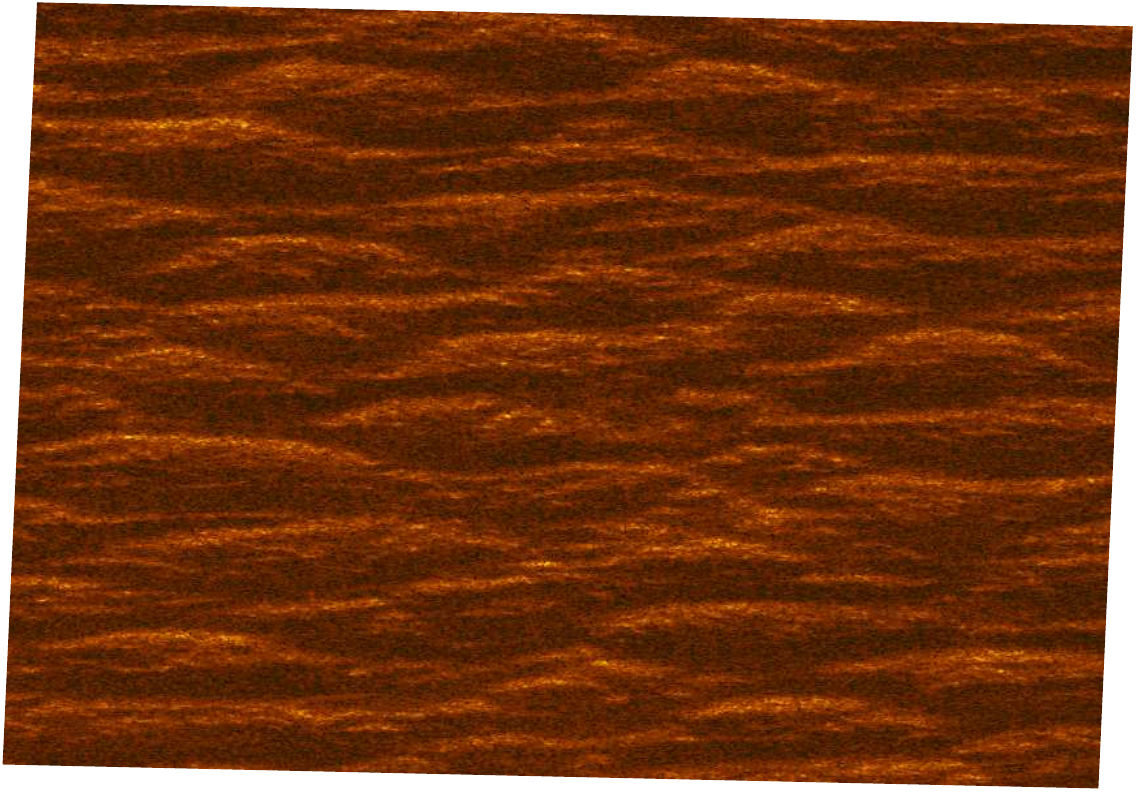}}
		\hfill
		\subfloat[GMRF\label{cs1b}]{%
			\includegraphics[trim={1cm 1cm 2cm 1cm},clip,width=0.25\linewidth]{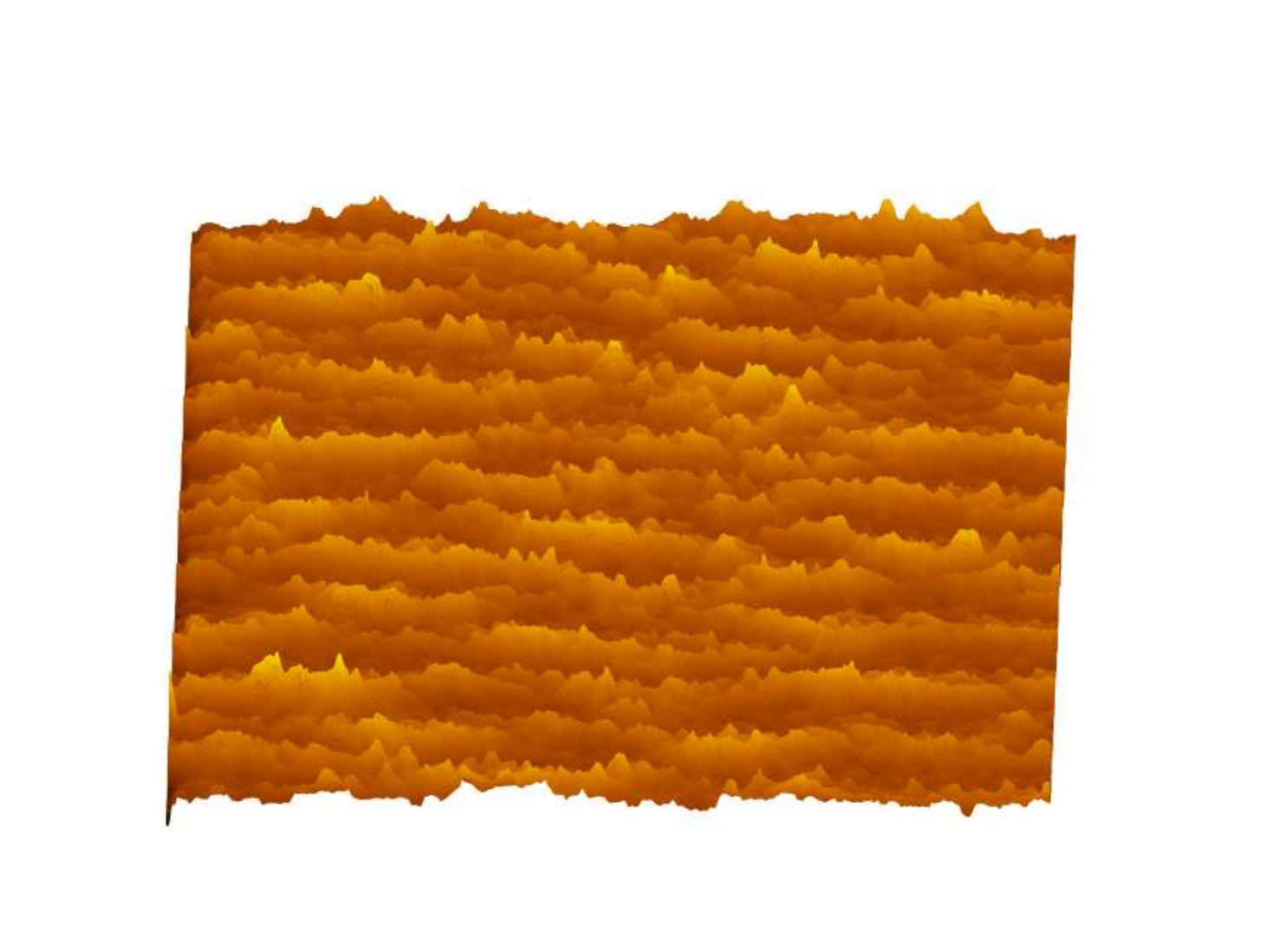}}
		\hfill
		\subfloat[UNet\label{cs1c}]{%
			\includegraphics[trim={1cm 1cm 2cm 1cm},clip,width=0.25\linewidth]{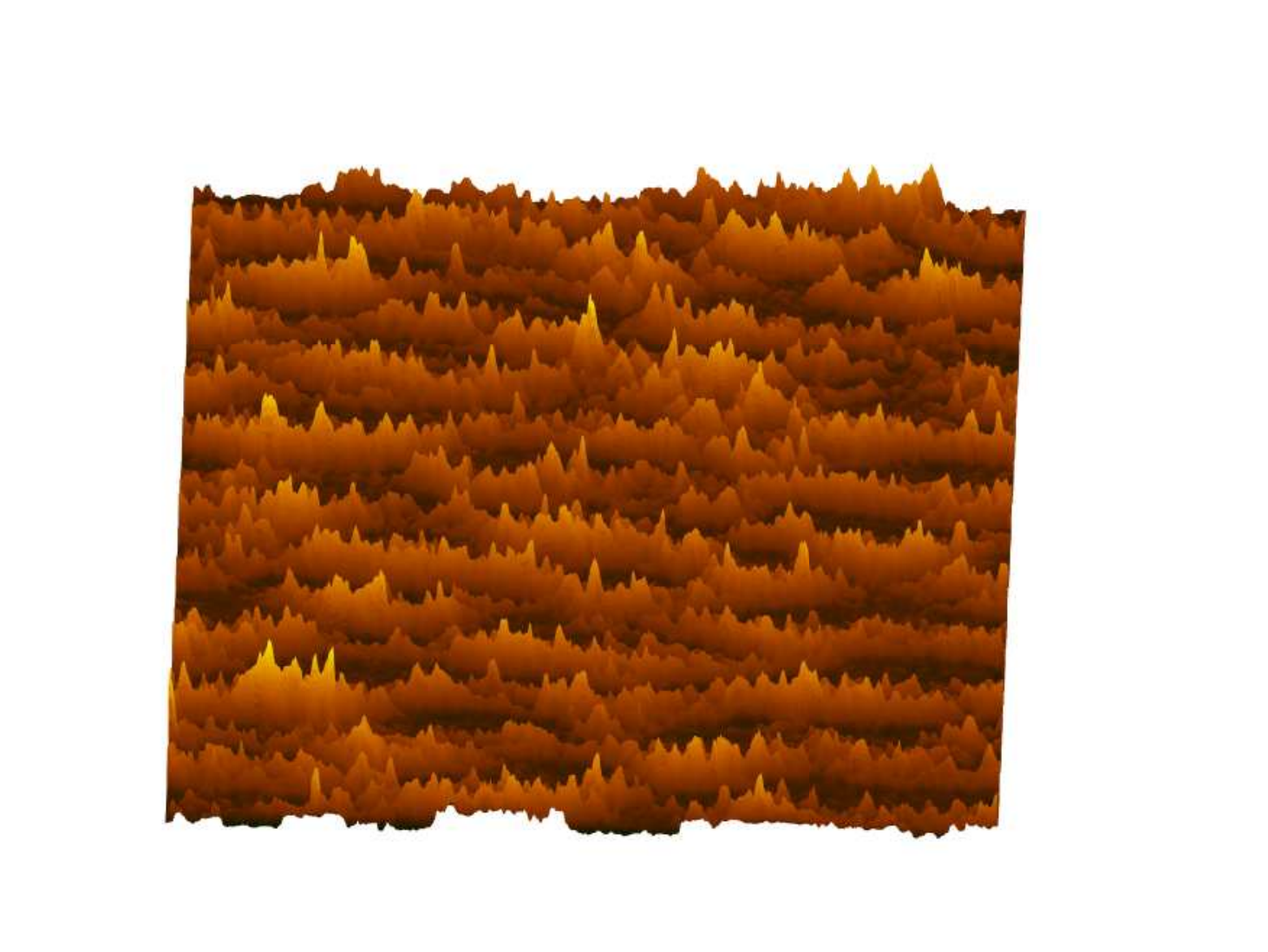}}
		\hfill
		\subfloat[UNet-opt-ftc\label{cs1d}]{%
			\includegraphics[trim={1cm 1cm 2cm 1cm},clip,width=0.25\linewidth]{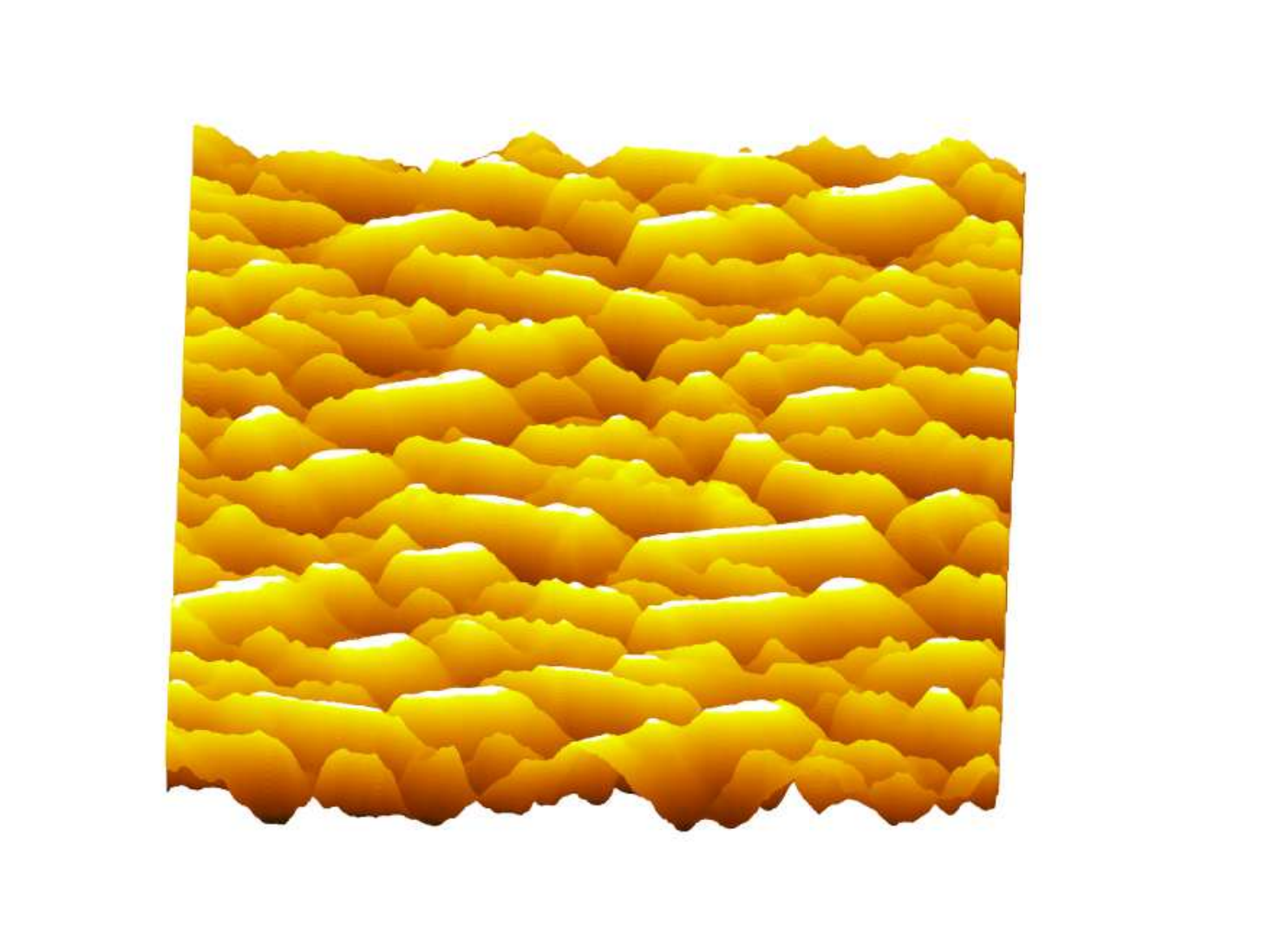}} \\
		
		\subfloat[$V_2$\label{cs2a}]{%
			\includegraphics[trim={1cm 1cm 1cm 1cm},clip,width=0.225\linewidth]{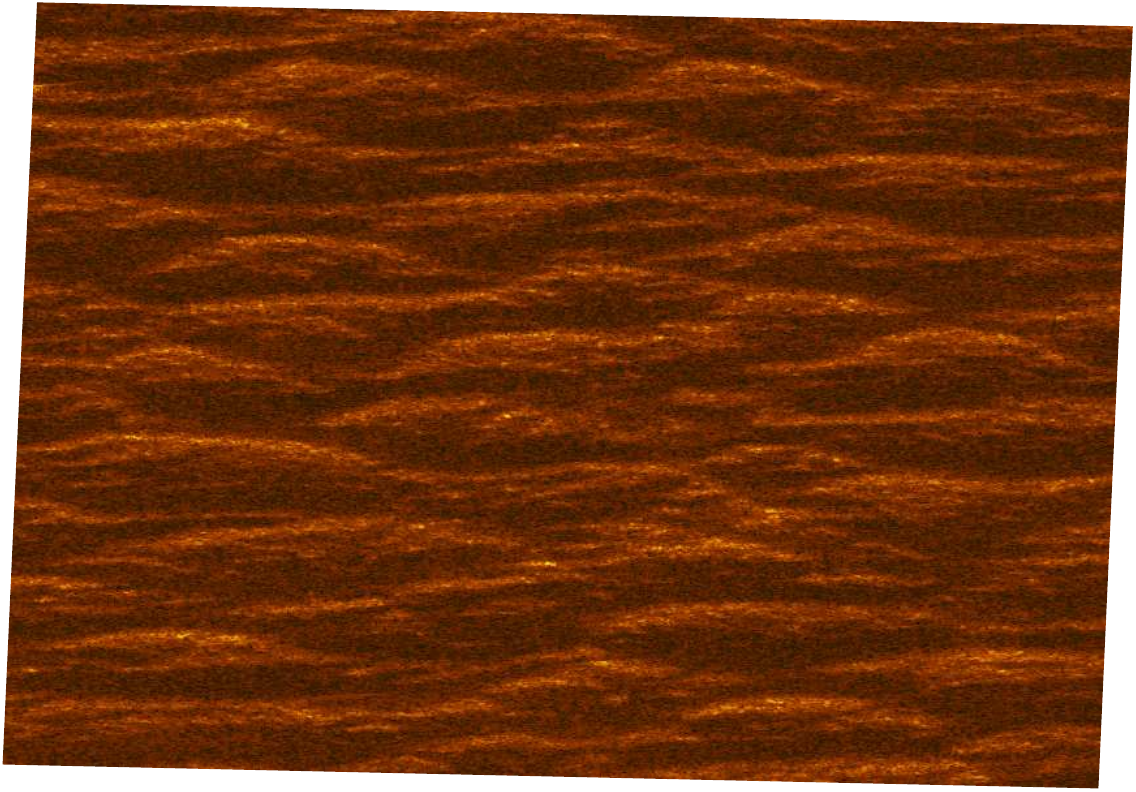}}
		\hfill
		\subfloat[GMRF\label{cs2b}]{%
			\includegraphics[trim={1cm 1cm 2cm 1cm},clip,width=0.25\linewidth]{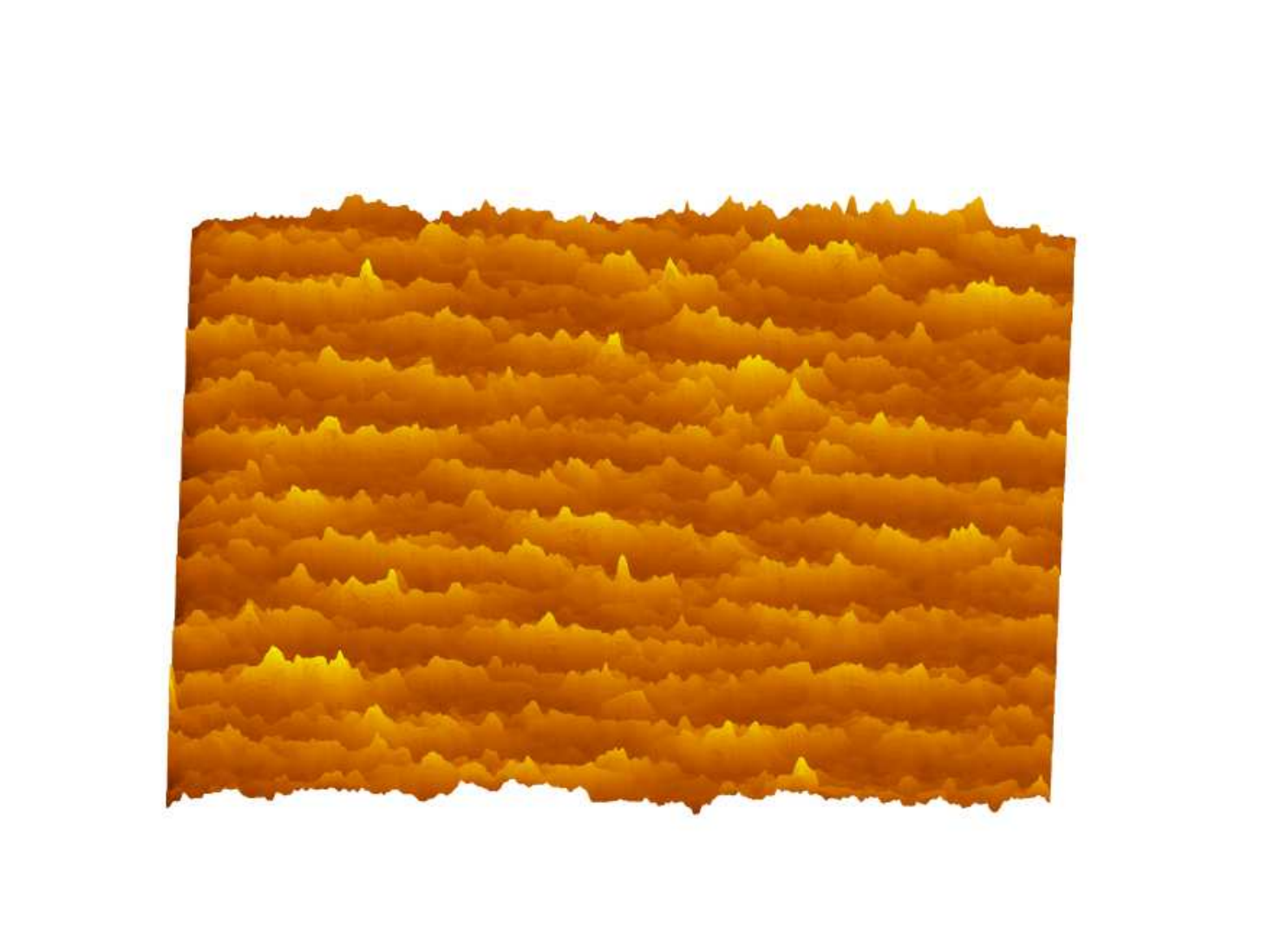}}
		\hfill
		\subfloat[UNet\label{cs2c}]{%
			\includegraphics[trim={1cm 1cm 2cm 1cm},clip,width=0.25\linewidth]{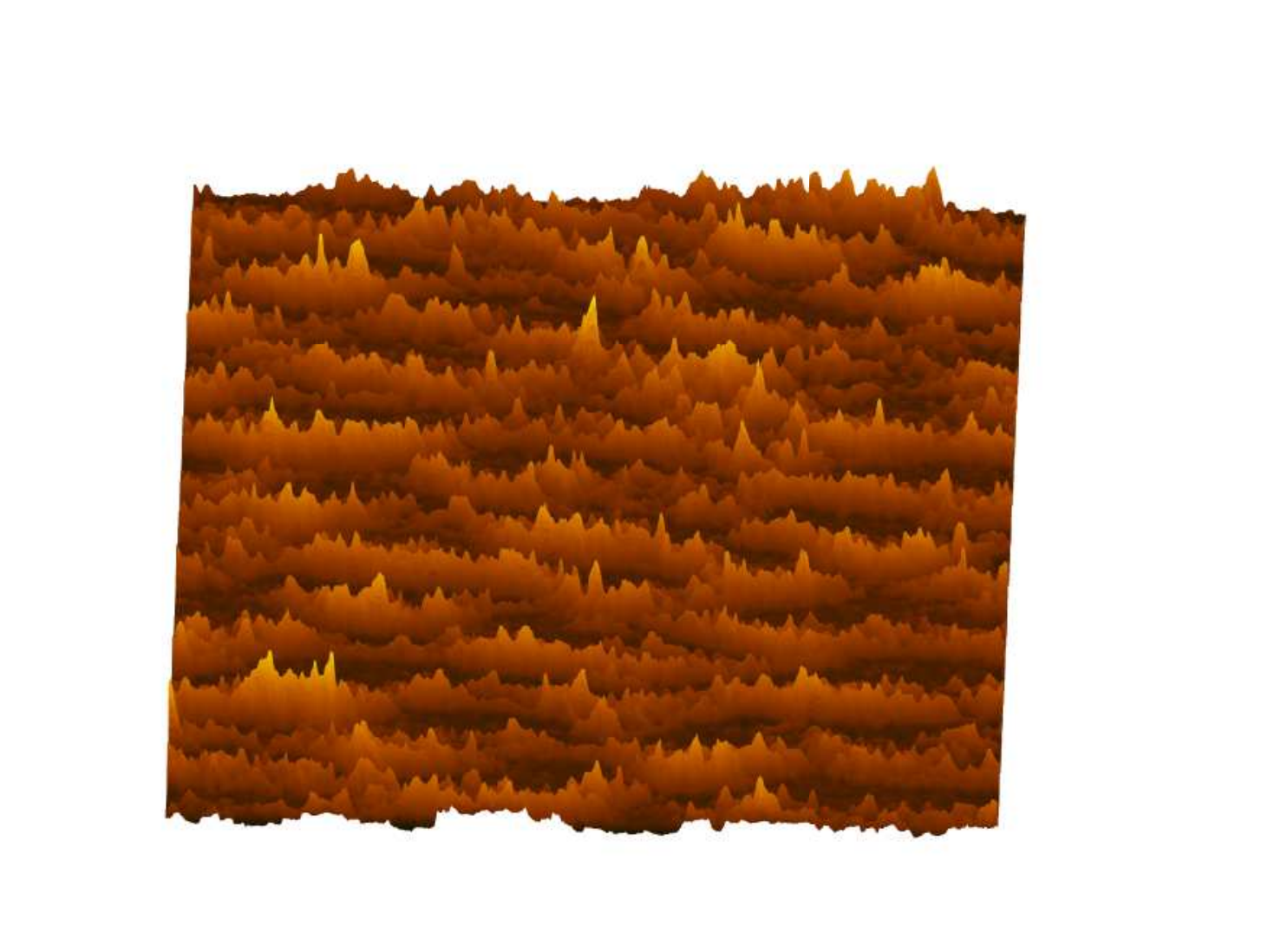}}
		\hfill
		\subfloat[UNet-opt-ftc\label{cs2d}]{%
			\includegraphics[trim={1cm 1cm 2cm 1cm},clip,width=0.25\linewidth]{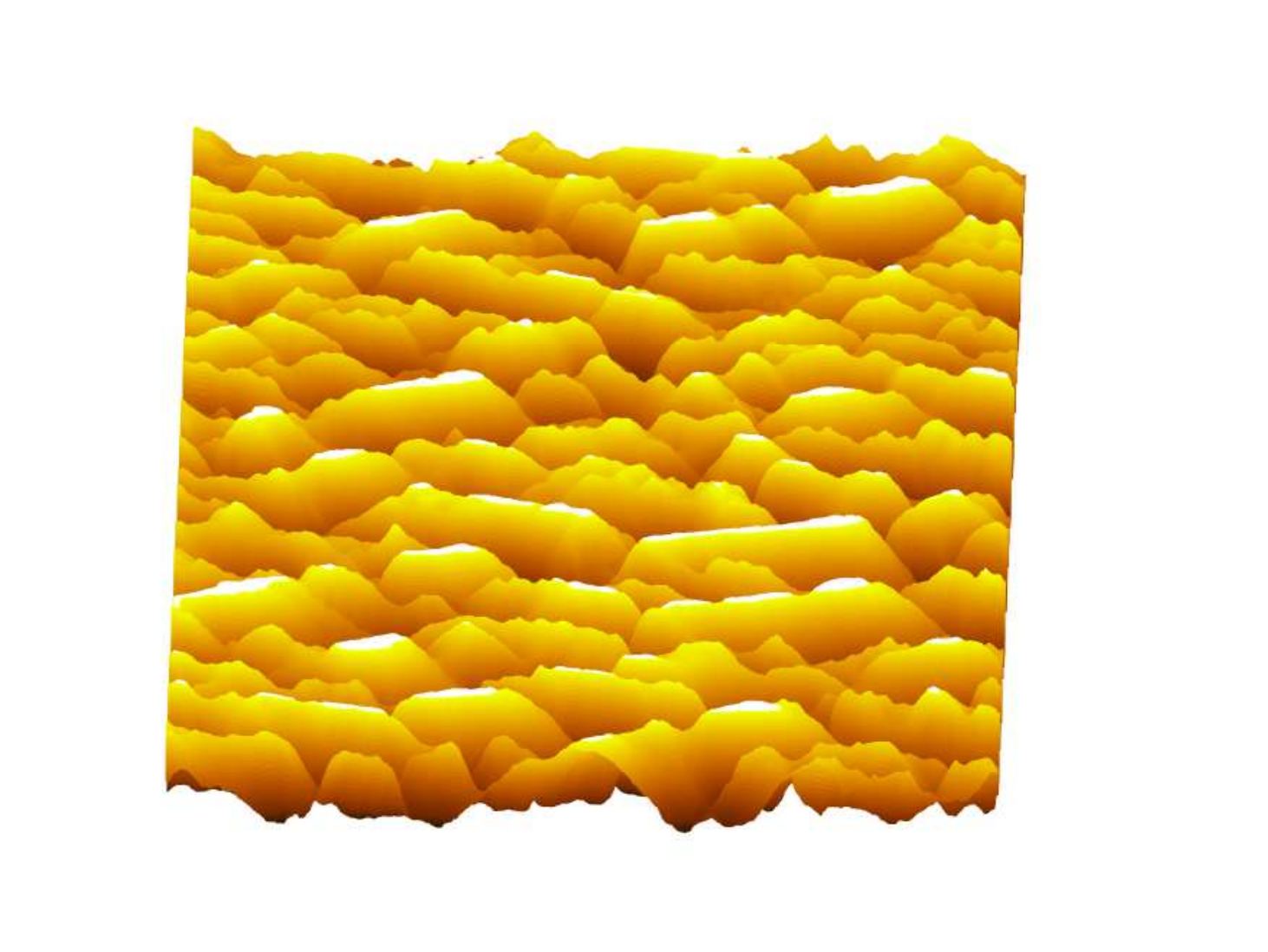}} \\

		\caption{Examples of seabed relief estimation on a pair of cSAS subaperture images containing sand-ripple patterns. Views $V_1$ and $V_2$ refer to a pair of coregistered intensity images (a,e) with $\Delta\phi=3.6\degree$ aspect difference. The GMRF model (b,f) and standard UNet produce noisy seabed relief maps (c,g). The UNet-opt-ftc model produces a seabed relief map with smooth ripples (d,h).}
		\label{fig:cSAS1} 
	\end{figure}

\begin{figure*}
		\includegraphics[width=\linewidth]{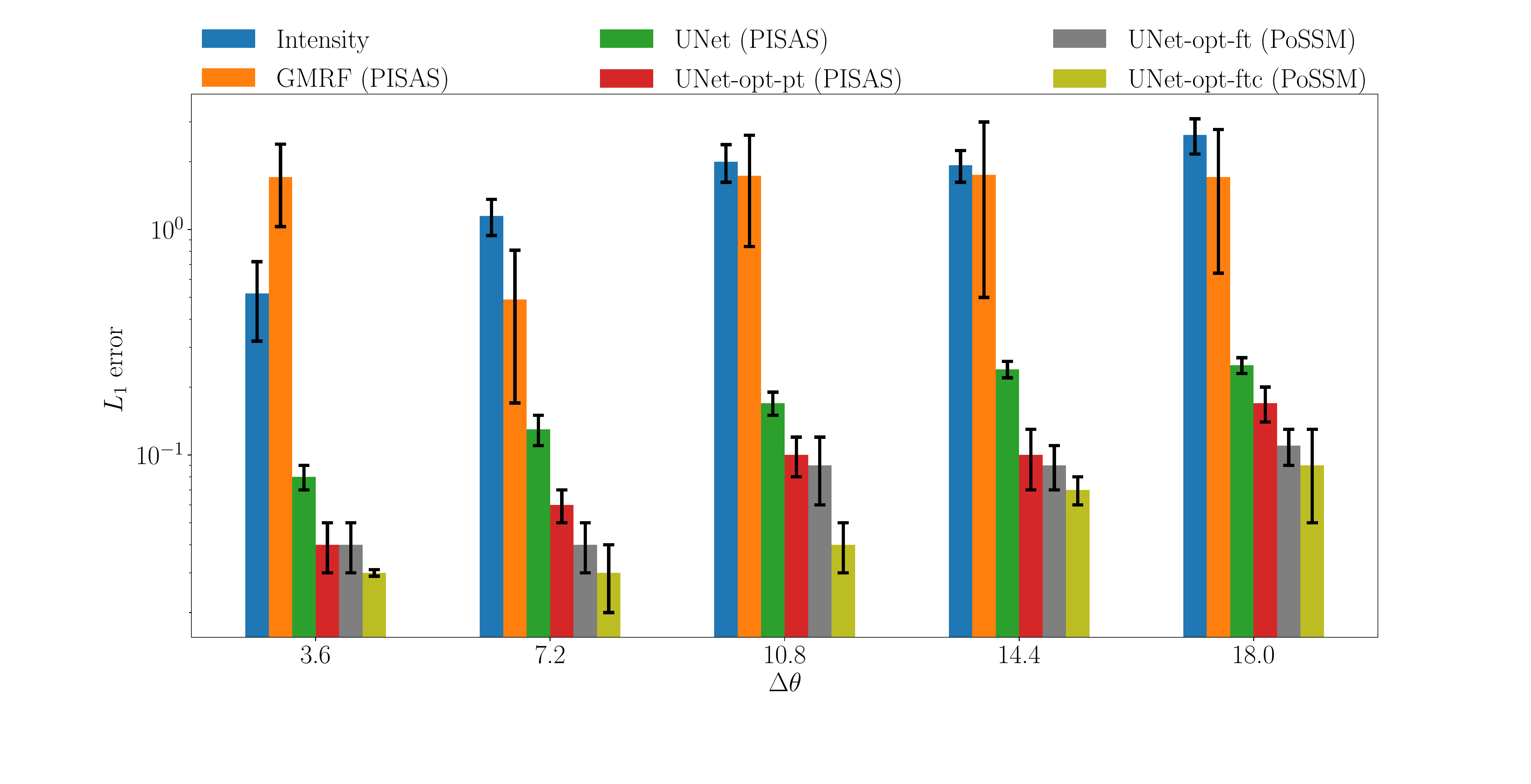}
	    \caption{Performance of each model with different training schemes on the cSAS test set. The data each model is trained on is listed next to the model name. Models with PoSSM listed were pre-trained with PISAS and fine-tuned with PoSSM. Other than the lowest angular differences ($\Delta\phi=3.6\degree$), all seabed relief estimation models produce more similar estimates than the raw intensity. The GMRF model slightly improves for each other angular difference, although the standard deviation is high for $\Delta\phi=14.4\degree$ samples. In each case the UNet models improve drastically in comparison to intensity. Each successive UNet model improves upon the performance of the previous models. The UNet-opt-ftc model outperforms all other models for every angular difference.}
		\label{fig:testcSAS} 
	\end{figure*}

	\paragraph{Circular SAS data}
	We apply our model to a dataset of aligned cSAS. Like the multiaspect SAS, there is no ground-truth seabed relief information for this dataset. The offset between subsequent images is $\Delta\phi=3.6\degree$. Therefore, we align pairs of images with varying angular differences ($\Delta\phi$) and measure the error in intensity and estimated seabed relief. The average and standard deviation for each set of angular differences is recorded to evaluate performance (Figure \ref{fig:testcSAS}). Estimated seabed relief maps for a pair of cSAS frames with $\Delta\phi=3.6\degree$ are shown in Figure \ref{fig:cSAS1}.

In the pair of cSAS frames (Figure \ref{fig:cSAS1}), two looks of sand-ripple with varying wavelengths are shown. Seabed relief maps produced by both the UNet and GMRF are noisy. Meanwhile, the UNet-opt-ftc model produces seabed relief maps that are the most similar across looks and not overly smooth. Although the ripples are similar to the training data, none of the models have been trained with ripples that have varying wavelengths present within a single image. However, the seabed relief maps produced by the UNet-opt-ftc model are able to estimate a seabed relief distribution analogous to the present ripples.
	
	Overall, error tends to increase as the angular difference increases (see Figure \ref{fig:testcSAS}). However, there are a few exceptions. The GMRF model has its smallest error for pairs that are $\Delta\phi=7.2\degree$. The error for the GMRF model is significantly larger than that of the UNet models. One can assume this is due to the lack of complexity of the model. Each of the UNet architectures rely on at minimum thousands of parameters rather than a few initial conditions. The error for the UNet-opt-pt models is the same between $\Delta\phi=10.8\degree$ and $\Delta\phi=14.4\degree$ with a slight increase in standard deviation. The UNet-opt-ft model has the same error between  $\Delta\phi=3.6\degree$ and $\Delta\phi=7.2\degree$ and $\Delta\phi=10.8\degree$ and $\Delta\phi=14.4\degree$. 
	
	Other than at the smallest angular differences, the GMRF model has reduced error compared to intensity. However, at each angular difference, the UNet as well as the UNet-opt variants each have substantially less error than the intensity and GMRF. The UNet-opt-pt model improves the similarity of information across aspect over relying on intensity information and the base UNet model estimates. Each fine-tuned model outperforms the pre-trained model other than the UNet-opt-ft model at the lowest angular difference. The UNet-opt-ftc model achieves the lowest error for each angular difference. This result shows the benefit of pre-training on low-fidelity and fine-tuning the last layer on high-fidelity simulations.  
	
	\subsection{Discussion}
	There are a few observations from our experiments that will be discussed further. First, the optimization of parameters for the pix2pix network was inherently difficult and may not necessarily represent the best possible results. This is a negative aspect of the cGAN approach. Although it could potentially obtain better results, the training procedure is difficult and hinders the success of the complex model. In addition to the 1,081,745 parameters for the generator network, a discriminator with over 250,000 parameters is needed to learn optimal parameters to distinguish between real and fake samples. The optimization for learning the pix2pix parameters is much more complex than the loss function we applied for the UNet architectures. Instead of investigating each possible combination of discriminator iterations and weighting each term of the multi-term loss function, we made a choice to investigate the UNet. Inherently, the UNet is much more simple to train and does not succumb to the same training failures that GANs do. Notably, the UNet base architecture that was used as the generator network for the pix2pix model was able to achieve better performance with a simple loss function. The base UNet achieved better results with 250,000 less parameters and a training time that is a fraction of the pix2pix. This lead us to discard the pix2pix network for further experiments. Future investigation could be done to try better training strategies for GANs, however, our interest in this work was to learn the simplest model that can perform domain translation of intensity to seabed relief. 
	
	Our investigation into UNet architectures provided insights into what part of the UNet architecture is most imporant for this problem. Reducing the depth of the UNet models led to higher error in each comparison with similar numbers of parameters.  Regardless of the number of channels used in each layer, the shallower UNet models each had worse performance than the deeper networks. This supports the importance of the skip connections by preserving more spatial information from the encoder stage of the network. Each network that had a reduced number of skip connections was worse than its counterpart.
	
	Additionally, our datasets are inherently imbalanced. For the 10 textures that were produced, four contained rough seabed relief maps with no ripples. The other six consisted of both ripples and roughness. Oversmoothing from pix2pix may be due to the large number of sand-ripple textures in the data. Parameters learned in each model were validated on random splits of the data that contain examples from multiple texture types. However, it is possible that the best parameters may have been those that bias for ripple textures. Nonetheless, the UNet models produced seabed relief maps consistent with each type of texture under much simpler training conditions. 
	
	Thirdly, the estimated seabed relief obtained from each domain translation method achieved smaller error between hand-aligned looks than intensity for the multi-aspect SAS dataset. While the GMRF had high standard deviation and only slightly less error than using intensity, this may be due to the reliance on a very simple model in comparison to the UNet architectures. The GMRF approach relies on an initial seabed relief and number of iterations while the UNet architectures each use at minimum thousands of parameters. The UNet-opt is able to use weights trained on low-fidelity samples to produce seabed relief maps from real SAS which correspond more than those from the more complex models. When trained solely on low-fidelity PISAS data, the UNet-opt uses $25\%$ as many parameters, but outperforms the UNet with an error decrease of $75\%$. Each variant of the UNet achieved better performance when fine-tuned with high-fidelity samples. Future work should investigate curriculum learning strategies to obtain possible improvements of the approach \cite{Bengio09}. Although the UNet-opt-ft model improved upon the pre-trained performance, the best model was one which only updated the output convolutional layer (UNet-opt-ftc). The UNet-opt-ftc model required nine parameters to be updated, rather than over 250,000 for the full model. Relying on stable weights for the entire model and fine-tuning the last nine led to a $22\%$ decrease in error with smaller standard deviation. This performance difference contradicts previous UNet fine-tuning experiments where the best results were achieved for fine-tuning the entire network \cite{amiri2020}. However, the application of semantic segmentation is unlike domain translation and may require different training strategies to optimize the models.

	Each model was trained with batches of samples with minimum $\Delta\phi = 5\degree$ differences in aspect. However, the UNet models were able to compare samples with $3.6\degree$ angular differences on the cSAS dataset and up to $45\degree$ differences in the multiaspect SAS dataset. No type of paired supervised training was performed. Batches of coregistered intensity and seabed relief were used to update parameters without any knowledge of how pairs of samples in a batch relate.
	
	Lastly, each model was trained on sand-ripple and roughness textures, not rocky textures. Our multi-aspect dataset contains some examples with rocky textures. In these experiments, the UNet architectures were able to relate pairs of looks without being trained on the specific textures present in the data. This warrants future investigation into what textures the model can generalize to without prior examples. The cSAS dataset contains highly variable sand-ripple patterns. While the models were trained with ripple patterns containing a single dominant ripple wavelength, each UNet model was able to relate images containing complex ripple patterns from real data.

	\section{Conclusion}
In this work, variations of the UNet architecture are proposed and investigated for domain translation of intensity to seabed relief. Each UNet model is able to learn a mapping from intensity to estimated seabed relief from simulated data. All UNet models outperform a GMRF and pix2pix model on predicting seabed relief from intensity imagery on the PISAS dataset. Comprehensive experiments demonstrate the ability of the UNet approaches to produce seabed relief estimates which are more similar across aspect than intensity information in both simulated and real data. The best UNet model (UNet-opt-ftc), when pre-trained on low-fidelity and fine-tuned on high-fidelity simulated samples, greatly outperforms the GMRF and standard UNet model on our multi-aspect and circular SAS datasets. This performance occurs on examples which contain textures that are not present in our training data. Additionally, the UNet-opt models outperform all other UNet variants and GMRF models on coregistered cSAS imagery. On interesting examples containing complex textures, the UNet-opt models are able to produce seabed relief maps with minimal error regardless of aspect differences. Results of the real data experiments support the stability of the UNet and its variants to produce seabed relief maps which relate more than intensity in interesting test cases.

Additionally, the UNet models present a few advantages over the comparison algorithms. Compared to the GMRF, the parameters are learned directly from the data. Rather than needing a grid search to select parameters, the UNet can be trained with a simple loss function and parameters are adapted to the training data. Learning these parameters is much easier than training a cGAN like the pix2pix model.

	
	%


	
	
	\bibliographystyle{IEEEtran}
	\bibliography{./references}
	%

	%
	
	
	
	
	
	
	

\end{document}